\documentclass{article}

    \PassOptionsToPackage{numbers, compress}{natbib}
 \usepackage[preprint]{neurips_2026}

\usepackage[utf8]{inputenc} 
\usepackage[T1]{fontenc}    
\usepackage{hyperref}       
\usepackage{url}            
\usepackage{booktabs}       
\usepackage{amsfonts}       
\usepackage{nicefrac}       
\usepackage{microtype}      
\usepackage{xcolor}         
\usepackage{amsmath}
\usepackage{amssymb}
\usepackage{mathtools}
\usepackage{amsthm}
\usepackage{bm}
\usepackage{caption}
\usepackage{bbm}
\usepackage{xcolor}
\usepackage{algorithm}
\usepackage{algorithmic}
\usepackage{wrapfig}
\usepackage[capitalize,noabbrev]{cleveref}
\usepackage{etoc}
\usepackage{multirow}
\usepackage{pifont}
\usepackage{makecell}
\newcommand{\cnum}[1]{\ding{\number\numexpr171+#1\relax}}
\title{DVD: Discrete Voxel Diffusion for 3D Generation and Editing}

\author{%
  Zhengrui Xiang\thanks{Work done during internship in Math Magic, Hitem3D.} \\
  Imperial College London\\
  \texttt{zx1321@imperial.ac.uk} \\
  \And
  Jiaqi Wu \\
  Math Magic \\
  \texttt{wujiaqi@mathmagical.com} \\
  \And
  Fupeng Sun \\
  Imperial College London \\
  \texttt{f.sun23@imperial.ac.uk} \\
  \And
  Heliang Zheng \\
  Math Magic \\
  \texttt{zhengheliang@mathmagical.com} \\
  \And
  Yingzhen Li \\
  Imperial College London \\
  \texttt{yingzhen.li@imperial.ac.uk} \\
}

\usepackage{xcolor}
\usepackage{xspace}

\def\E{{\mathbb E}}

\def \x{\mathbf{x}}

\def \m{\mathbf{m}}
\def \s{\mathbf{s}}
\def \p{\mathbf{p}}

\def \bpi{\bm{\pi}}

\newcommand{\SLAT}{\textsc{SLat}\xspace}

\begin{document}

\maketitle

\vspace{-15pt}
\begin{abstract}
We introduce \emph{Discrete Voxel Diffusion} (DVD), a discrete diffusion framework to generate, assess, and edit sparse voxels for \textsc{SLat} (Structured LATent) based 3D generative pipelines. Although discrete diffusion has not generally displaced continuous diffusion in image-like generation, we show that it can be an effective first-stage prior for sparse voxel scaffolds. By treating voxel occupancy as a native discrete variable, DVD avoids continuous-to-discrete thresholding and provides a simple framework for voxel generation, uncertainty estimation, and editing. Beyond quality gains, DVD provides more interpretable generation dynamics through explicit categorical modeling. Furthermore, we leverage the predictive entropy as a robust uncertainty metric to identify ambiguous voxel regions and complicated samples, facilitating tasks such as data filtering and quality assessment. Finally, we propose a lightweight fine-tuning strategy using block-structured perturbation patterns. This approach empowers the model to inpaint and edit voxels within a single sampling round, requiring negligible auxiliary computation and no additional model evaluations. Code is available at \url{https://github.com/TeCai/DVD}.
\end{abstract}

\vspace{-10pt}

\etocdepthtag.toc{main}
\section{Introduction}\label{sec:introduction}
\begin{figure*}[htbp]
    \centering
    \includegraphics[width=\linewidth]{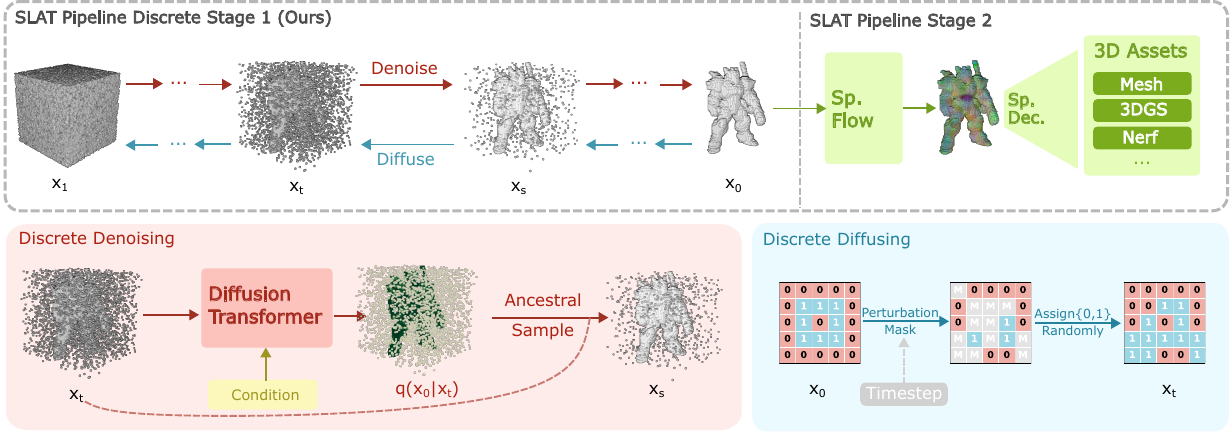}
    \caption{\textsc{SLat} pipeline and uniform state discrete diffusion (USDM). We abuse the notation here to denote a whole voxel data as $\x_0$. 
    \textbf{Top:} The two-stage \textsc{SLat} pipeline. Stage~1 generates low-resolution sparse voxels. Stage~2 predicts a latent vector for each occupied voxel using flow matching with a sparse transformer (\emph{Sp.~Flow}) and decodes the latents (together with the scaffold) into downstream 3D representations via a sparse decoder. We substitute Stage~1 with discrete voxel diffusion.
    \textbf{Bottom:} Uniform state discrete diffusion. \emph{Reverse process}: given a noisy sample $\x_t$, the model predicts the distribution over clean states $q(\x_0 | \x_t)$ (darker voxels indicate higher occupancy probability) and draws $\x_{s}$ with ancestral sampling (refer to Section~\ref{sec:DiscreteDiffusionModels}). 
    For clarity, we visualize predictions only at occupied locations, although the model outputs $q(\x_0 \mid \x_t)$ for all grid positions. 
    \emph{Forward process}: the forward process randomly reassigns each voxel to $0$ or $1$ with probability depend on $t$. 
    }
    \label{fig:pipeline}
\end{figure*}

Recent progress in 3D generative modeling has enabled scalable creation of assets for applications such as game development, embodied AI, and immersive media. 
However, high-quality 3D generation remains challenging because 3D data admits diverse representations, including meshes, signed distance fields (SDFs), point clouds, 3DGS~\cite{kerbl20233dgaussiansplattingrealtime}, etc. 
Unlike images, these representations differ substantially in topology, geometry, and decoding procedure, making the choice of representation central to both modeling and generation quality. 
One influential direction represents 3D assets as compact sets of latent vectors, which are decoded into implicit fields or surfaces~\cite{zhang20233dshape2vecset3dshaperepresentation,wu2024direct3dscalableimageto3dgeneration}. 
Such vector-set representations have powered many recent systems~\cite{yang2025hunyuan3d10unifiedframework,zhao2025hunyuan3d20scalingdiffusion,lai2025hunyuan3d25highfidelity3d,zhang2024claycontrollablelargescalegenerative,BANG,li2025triposghighfidelity3dshape}, producing high-quality and smooth meshes. 
Nevertheless, they often rely on sophisticated preprocessing to obtain training targets such as SDFs, and their sampled implicit-field training can make fine-grained geometric structures difficult to capture.

An alternative line of work adopts hybrid voxel-anchored representations, where a sparse voxel scaffold is generated first, and continuous latent features are then generated and attached to active voxels for high-fidelity decoding~\cite{xiang2025structured3dlatentsscalable,chang2025reconviagenaccuratemultiview3d,li2025sparc3dsparserepresentationconstruction,lai2025latticedemocratizehighfidelity3d,jia2025ultrashape,wu2025direct3d,xiang2025native}. 
This formulation, commonly referred to as \SLAT (Structured LATent representation), combines explicit spatial structure with expressive latent features. 
This hybrid representation enables intuitive operations such as editing and inpainting while enhancing fine-grained geometric details.

A key but often under-examined component in \SLAT-based pipelines is the \emph{first stage}: generating the sparse voxel scaffold. 
Since the downstream stage predicts latent features on the active voxels and decodes them together with the scaffold, errors in voxel occupancy can propagate to the final mesh. 
Existing frameworks typically implement this stage by compressing the voxel grid into continuous VAE latents~\cite{kingma2013auto} and modeling them with continuous diffusion models~\cite{ho2020denoising}. 
While effective, this design is not fully aligned with the native structure of the data: voxel occupancy is sparse and binary, but is generated continuously and discretized by decoding and thresholding. 
Such a continuous-to-discrete conversion can reduce interpretability and be less suitable for downstream operations such as post-editing.
In practice, we observe failure cases where small errors near the binarization boundary flip voxels to empty, leading to local holes that persist through later meshing (see Figure \ref{fig:contHoles}).

This motivates modeling sparse voxel occupancy directly as discrete data. 
Discrete diffusion is a natural candidate: sparse 3D voxel grids do not admit an obvious preferred generation order, while diffusion models denoise voxel states in parallel and are therefore well matched to long voxel sequences. 
However, this choice is not guaranteed to improve quality, since continuous diffusion remains dominant and often stronger in image generation~\cite{sahoo2025diffusionduality,austin2023structureddenoisingdiffusionmodels}. 
We therefore examine whether discrete diffusion can serve as a practical first-stage prior for sparse voxel generation.

We find that discrete diffusion provides a practical and informative first-stage prior for sparse voxel generation. 
Modeling occupancy directly as categorical states improves sparse voxel quality over continuous counterparts across most evaluation metrics and exposes voxel-wise predictive probabilities throughout the reverse process. 
This probabilistic voxel interface enables us to inspect generation dynamics, identify ambiguous geometric regions, derive shape-level uncertainty scores, and support localized editing through simple fine-tuning. 
Our main contributions are summarized as follows:

\begin{itemize}
\item \textbf{Discrete diffusion for sparse voxel generation.} We introduce a discrete diffusion formulation for generating sparse voxels in \SLAT-based 3D pipelines, avoiding the continuous-to-discrete conversion and achieving improved performance on most evaluation metrics with reduced model size and training data.
\item \textbf{Uncertainty at voxel and shape level.}
We show that predictive uncertainty in discrete diffusion localizes ambiguous (``difficult'') voxels, and we derive a lightweight entropy-based shape-level uncertainty score by aggregating voxel entropies, enabling applications such as data filtering and quality assessment.
\item \textbf{Efficient voxel editing via block-structured perturbations.} We propose a block-structured perturbation fine-tuning strategy that enables efficient voxel editing and inpainting \emph{within a single sampling process}, without additional neural network evaluations beyond the standard sampler.
\end{itemize}

\section{Related Work}

\paragraph{\SLAT-style 3D generative models.}
Recent 3D generative systems increasingly adopt two-stage pipelines that first generate explicit sparse spatial voxels and then synthesize higher-fidelity geometry/appearance latents attached to them.
XCube \cite{ren2024xcubelargescale3dgenerative} generates high-resolution sparse voxel grids using a hierarchical sparse-voxel hierarchy (VDB) and a coarse-to-fine latent diffusion process.
TRELLIS \cite{xiang2025structured3dlatentsscalable} introduces \SLAT  representation and a two-stage generation scheme that first predicts sparse structure and then predicts per-cell structured latents decodable to multiple 3D formats (e.g., meshes and 3D Gaussians).
SparC3D \cite{li2025sparc3dsparserepresentationconstruction} combines a sparse deformable marching-cubes representation with a sparse-convolutional VAE to enable near-lossless high-resolution surface compressing.
ReconViaGen \cite{chang2025reconviagenaccuratemultiview3d} integrates diffusion-style 3D generative priors into multi-view reconstruction and introduces mechanisms to improve cross-view consistency and controllability during denoising.
LATTICE \cite{lai2025latticedemocratizehighfidelity3d} and UltraShape 1.0 \cite{jia2025ultrashape} sample a portion of generated sparse voxels for the second stage instead of using all of them. Although numerous \SLAT-based methods have been proposed, the majority generate voxels either through continuous diffusion models operating on VAE latents or by voxelizing alternative 3D representations.
Direct3d-S2 \cite{wu2025direct3d} proposed Spatial Sparse Attention, which significantly improved the efficiency of training and inference on sparse volumetric data.
TRELLIS.2 \cite{xiang2025native} proposes a native omni-voxel (O-Voxel) for the second stage,
enhancing the generation quality.
In contrast, our work revisits the first stage and directly models sparse voxel occupancy using discrete diffusion mechanisms.

\paragraph{Discrete diffusion models.} Discrete diffusion models~(DDMs) \cite{austin2023structureddenoisingdiffusionmodels,campbell2022continuoustimeframeworkdiscrete,gat2024discreteflowmatching,campbell2024generativeflowsdiscretestatespaces, shi2025simplifiedgeneralizedmaskeddiffusion}  connect a prior noise distribution and data distribution with continuous time Markov Chains (CTMCs) over discrete state spaces. While the prior distribution could be arbitrary, they are mainly split into two parts: uniform state diffusion models (USDMs) and mask diffusion models (MDMs). USDMs have a uniform prior distribution across all possible states and all tokens, while MDMs mask every position with a special MASK token. 
Discrete diffusion models have been developed and studied in various tasks, such as text generation \cite{sahoo2024simpleeffectivemaskeddiffusion}, image generation \cite{you2025effectiveefficientmaskedimage}, multimodal modeling \cite{swerdlow2025unified}, protein generation~\cite{campbell2024generativeflowsdiscretestatespaces}, and pose estimation \cite{wang2024text}, etc. For applications of discrete diffusion in 3D tasks, \citet{song2025topologysculptorshaperefiner} investigated mesh generation using discrete diffusion models, and TD3D~\cite{zhao2024td3d} leveraged discrete diffusion models for shape generation in a quantized latent space, scaffold diffusion~\cite{jung2025scaffold} used DDMs to generate labels for given voxels. Most relevant to our work, CADD~\cite{hu2025large} and pyramid diffusion~\cite{liu2024pyramiddiffusionfine3d} leverage discrete diffusion for generating large 3D voxel scenes. However, large scene generation is less sensitive to small errors compared to object-level generation, and they mainly focused on generation results without further investigating the benefits of discrete modeling. 

\paragraph{Diffusion inpainting.}
Inpainting for continuous diffusion models has been widely studied, including sampling-based constraint enforcement and posterior sampling formulations \cite{lugmayr2022repaint,kawar2022denoising,daras2024survey}. These method often contains additional network evaluations or complicated post-computation.
For masked diffusion models (MDMs), inpainting is largely built into the training objective: the model is trained to predict masked tokens conditioned on the unmasked context \cite{secretcondition}. In contrast, inpainting for USDMs is less explored. While techniques developed for continuous diffusion—such as conditioning on corrupted inputs or enforcing known regions during sampling—can often be adapted in principle \cite{lugmayr2022repaint,saharia2022palette}, they may incur additional sampling overhead (e.g., extra sampling steps) or require task-specific conditioning.
\section{Method}

\subsection{Generation Pipeline}\label{sec:DiscreteDiffusionModels}

\paragraph{Structured latent representation.}
For a 3D asset $\mathcal{O}$, the geometry and appearance information is encoded by an \SLAT \cite{xiang2025structured3dlatentsscalable} representation $\mathcal{S}$ on a 3D grid:
\begin{equation}
    \mathcal{S} = \{ (\s^i, \p^i) \}^{L'}_{i=1},\; \s^i\in\mathbb{R}^C,\;\p^i\in\{0,1,\cdots,N-1\}^3,
\end{equation}
where $\p^i$ is the positional index of the $i$-th active (occupied) voxel, and $\s^i$ denotes the local latent vector attached to it. $N$ is the resolution of the 3D grid, and $L'$ is the total length of the active voxels. The active voxels serve as spatial scaffolds that determine the rough shape of 3D assets, while the latent vector encodes local details.
The generation process of $\mathcal{S}$ consists of two stages (Figure \ref{fig:pipeline}): 
\begin{itemize}
    \item  In stage 1, the active voxels $\{\p^i\}^{L'}_{i=1}$ are generated.
    \item In stage 2, continuous latent vectors  $\{\s^i\}^{L'}_{i=1}$ are generated on the active voxels via flow matching. Then,  $\{\p^i\}^{L'}_{i=1}$ and $\{\s^i\}^{L'}_{i=1}$ are jointly decoded to various 3D assets.
\end{itemize}
In our work, we leverage a discrete diffusion model for stage 1 and keep stage 2 the same as TRELLIS.
In practice, instead of generating active indices directly, a binary voxel occupancy grid is generated as
$\mathbf{X}\in\{0,1\}^{N\times N\times N}$, where $\mathbf{X}[\p]=1 \iff \p\in\{\p^i\}^{L'}_{i=1}$.

\paragraph{Discrete diffusion models.} Following the notations of \cite{schiff2025simpleguidancemechanismsdiscrete}, we denote a scalar discrete random variable with $K$ possible values as one-hot vectors, whose set is denoted as $\mathcal{V}=\{\x \in \{0,1\}^K:\sum^K_{i=1}\x_i=1\}$. Denote $\Delta$ as the $K$-simplex, and Cat$(\cdot;\bpi)$ as the categorical distribution given by $\bpi\in\Delta$. Let $\x^{1:L}\in \mathcal{V}^L$ and $\{\x^i\}^{L}_{i=1}$  denote a sequence of length $L$. For sparse voxels, each position of $\mathbf{X}$ is modeled as a Bernoulli variable ($K=2$) and results in a sequence length of $L=N^3$.

Denote the distribution of data $p_{data}$ as $p_0$, and some prior distribution as $p_1=\text{Cat}(\cdot;\bpi)$. Also, denote $\x_0$ as a data sample. For $t\in[0,1]$, we can get the marginal distribution of the forward process $q_t$ by interpolating $p_0$ and $p_1$:
\begin{equation}
    p_t(\cdot|\x_0;\alpha_t) = \text{Cat}(\x_t;\alpha_t\x_0 + (1-\alpha_t)\bpi),
\end{equation} where $\alpha_t$ is a strictly decreasing function of $t$ with boundary conditions $\alpha_1\approx0$, $\alpha_0\approx1$.

Denote $\x_t$ as a random variable follows the  marginal distribution $p_t$, the true reverse posterior $q_{s|t}$ for $s<t$ given $\x_0$ writes: 
\begin{equation}\label{equ:posterior}
    q_{s|t}(\mathbf{x}_s|\mathbf{x}_t,\mathbf{x}_0) = \text{Cat}\Big(\mathbf{x}_s;
    \frac{[\alpha_{t|s}\mathbf{x}_t+(1-\alpha_{t|s})\mathbf{1}\bpi^\intercal\mathbf{x}_t] \odot [\alpha_s\mathbf{x}_0 + (1-\alpha_s)\bpi]}{\alpha_t\mathbf{x}_t^\intercal\mathbf{x}_0+(1-\alpha_t)\mathbf{x}_t^\intercal\bpi}\Big),
\end{equation}
where $\odot$ denotes the Hadamard product and $\alpha_{t|s}$ denotes $\alpha_t/\alpha_s$. During inference, we replace $\x_
0$ with the predictive posterior $q(\x_0|\x_t)$, approximated by a neural network $\x_\theta:\mathcal{V}\times[0,1]\rightarrow\Delta^K$. The raw output $\tilde\x_\theta(\x_t,t)$ is often the unnormalized log probability across categories, and we can obtain the probability as $\x_\theta(\x_t,t)$ by a softmax operation. 
A single denoising step is performed via ancestral sampling~\cite{sahoo2025diffusionduality}, i.e., sample from predictive posterior $q^\theta_{s|t}(\x_s|\x_t):=q_{s|t}(\x_s|\x_t,\x_\theta)$.

When applied to voxels $\x^{1:L}$, the forward process and backward process are factorized over tokens (positions) as $p_t(\x_t^{1:L}|\x_0^{1:L};\alpha_t) = \prod_{i=1}^{L}p_t(\x_t^i|\x_0^{1:L};\alpha_t)$ and $q^\theta_{s|t}(\x_s^{1:L}|\x_t^{1:L}) = \prod_{i=1}^{L}q_{s|t}(\x_s^i|\x_t^{1:L},\x_\theta(\x_t^{1:L},t))$. The denoising model $\x_\theta:\mathcal{V}^L\times[0,1]\rightarrow\Delta^{KL}$ predicts all tokens given the whole sequence of data.
The training of $\mathcal{\x_\theta}$ is by minimizing the negative ELBO~\cite{sahoo2025diffusionduality}:
\begin{equation}\label{equ:elbo}
    \text{NELBO}= \sum^L_{i=1}\E_{t\sim \mathcal{U}[0,1],p_t}\left[f(\x^i_t,\x^i_\theta(\x^{1:L}_t,t),\alpha_t;\x_0^i)\right],
\end{equation} where $\x^i_\theta(\x^{1:L}_t,t)$ denote the prediction of $i$th voxel and $f$ defined in Appendix \ref{appendix:ELBODM}.

For USDMs, $\bpi=\mathbf{1}/K$; for MDMs $\bpi=\mathbf{m}$ where $\m\in\mathcal{V}$ is a special mask token. During early experiments, we found that USDMs perform better than MDMs (discussion in Appendix~\ref{appendix:PriorDiffusionModels}). Thus, without further clarification, a uniform prior is used by default for the rest of the paper.

\subsection{Uncertainty of Voxels}
In contrast to continuous diffusion models, which typically approximate a Gaussian predictive posterior by estimating the mean \cite{ho2020denoising}, discrete diffusion models explicitly parameterize the predictive posterior as a full categorical distribution. 
Given a timestep $t$ and latent state $\mathbf{x}_t^{1:L}$,  $\mathbf{x}_\theta(\mathbf{x}_t^{1:L}, t)$ represents the predictive distribution over all possible states. 
This allows us to quantify the model's uncertainty by the predictive entropy. Precisely, we compute a weighted average entropy across all $L$ voxels given a data $\x_0^{1:L}$:
\begin{equation}
    \gamma_{t,\rho}(\x_0^{1:L}) = \log \left[ \frac{1}{L} \sum_{i=1}^L 
    \mathbbm{1}_{\left[H^i_\theta(\x_t^{1:L})>\rho\right]}H^i_\theta(\x_t^{1:L})\right],
\end{equation} where $H^i_\theta(\x_t^{1:L}) = H\!\left({\mathbf{x}}_\theta^{\,i}(\mathbf{x}^{1:L}_t, t)\right)$ indicates the entropy of the output probability of the $i$th voxel.
Here, $\gamma_{t,\rho}$ denotes the expected predictive entropy of the model at timestep $t$, computed after assigning zero weight to voxels whose entropy falls below a threshold $\rho$. When $t$ is close to $1$, $\mathbf{x}_t^{1:L}$ is highly corrupted, and the model often exhibits high entropy in its prediction. When $t \approx 0$, the corrupted input $\mathbf{x}_t^{1:L}$ is almost identical to the ground-truth $\mathbf{x}^{1:L}_0$, and the model is generally confident. Nevertheless, the entropy still varies across samples. 
Evaluation of voxel data can be performed by computing $\gamma_{t,\rho}$ at a near-zero timestep $t=\epsilon$, preserving most of the structural information.
In this regime, the model is confident about the majority of voxels, causing the entropy values to concentrate near zero. Thus, we analyze these values on a logarithmic scale to improve discriminability between assets.  As an evaluation of voxel ambiguity, shapes with larger $\gamma_{\epsilon,\rho}$ are often more intricate examples.
often contain more detailed or intricate structures.  We find $\gamma_{\epsilon,\rho}$ remains stable even without conditioning (see Section.~\ref{sec:generationDynamics}), suggesting this uncertainty is a property of the voxel's shape regardless of the type of conditioning. The proposed uncertainty score is versatile and can be applied across various downstream tasks. In Section~\ref{sec:generationDynamics}, we demonstrate a practical application by employing the score in data augmentation.

\subsection{Block-Structured Perturbations for Fine-tuning}
\label{sec:multi_scale}
In voxel editing, a common setting is to preserve a subset of voxels specified by a binary mask $\mathbf{M}_b$
, while allowing modifications to the complementary region.
However, standard discrete models are trained with spatially homogeneous noise and rarely encounter the heterogeneous mixture of clean context and corrupted target inherent to inpainting. This training-inference mismatch might cause the model to fail to generate structures consistent with the given voxels. 

\begin{wrapfigure}{r}{.5\linewidth}
    \centering
    \vspace{-1em}
    \includegraphics[width=\linewidth]{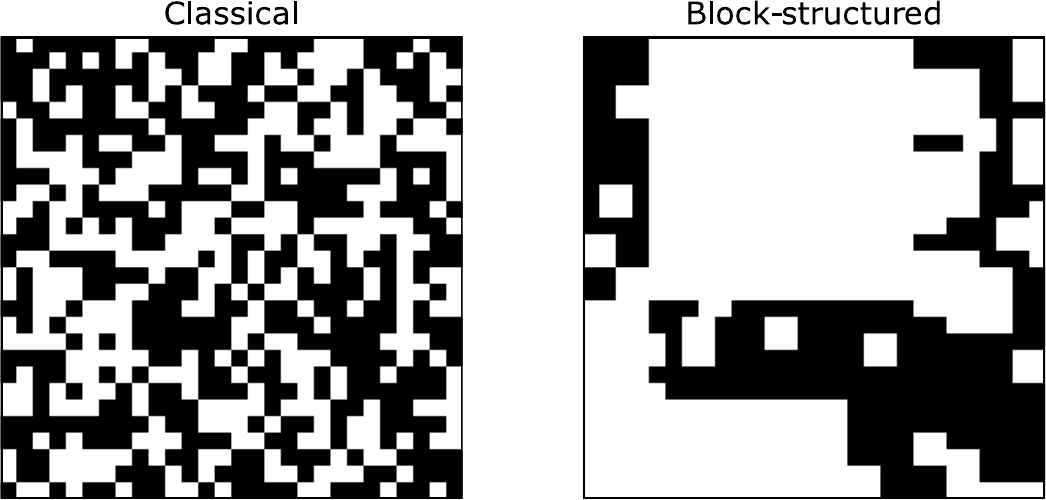}
    \caption{Perturbation Pattern Examples. The white area indicates the positions where the voxels should be perturbed.}
    \label{fig:ComparePertubation}
    \vspace{-1em}
\end{wrapfigure}

To address this mismatch, we introduce a simple yet effective
\emph{block-structured perturbation} (BSP) strategy for fine-tuning. Instead of
perturbing every voxel independently, we randomly select several axis-aligned hypercubes (``blocks'') at multiple scales, and apply the
forward corruption kernel only within these blocks. Intuitively, this creates
training examples that closely resemble inpainting or local editing scenarios:
Some contiguous regions are highly corrupted, while the rest of the volume
remains largely intact, as shown in Figure \ref{fig:ComparePertubation}. This increases the frequency of training examples where the model must predict a heavily corrupted region conditioned on a largely clean context.

These block patterns can be generated by sampling a union of axis-aligned hypercubes with different side lengths. For 
each length, we
\begin{itemize}
    \item [(i)] Compute an upper bound on the number of
blocks that would approximately achieve a target perturbation fraction $t$.
    \item [(ii)] Allocate a random number of blocks at each scale via normalized random weights.
    \item [(iii)]  Place blocks at random locations on the $N^\text{dim}$ grid.
\end{itemize}The final perturbation block is given by the union of all blocks; overlaps are allowed
and naturally reduce the realized coverage. We provide algorithms for generating these block patterns and an inpainting algorithm in Appendix \ref{appendix:PurturbInpaint} as Algorithm \ref{alg:block_masks} and \ref{alg:udlm_ancestral_inpainting}.

\section{Experiments} \label{sec:experiments}

Our experiments address three questions: (i) whether discrete diffusion improves the \SLAT generation quality compared to continuous baselines, (ii) what generation dynamics and uncertainty signals the discrete formulation provides, and (iii) the effectiveness of BSP fine-tuning for voxel inpainting.

\paragraph{Datasets.}
We use the same training set as TRELLIS presented ($\sim$500K); the actual number of data we used is around 90\% of it ($\sim$450K) due to some failure cases in the data preprocessing stage. For evaluation, we randomly select 2500 meshes from the training set, which we call the training subset later. Also, for evaluating generalization ability, we randomly select 1000 meshes from the Toys4K \cite{stojanov2021using} dataset, which consists of our second evaluation set. 

\paragraph{Training and sampling settings.}
For training, we applied classifier-free guidance (CFG) \cite{ho2022classifierfreediffusionguidance,schiff2025simpleguidancemechanismsdiscrete,nisonoff2025unlocking} with a drop rate of 0.1 and an AdamW \cite{loshchilov2019decoupledweightdecayregularization} optimizer with a learning rate of $1e-4$ with a weight decay of $1e-2$. We train a medium-sized ($\sim$400M) model for both image and text tasks, with shared time encoding DiTs.  The image model is trained for 400K steps with a batch size of 128. The text model is trained for 600K steps, as we have observed a continuous improvement in the generation results. 
The CFG strength for image generation tasks is set to be 0.7 if $t>0.5$ else 0.4, and 1.0 if $t>0.5$ else 0.4 for text generation tasks.  Here we use a sampling step of 256 ($\sim$20s on a NVIDIA A800 GPU), because for discrete diffusion models, an obvious performance drop may be observed \cite{sahoo2025diffusionduality} for a constrained number of function evaluations (NFEs). See Appendix~\ref{appendix:MoreImplementationDetails} for more implementation details.

\paragraph{Baselines.}
We leverage stage 1 from TRELLIS as the most direct continuous comparison and use its pretrained checkpoints. 
Their checkpoints for both image and text generation have around 550M parameters and are trained with a batch size of 256 for 400K steps. We also include image-to-3D tasks to Direct3D-S2 \cite{wu2025direct3d} and TripoSR \cite{li2025triposghighfidelity3dshape} as broader baselines.

\begin{table*}[t]
\centering
\caption{Quantitative comparison on the training subset and Toys4K.}
\label{tab:metric_all}
\setlength{\tabcolsep}{3.5pt}
\small
\scalebox{0.9}{
\begin{tabular}{llcccccccccc}
\toprule
\multirow{2}{*}{Dataset} & \multirow{2}{*}{Method}
& \multicolumn{6}{c}{Image-to-3D}
& \multicolumn{4}{c}{Text-to-3D} \\
\cmidrule(lr){3-8}
\cmidrule(lr){9-12}
&& CLIP $\uparrow$
& FID$_{\text{D}}$ $\downarrow$
& FID$_{\text{PC}}$ $\downarrow$
& CD $\downarrow$
& FID$_{\text{V}}$ $\downarrow$
& CD$_{\text{V}}$ $\downarrow$
& CLIP $\uparrow$
& FID$_{\text{D}}$ $\downarrow$
& FID$_{\text{PC}}$ $\downarrow$
& FID$_{\text{V}}$ $\downarrow$ \\
\midrule

\multirow{4}{*}{\makecell{Training\\subset}}
& TripoSR
& 82.05 & 321.22 & 14.45 & 0.0910 & 10.20 & 0.0281
& - & - & - & - \\
& Direct3D-S2
& - & - & 4.09 & 0.0045 & 4.11 & 0.0127
& - & - & - & - \\
& TRELLIS$^\dagger$
& 87.52 & 87.53 & 2.60 & 0.0048 & 1.99 & 0.0199
& \textbf{25.59} & 221.57 & 5.34 & 4.68 \\
& Ours
& \textbf{87.66} & \textbf{84.76} & \textbf{2.43} & \textbf{0.0029} & \textbf{1.87} & \textbf{0.0107}
& 25.54 & \textbf{201.70} & \textbf{3.91} & \textbf{3.24} \\
\midrule

\multirow{4}{*}{Toys4K}
& TripoSR
& 83.87 & 374.05 & 21.45 & 0.0113 & 16.92 & 0.0316
& - & - & - & - \\
& Direct3D-S2
& - & - & 7.42 & 0.0036 & 6.91 & 0.0089
& - & - & - & - \\
& TRELLIS$^\dagger$
& 87.13 & 106.71 & \textbf{4.51} & 0.0039 & 3.78 & 0.0150
& \textbf{26.10} & 328.18 & 8.97 & 7.93 \\
& Ours
& \textbf{88.85} & \textbf{104.64} & 4.64 & \textbf{0.0024} & \textbf{3.69} & \textbf{0.0085}
& 26.05 & \textbf{312.21} & \textbf{8.74} & \textbf{7.30} \\
\bottomrule
\end{tabular}}
\vspace{2pt}

{\footnotesize $^\dagger$ Metric reimplemented by us.}
\end{table*}
\paragraph{Metrics.}
To evaluate the generation ability, we compute FID \cite{heusel2018ganstrainedtimescaleupdate} of the rendering images under the feature space of DINOv2 \cite{oquab2024dinov2learningrobustvisual}, noted as $\text{FID}_{\text{D}}$. We compute the point cloud FID of the finalized mesh and cubified voxels under the feature space of PointNet++  \cite{qi2017pointnetdeephierarchicalfeature}, denoted as $\text{FID}_{\text{PC}}$ and $\text{FID}_{\text{V}}$, respectively.  We compute the CLIP score~\cite{radford2021learning} between the rendered images of the generated assets and the condition. For the image-conditioned task, we compute Chamfer Distance between the generated and GT mesh as a metric of reconstruction, denoted as CD. We also report the chamfer distance of the GT voxel and the generated voxel as $\text{CD}_{\text{V}}$, by treating the voxel centers as points. For certain experiments, negative log likelihood~(NLL) is reported as an evaluation metric for discrete diffusion models.
While computing metrics, we pre-align the generated voxels with the GT before we pass them to the second stage to minimize the influence caused by orientation mismatch. 

\subsection{Quantitative Results}
We present the results in Table \ref{tab:metric_all}. 
Quantitatively, our approach outperforms the continuous baseline across a majority of evaluation metrics while remaining competitive with existing state-of-the-art methods. In image-conditioned tasks, DVD demonstrates superior prompt alignment and geometric fidelity, as evidenced by improved metrics. Significantly, these gains are achieved using a more compact architecture and a more restricted training set, suggesting that discrete modeling provides a more effective inductive bias for representing the binary nature of sparse voxel occupancy. While text-conditioned CLIP scores are slightly lower due to limited training supervision, we anticipate that scaling the text-labeled dataset will bridge this gap, further leveraging the structural advantages of our discrete formulation.

\subsection{Generation Results}
We present some of the generation results for image and text prompts in Appendix~\ref{appendix:moreGenResults}. We also present qualitative comparisons of the DVD-generated voxel and the continuously generated voxel with different second-stage results in the Appendix~\ref{appendix:diffpipeline}.

\begin{figure}[tbp]
    \centering
    \includegraphics[width=0.49\linewidth]{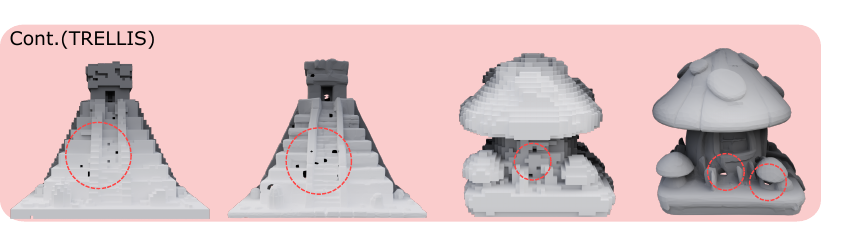}
    \includegraphics[width=0.50\linewidth]{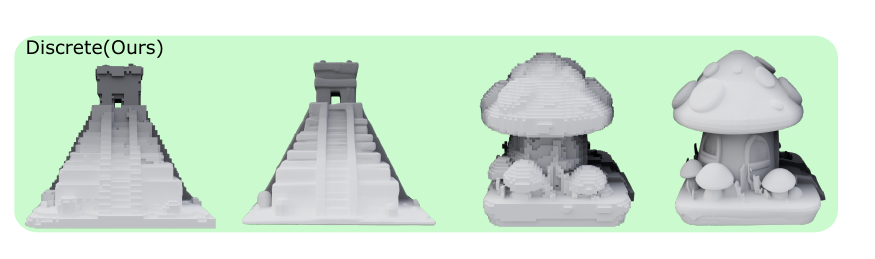}
    \caption{Degraded examples with continuous methods. We present some of the generated voxels and mesh of TRELLIS (Left) and our method (Right). The lightness of the mesh represents the distance of the mesh from the camera. Holes are circled out.  }
    \label{fig:contHoles}
\end{figure}
As mentioned in section~\ref{sec:introduction}, we sometimes observe holes in voxels generated by the continuous method, see Figure \ref{fig:contHoles} for an example.
In the continuous generation pipeline, voxels are obtained by decoding latent vectors and thresholding at $0$: the voxel is active if the decoded value is greater than $0$. This could pose a natural difficulty in ``thin-layer'' structures: the resulting continuous field needs to be positive on a thin layer of voxel, and negative in other places. These sharp ridges will pose extra difficulty for fitting the VAE, as small reconstruction errors around the zero level set can flip the sign and erase the voxel, causing holes or disconnected fragments.

\subsection{Generation Dynamics and Uncertainty Assesment}\label{sec:generationDynamics}
\begin{table*}[tb]
\centering
\setlength{\tabcolsep}{1.5pt}
\small
\caption{Quantitative comparison with and without augmented data fine-tune. The total training budget is 400K steps; with augmentation, the model is fine-tuned under the augmented dataset from 300K to 400K steps.}
\label{tab:LikelihoodAfterextend}
\scalebox{0.89}{
\begin{tabular}{lcccccccccccccc}
\toprule
\multirow{2}{*}{\raisebox{-0.5ex}{Method}}
& \multicolumn{7}{c}{Toys4K} & \multicolumn{7}{c}{Training subset}
\\
\cmidrule(lr){2-8}\cmidrule(lr){9-15}
& NLL $\downarrow$
& CLIP $\uparrow$
& FID$_{\text{D}}$ $\downarrow$
& FID$_{\text{PC}}$ $\downarrow$
& CD $\downarrow$
& FID$_{\text{V}}$ $\downarrow$
& CD$_{\text{V}}$ $\downarrow$
& NLL $\downarrow$
& CLIP $\uparrow$
& FID$_{\text{D}}$ $\downarrow$
& FID$_{\text{PC}}$ $\downarrow$
& CD $\downarrow$
& FID$_{\text{V}}$ $\downarrow$
& CD$_{\text{V}}$ $\downarrow$\\
\midrule
  w/o. Aug
& 0.0126 & 88.85 & 104.64 & 4.64
& \textbf{0.0024}
& 3.69 &\textbf{0.0085}&
0.0114 &87.66 & \textbf{84.76} & 2.43 
& 0.0029 
&1.87&0.0107\\
 w. Aug
& \textbf{0.0124} & \textbf{88.89} & \textbf{102.83} & \textbf{4.50} 
& \textbf{0.0024} 
&\textbf{3.60}&\textbf{0.0085}
& \textbf{0.0111} &\textbf{87.73} & 84.99 & \textbf{2.35} 
& \textbf{0.0028} 
&\textbf{1.85}&\textbf{0.0103}\\

\bottomrule
\end{tabular}}
\end{table*}

\begin{figure}[tbp]
    \centering
    \includegraphics[width=\linewidth]{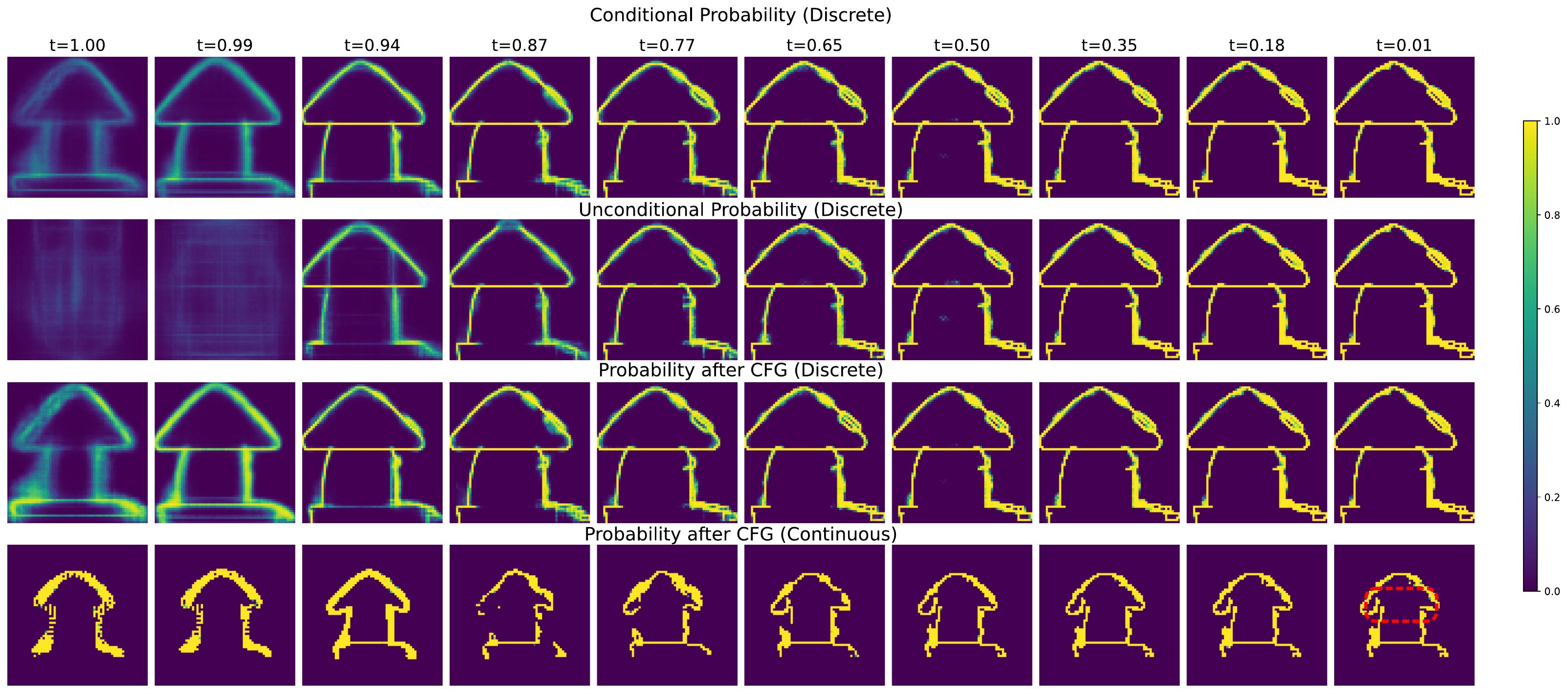} 
    \caption{ $\x_0$ prediction at different timesteps of one sampling process. A disconnected hole is marked in a red rectangle. Condition image from Figure \ref{fig:voxeluncertainty}.}
    \label{fig:ValueAfterCFG}
\end{figure}

To understand the generation dynamics, we visualize the $\x_0$ prediction for both continuous methods (TRELLIS) and discrete methods (ours) in Figure~\ref{fig:ValueAfterCFG}. For TRELLIS, the visualized values are first passed to the stage-1 VAE, and then a sigmoid function; for ours, we visualize the prediction of $\x_0$ in conditional and unconditional probability directly.

Figure \ref{fig:ValueAfterCFG} demonstrates that our model establishes global geometry early in the reverse process; by $t=0.87$, both conditional and unconditional trajectories have converged on a stable structural scaffold. Subsequent sampling introduces high-frequency details as predictive certainty increases, exhibiting a clear coarse-to-fine progression. Conversely, continuous dynamics (Figure \ref{fig:ValueAfterCFG}) appear less stable, with structures intermittently appearing and vanishing during intermediate steps, lacking the intuitive convergence of our discrete approach.

During sampling, the model becomes confident in the majority of locations, while a subset of voxels remains ambiguous with similar patterns for conditional and unconditional predictions. We interpret this ambiguity as a resolution effect: when occupancy and emptiness are both plausible at the voxel scale, small geometric variations near the discretization limit (on the order of $1/N$) can flip a voxel’s state without materially changing the perceived shape. This motivates us to use a near-zero uncertainty statistic $\gamma_{\epsilon,\rho}$  as a lightweight uncertainty score as a proxy to assess the level of fine details. Unless otherwise stated, we set $\epsilon=10^{-3}$ and $\rho=0.4$ in all experiments.

\begin{figure}[tbp]
    \centering
    \includegraphics[width=0.15\linewidth]{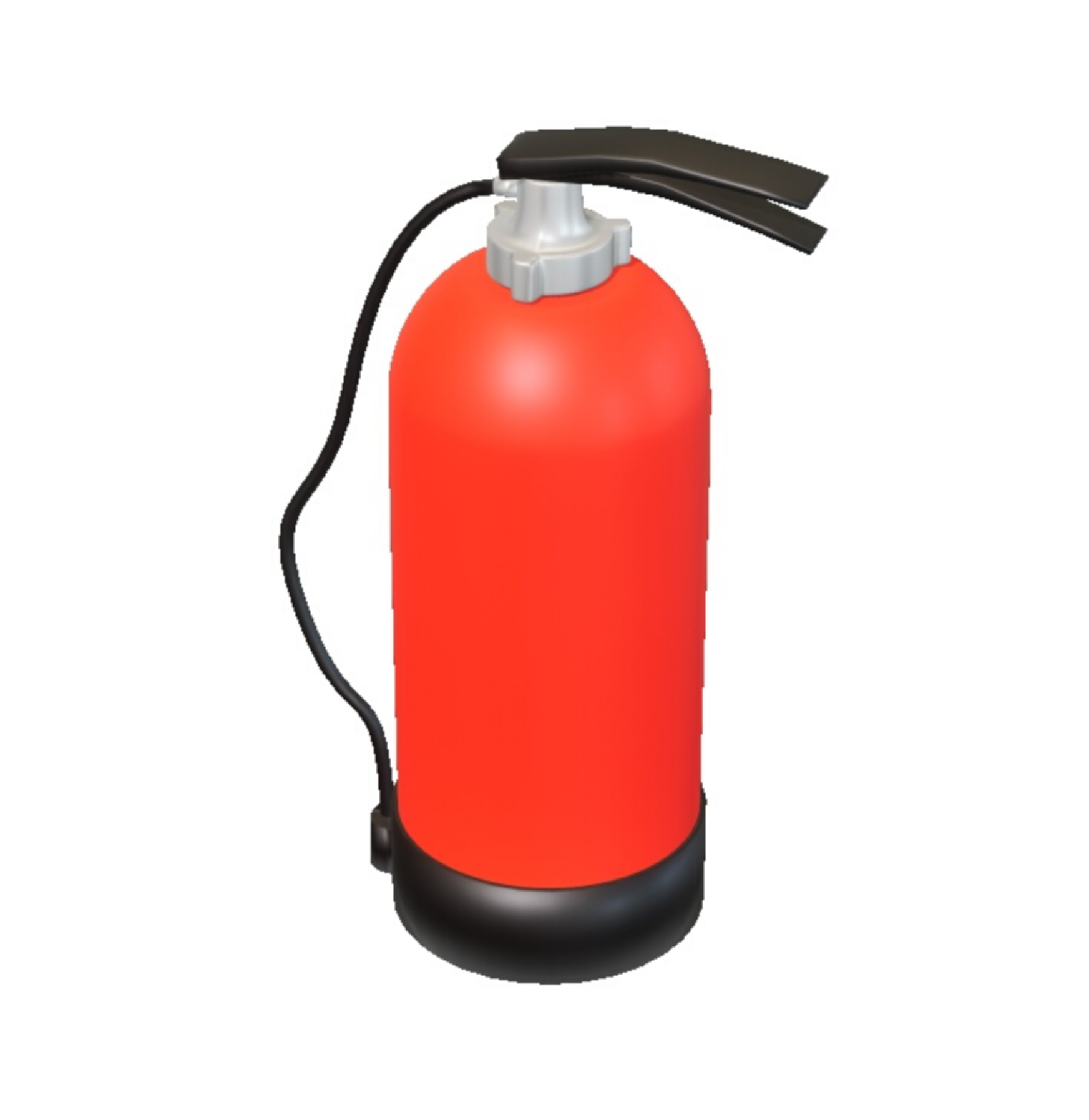}
    \includegraphics[width=0.165\linewidth]{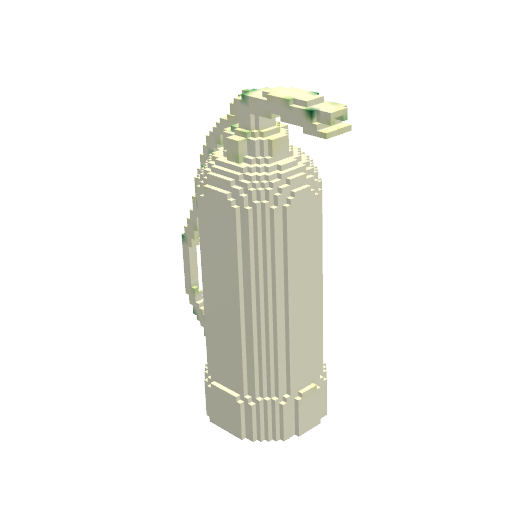}
    \includegraphics[width=0.165\linewidth]{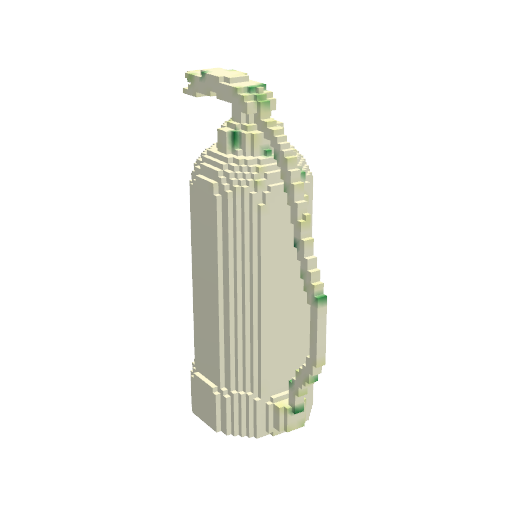}
    \includegraphics[width=0.15\linewidth]{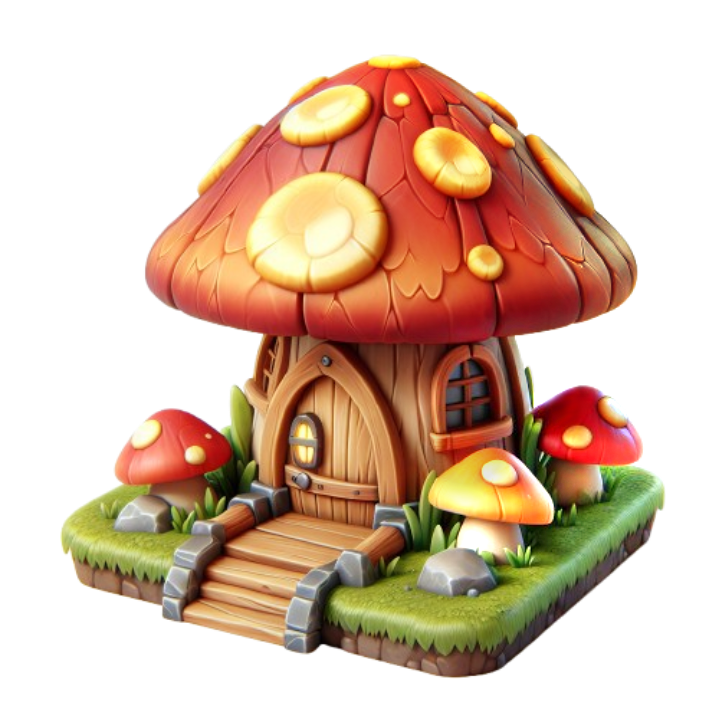}
    \includegraphics[width=0.165\linewidth]{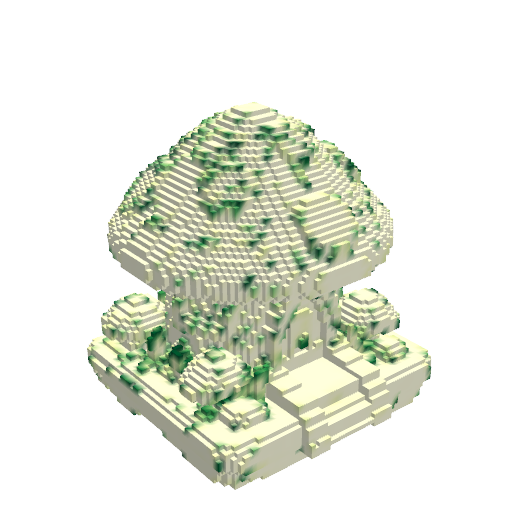}
    \includegraphics[width=0.165\linewidth]{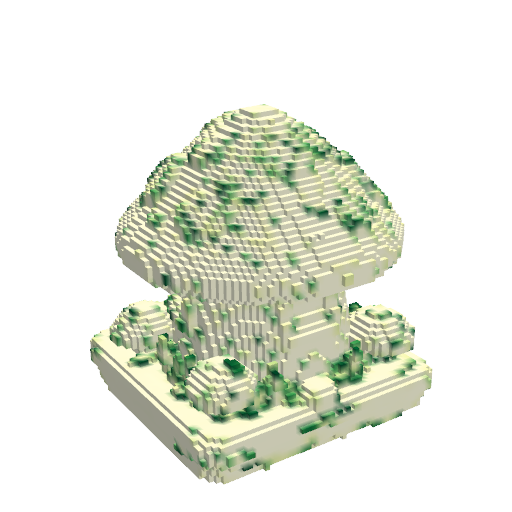}
    \caption{Visualizations of entropy on generated voxel grid. Prompts include a fire extinguisher and a mushroom house (from TRELLIS). The voxels are colored with entropy calculated from unconditional predictive probability. Darker shades denote voxels with higher entropy.}
    \label{fig:voxeluncertainty}
\end{figure}

We present two voxel visualizations as examples, one relatively complex and one relatively simple, in Figure \ref{fig:voxeluncertainty},  
which shows that voxels with higher entropy appear in areas with more detailed structures. For the fire extinguisher, the uncertain areas concentrate around the handle and the tube. For the mushroom house, the uncertain areas concentrate around regions such as grasses, stones, the bumps on the roof, and the small mushrooms, etc. The uncertainty score $\gamma$ is calculated for every voxelized mesh in the TRELLIS500K dataset, reported in Appendix~\ref{appendix:quantativeResults}.

To demonstrate its practical utility, we perform a data augmentation by oversampling the top 10K meshes ranked by their $\gamma$ scores, doubling their frequency within the training set. We then fine-tune our image-conditioned model for 100K steps from 300K, resulting in a total training step of 400K. We evaluated the likelihood of the training subset and Toys4k, as well as the generation results in Table \ref{tab:LikelihoodAfterextend}, which shows that the data augmentation efficiently strengthens the generation results.

\subsection{Editing After Finetuning}\label{sec:inpaintAfterFinetune}
For editing, we finetune both the image and text conditioned model, replacing half of the noise pattern with the BSP pattern (Section \ref{sec:multi_scale}) for 100k steps. During sampling, the CFG is set to 0.45 across all $t$, and we use a sampling step of 128 for all inpainting algorithms.

\begin{figure}[tbp]
    \centering
    \includegraphics[width=\linewidth]{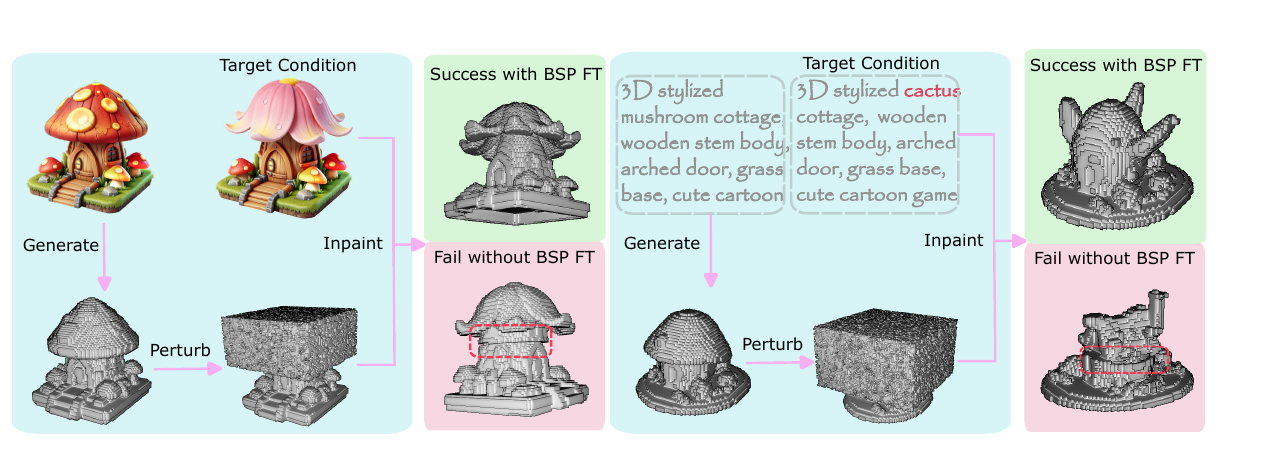}
    \caption{Mesh Editing with Image Condition. Target condition edited by Google Gemini Pro. }
    \label{fig:image_editing}
\end{figure}

The proposed BSP strategy enables DVD to perform seamless editing and inpainting within a single sampling trajectory while preserving the structural integrity of unmasked regions (Figure \ref{fig:image_editing}). In this workflow, the target edit region is initialized with uniform noise, followed by the sampling procedure detailed in Algorithm~\ref{alg:udlm_ancestral_inpainting}. Crucially, models without BSP fine-tuning fail to maintain spatial consistency, yielding fragmented or misaligned structures (highlighted by red dashed rectangles). Conversely, the fine-tuned model leverages the BSP inductive bias to generate coherent, boundary-aligned geometry that integrates naturally with the existing voxels.

To quantitatively evaluate the efficacy of image-conditioned inpainting and editing, we partition the voxel grid into two halves. By masking a half-space volume, we task both our discrete model and the continuous baseline (TRELLIS) with reconstructing the missing geometry conditioned on the remaining observed voxels. We report the results of our method before and after fine-tuning, and the results of TRELLIS with Repaint~\cite{lugmayr2022repaint} model in Table \ref{tab:metric_withwoft}. More experimental details are included in Appendix \ref{appendix:quantativeResults}.
\begin{table}[tbp]
\centering
\small
\caption{Image-conditioned inpainting result on training subset, NFE = 128. Baselines are colored in gray.}
\label{tab:metric_withwoft}
\setlength{\tabcolsep}{3pt}
\begin{tabular}{lcccccc}
\toprule
Method 
& CLIP $\uparrow$
& FID$_{\text{D}}$ $\downarrow$
& FID$_{\text{PC}}$ $\downarrow$
& CD $\downarrow$
& FID$_{\text{V}}$ $\downarrow$
& CD$_{\text{V}}$ $\downarrow$\\
\midrule

\textcolor{lightgray}{Ours baseline}&\textcolor{lightgray}{87.66} & \textcolor{lightgray}{84.76} & \textcolor{lightgray}{2.43} & \textcolor{lightgray}{0.0029} & \textcolor{lightgray}{1.87} &\textcolor{lightgray}{0.0107}
\\

Ours w/o. BSP
& 87.63 & 85.75 & 2.37 & 0.0031&1.89&0.0123\\

Ours w. BSP 
& \textbf{87.70} & \textbf{84.94} & \textbf{2.32} & \textbf{0.0028} &\textbf{1.82} &\textbf{0.0108}\\

\midrule
\textcolor{lightgray}{Cont. (TRELLIS) baseline}
& \textcolor{lightgray}{87.52} & \textcolor{lightgray}{87.53} & \textcolor{lightgray}{2.60} & \textcolor{lightgray}{0.0048} &\textcolor{lightgray}{1.99} & \textcolor{lightgray}{0.0199}\\
Cont. (TRELLIS) w. Repaint 
& 87.25 & 88.28 & 2.66 & 0.0058&2.49&0.0243\\
\bottomrule
\end{tabular}
\end{table}
As shown in Table \ref{tab:metric_withwoft}, our fine-tuned model completes the 3D model perfectly and even surpasses the baseline on several metrics. For models without fine-tuning and the continuous with Repaint suffer from a significant degradation in performance.

\section{Conclusion}\label{sec:conclusion}
We presented DVD for sparse voxel generation as the first stage of \textsc{SLat} 3D pipelines. Modeling occupancy directly in a discrete space avoids artifacts from continuous-to-discrete thresholding. Across image-to-3D and text-to-3D, our approach improves voxel fidelity and geometry-oriented metrics over the continuous baseline while using a smaller model and less training data. We also proposed an entropy-based score that correlates with local geometric detail, which potentially supports various applications. Additionally, we introduced a block-structured perturbation fine-tuning strategy to better match localized corruptions in editing and inpainting. Overall, our results suggest discrete diffusion is a practical first-stage prior for voxel scaffolds. However, DVD still requires a few hundred sampling steps for generation from scratch, and effective training and sampling strategies remain underexplored. Promising directions for future work include accelerating sampling for broader practical use and advancing 3D understanding through multimodal modeling of text and voxels.

\paragraph{Broader impacts.} DVD may support more efficient creation and manipulation of 3D assets for applications such as design, simulation, robotics, embodied AI, and immersive media. The proposed uncertainty estimates may also help identify ambiguous or difficult geometric regions, providing a lightweight tool for data filtering and quality assessment. At the same time, improved 3D generation tools could be misused to create unauthorized digital replicas, synthetic assets for deception, or copyrighted content. We therefore encourage responsible deployment for generated 3D content.
\clearpage

\bibliographystyle{unsrtnat}
\bibliography{example_paper}

\newpage
\clearpage
\appendix
\begin{center} {\huge \bf Appendix} \end{center}
\begingroup
\etocsettocstyle{\subsection*{Contents}}{}
\etocsetnexttocdepth{subsection}
\etocsettagdepth{main}{none}
\etocsettagdepth{appendix}{subsection}
\tableofcontents
\endgroup

\etocdepthtag.toc{appendix}

\section{More Related Works}
\paragraph{Quality of 3D models.}
With the development of 3D generative models, the quality of 3D assets has been given more emphasis. \cite{chen2025dora} evaluates the complexity of 3D mesh based on the number of salient edges. \cite{xiang2025structured3dlatentsscalable} leverages a pretrained image aesthetic model\footnote{https://github.com/christophschuhmann/improved-aesthetic-predictor} to evaluate the rendering of 3D assets. \cite{lin2025objaverse++} introduces Objaverse++, which manually annotates 10K Objaverse~\cite{deitke2023objaverse} assets with multi-view quality/aesthetic scores and attribute tags, then trains a predictor to scale these annotations to a curated subset of roughly 500K objects. They further show that filtering training data by these quality labels can improve image-to-3D generation performance and speed up training convergence.

\section{ELBO of Discrete Diffusion Models}\label{appendix:ELBODM}
For USDMs, $f$ from equation \ref{equ:elbo} is usually written as \cite{schiff2025simpleguidancemechanismsdiscrete} :
\begin{equation}
    \tilde f(\x_t,\x_\theta(\x_t,t),\alpha_t;\x_0) = -\frac{\alpha_t'}{K\alpha_t}
    \left[ \frac{K}{\bar\x_i} - \frac{K}{(\bar\x_\theta)_i} - \sum_j \frac{\bar\x_j}{\bar\x_i}\log\frac{(\bar\x_\theta)_i\cdot\bar\x_j}{(\bar\x_\theta)_j\cdot \bar\x_i} \right],
\end{equation} $\bar\x = K\alpha_t\x_0 + (1-\alpha_t)\mathbf{1},\;\bar\x_\theta = K\alpha_t\x_\theta(\x_t,t) + (1-\alpha_t)\mathbf{1}$. $\alpha_t'$ represents the time derivative of $\alpha_t$, $i=\arg\max_{j\in{K}}(\x_t)_j$ as the non-zero entry of $\x_t$.

Here we adopted its Rao Blackwellized Version as used in \cite{sahoo2025diffusionduality}:
\begin{align}
    f(\x_t,\x_\theta(\x_t,t),\alpha_t;\x_0) =& -\frac{\alpha_t'}{K\alpha_t}\Bigl[ \frac{K}{\bar\x_i} - \frac{K}{(\bar\x_\theta)_i} - (\kappa_t\mathbbm{1}_{\x_t=\x_0} + \mathbbm{1}_{\x_t\neq \x_0)})\sum_j\log\frac{(\bar\x_\theta)_i}{(\bar\x_\theta)_j} - \\&K\frac{\alpha_t}{1-\alpha_t}\log\frac{(\bar\x_\theta)_i}{(\bar\x_\theta)_m}\mathbbm{1}_{\x_t\neq \x_0}  -
    \left( (K-1)\kappa_t\mathbbm{1}_{\x_t=\x_0} - \frac{1}{\kappa_t}\mathbbm{1}_{\x_t\neq \x_0} \right)\log\kappa_t \Bigr],
\end{align}where $\kappa_t = (1-\alpha_t)/(K\alpha_t + 1-\alpha_t)$ and $m$ denote the index where $\x_m=1$. Emperically, this works better than cross entropy.

\section{Prior of Discrete Diffusion Models}\label{appendix:PriorDiffusionModels}

As mentioned in Sec. \ref{sec:DiscreteDiffusionModels}, the prior distribution of USDMs is $\bpi=\mathbf{1}/K$, which means an equal probability of all states. For MDMs, the prior distribution $\bpi=\m$, which is a degenerated distribution with probability 1 on the special MASK token. Despite the difference in prior distribution, these two frameworks demonstrate different behaviors during the diffusion process. In USDMs, each categorical latent can either stay the same or transition uniformly to any other category in 
$\mathcal{V}$, with the transition probabilities controlled by the diffusion timestep. In contrast, MDMs allow a token to either remain as a mask or to be revealed to other tokens, and once revealed, it stays masked throughout the reverse process. In other words, if MDMs make a mistake, there is no chance to fix them.

\begin{figure}[htbp]
    \centering
    \includegraphics[width=0.3\linewidth]{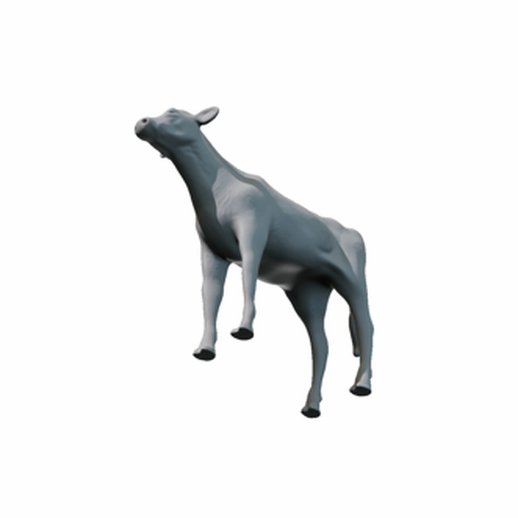}
    \includegraphics[width=0.3\linewidth]{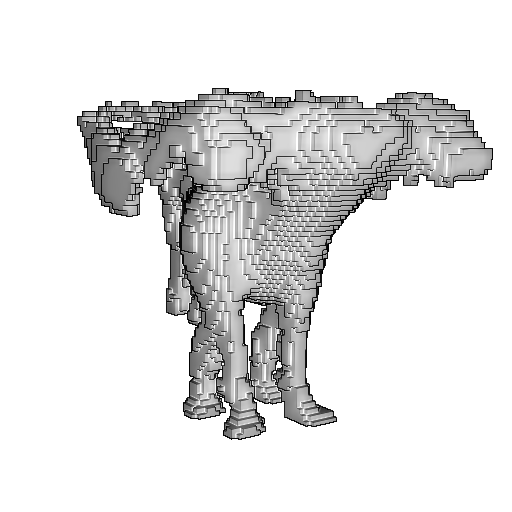}
    \includegraphics[width=0.3\linewidth]{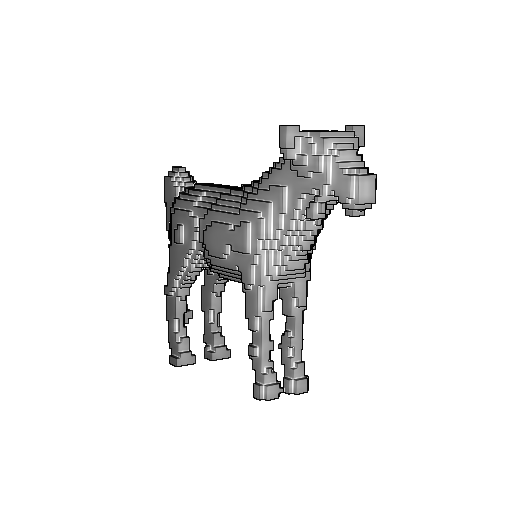}
    \caption{Example of the generation result of USDM and MDM at 60K steps of training, with ancestral sampler. CFG set to 2.0 and NFE=256. \textbf{Left}: Condition Image. \textbf{Middle}: MDM generated voxel. \textbf{Right}: USDM generated voxel.}
    \label{fig:USDMvsMDM}
\end{figure}

During our early experiment, we found that USDM outperforms MDM even in an early stage, where USDM starts to generate image-aligned voxels, while MDM still struggles to generate meaningful objects, as shown in Figure \ref{fig:USDMvsMDM}; thus, we decided to use USDMs for this task.

The main reason for the degraded performance of MDMs, we suspect, is the sequence length. In our work, we use voxel grids with a resolution of 64, yielding sequences of $L\approx$200K
tokens. This creates significant difficulty for MDMs: during each sampling step, thousands of tokens may be revealed at once, which leads to a high likelihood of early-stage errors.  From Figure \ref{fig:ValueAfterCFG}, it can be seen that in early sampling stages, the model is less certain about the occupancy of voxels, which further amplifies the chances of making early mistakes. These errors then accumulate through the reverse process and ultimately degrade generation quality. We tried to leverage several inference time scaling techniques for correcting MDMs, such as \cite{wang2025remaskingdiscretediffusionmodels, kimtrain,yang2025powerfulwaysgenerateautoregression}. While these methods demonstrate improvements compared to pure MDM settings, USDM still presents better sample quality without introducing additional computations.
The intrinsic self-correcting behavior of USDMs makes them better suited for handling such long-sequence but simple states settings.

\section{Design Space of Discrete Diffusion Models}\label{appendix:DesignSpace}
Due to the lack of a general voxel perceptual model, it can be difficult and expensive to evaluate the performance of a voxel generation model. Thus, we leverage a simpler surrogate dataset, MNIST \cite{mnist}, to provide partial information for different settings as a reference. For all the tests, we discretize the MNIST images into 8 categories, according to their pixel values. We train a tiny DiT (around 6M parameters) on the datasets for 50k steps under different settings, and we evaluate its FIDs over 10k generated images and likelihood on the validation set. The images are generated with 256 denoising steps. With the MNIST result as a reference, we also conduct several ablation studies on a confidential inner dataset (INNER3D for short). For evaluation, we select 20K meshes from INNER3D randomly as the validation set. We train a small DiT model ($\sim$140M params) for 140K steps with AdamW optimizer with no weight decay and $lr=1e-4$. Evaluate the negative log likelihood under the EMA 0.9999 weights for experiments with the INNER3D dataset. The exploration of design spaces is done with image-conditioned generation tasks.

\subsection{Sampling Distribution of $t$}

\begin{figure}[ht]
    \centering
    \begin{minipage}{0.48\linewidth}
        \centering
        \includegraphics[width=\linewidth]{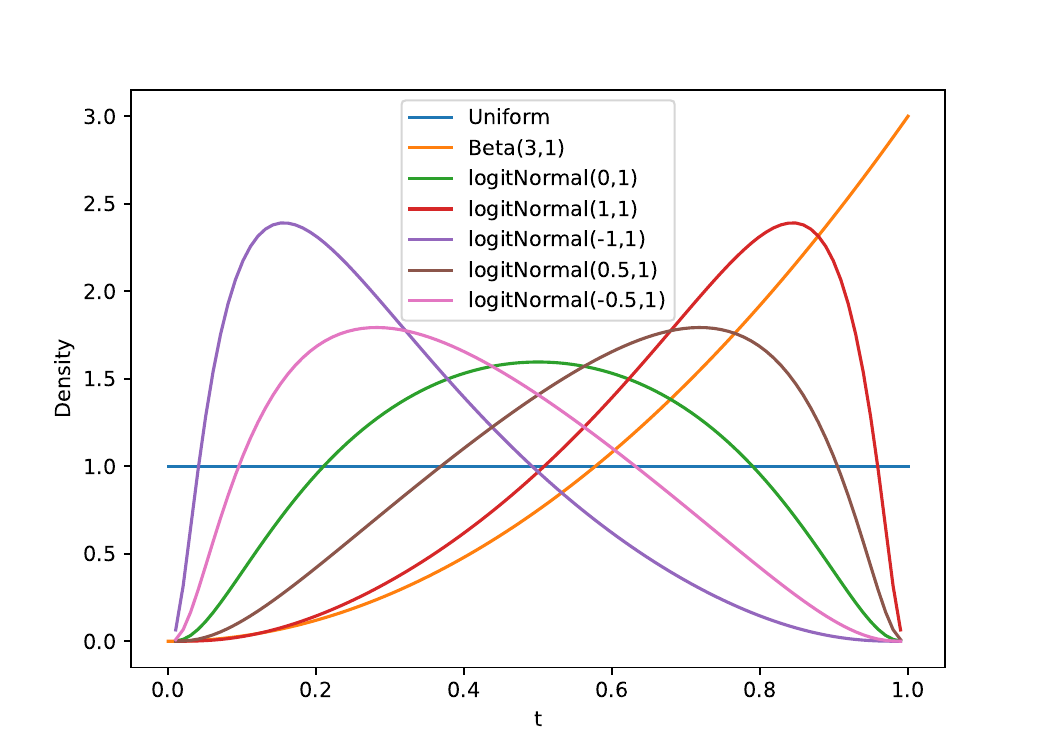}
        \caption{PDF of different distributions.}
        \label{fig:pdfoft}
    \end{minipage}
    \hfill
    \begin{minipage}{0.48\linewidth}
        \centering
        \captionsetup{type=table}
        \caption{NLL and FID results on discretized MNIST.}
        \label{tab:tdis_mnist}
        \begin{tabular}{lcc}
        \toprule
             Dist.& NLL $\downarrow$& FID $\downarrow$ \\\midrule
             Uniform& 0.2816 & 11.35\\
             LogitNormal(-1,1) & 0.2880&6.56\\
             LogitNormal(-0.5,1) & 0.2795& 4.89\\
             LogitNormal(0,1) & 0.2764&3.34\\
             LogitNormal(0.5,1) & 0.2781&3.50\\
             LogitNormal(1,1) & 0.2787&3.94\\
             Beta(3,1)& 0.2784&3.73 \\ \bottomrule
        \end{tabular}
    \end{minipage}
\end{figure}

\paragraph{Training.} In the ELBO objective \ref{equ:elbo}, the $t$ is sampled from $\mathcal{U}[0,1]$. 
We tested distributions that focus on different time sampling intervals on discretized MNIST, including: uniform distribution, logit Normal distributions, and Beta distributions, as shown in Figure \ref{fig:pdfoft} and Table \ref{tab:tdis_mnist}.

Table~\ref{tab:tdis_mnist} suggests that more training steps around $t$ close to $1$ might potentially lead to better results.
For further investigation on the 3D task, we choose logit Normal $(1,1)$, logit Normal $(0,1)$, and Beta$(3,1)$ as our candidate distributions, with different focuses on time intervals. We evaluate the negative log-likelihood at 140K steps of training on the INNER3D dataset in Table \ref{tab:ExpSmallDit}.
\begin{table}[htbp]
    \centering
    \caption{The evaluation negative log-likelihood on INNER3D}
    \label{tab:ExpSmallDit}
    \begin{tabular}{lc}
    \toprule
        Dist. & NLL $\downarrow$ \\\midrule
        LogitNormal(0,1) & 0.01369 \\
        LogitNormal(1,1) & 0.01457 \\
        Beta(3,1)        & 0.01341 \\
        \bottomrule
    \end{tabular}
\end{table}  Although it suggests that Beta$(3,1)$ might be the best candidate distribution, we found that as training continues, both logit Normal $(0,1)$ and Beta$(3,1)$ exhibit unstable sampling behavior and degraded quality, posing extra difficulties in CFG tuning. Thus, we choose logit Normal $(1,1)$ for image generation tasks. For text-conditioned generation tasks, we found that the Beta$(3,1)$ reports a better metric and generation result, possibly due to text-conditioned generation being a more difficult task. So for text-conditioned models, we choose Beta$(3,1)$.

\subsection{Structural Design}\label{sec:structuralDesign}

\paragraph{Positional embedding.} While RoPE \cite{SU2024127063} has been a standard technique in language modeling, its relative attention nature makes it also a good fit in 3D generative modeling. We conduct an ablation study in both absolute positional embedding (APE) and RoPE. By default, ROPE is only performed on the self-attention part, but not on the cross-attention part. We think the absolute position is also an important feature in 3D generative modeling, especially for image conditioning; thus, we optionally add APE in the cross-attention part to make cross-attention have access to local information. For all the positional embeddings, we keep the t sampling distribution to logit-Normal$(1,1)$. Thus, we also choose RoPE with absolute positional embedding for our final version model.

\begin{table}[htbp]
    \centering
    \caption{Evaluated negative log-likelihood of the validation set from INNER3D.}
    \label{tab:ablationpositionalembedding}
    \begin{tabular}{lc} 
    \toprule
        Model & NLL $\downarrow$ \\\midrule
         APE& 0.01457\\
         RoPE& 0.01451\\
         APE+RoPE& \textbf{0.01448}\\
    \bottomrule
    \end{tabular}
\end{table}

\paragraph{Stabilized training.} During our experiments, we have observed undesirable training dynamics: the grad norm will stop decreasing at some point around (250K steps), and start to increase with occasional spikes. Keep training will, counterintuitively, produce lower negative log-likelihood but, subsequently, worse generation quality. 

As discussed in Sec.~\ref{sec:DiscreteDiffusionModels}, the model output 
$\tilde{\x}_\theta$ corresponds to the unnormalized log-probabilities 
(logits) of the discrete states. During training, as the model becomes increasingly 
confident, these logits may grow to extremely large magnitudes, leading to numerical 
instability and undesirable optimization behavior. For instance, we observed the sampling processes where a normalized log-probability reached $-33$, corresponding to a 
probability of approximately $10^{-15}$. Such over-sharpened logits can destabilize the training 
dynamics and ultimately degrade sampling quality.

To mitigate this issue, we introduce a centered and $\tanh$-truncated logit parameterization, specifically designed for binary (two-state) discrete diffusion models. Concretely, we first center the logits by subtracting their mean, thereby removing any uninformative global shift. We then apply a scaled $\tanh$ function with parameter $\tau$, which softly bounds the logit differences and prevents the model from producing excessively confident predictions. This modification stabilizes both training and sampling without changing the underlying discrete diffusion objective. 
\begin{algorithm}[H]
\caption{Centered and $\tanh$-Truncated Logits}
\label{alg:centered_tanh_logits}
\begin{algorithmic}[1]
\REQUIRE Raw logits from the model: $\mathbf{x}_{\text{raw}} \in \mathbb{R}^{2}$
\REQUIRE Saturation parameter $\tau > 0$
\vspace{0.25em}
\STATE 
$
\mathbf{x}_{\text{centered}}
\gets \mathbf{x}_{\text{raw}} 
- \operatorname{mean}(\mathbf{x}_{\text{raw}})
$

\STATE 
$
\mathbf{x}_{\text{saturated}}
\gets \tau \cdot \tanh\!\left( \frac{\mathbf{x}_{\text{centered}}}{\tau} \right)
$
\STATE $\text{logits} \gets \mathbf{x}_{\text{saturated}}$
\OUTPUT logits
\end{algorithmic}
\end{algorithm}
We choose the truncate parameter $\tau=5$, which approximately corresponds to a limit probability of $5e-5$. We found that this parametrization could significantly stabilize both training and sampling behavior.

\paragraph{Abandon time conditioning.} In MDMs, people usually drop the time-step conditioning in the model \cite{kimtrain,nie2025largelanguagediffusionmodels}, since the number of mask tokens can itself be an indicator of the current time-step. In continuous diffusion models, some works \cite{sun2025noiseconditioningnecessarydenoising,wang2025equilibriummatchinggenerativemodeling} also studied the necessity of time conditioning. 
In USDMs, the time step is also inherited in the chaoticity of the input, though less straightforward than MDMs. In experiments of MNIST, we found that removing the time-step conditioning, while slightly harming the NLL, demonstrates better sampling quality:
\begin{table}[htbp]
    \centering
    \caption{MNIST experiments with and without time conditioning.}
    \label{tab:MNISTTimeCondition}
    \begin{tabular}{lcc}
    \toprule
        Setting&NLL&FID\\ \midrule
        Uniform& 0.2816 & 11.35\\
        Uniform w/o $t$ & 0.2861 & 7.11\\
        logitNormal(1,1) &  0.2787&3.94\\
         logitNormal(1,1) w/o $t$& 0.2807&3.74 \\
         \bottomrule
    \end{tabular}
\end{table}

While we keep time conditioning in our generation model to ensure minimal variation from typical of USDMs, we drop it for our fine-tuning tasks in Section. \ref{sec:multi_scale} by setting the time step to $0$. This also simplifies the inpainting algorithm since we don't need to calculate the actual time $t$ every step.

\section{More Implementation Details}\label{appendix:MoreImplementationDetails}
\subsection{Training and Sampling}
In addition to the design choices mentioned above, we also incorporate a cosine learning rate decay schedule from HuggingFace transformers~\cite{wolf-etal-2020-transformers} with a warm-up step of 5000. The training and fine-tuning are conducted on 64 NVIDIA A800 GPUs. During sampling, we leverage a cosine sampling grid \cite{shi2025simplifiedgeneralizedmaskeddiffusion}.

\subsection{Fine-tuning} Even though we tried to keep the block structures in a fraction of a predefined $t_b$ (see Algorithm \ref{alg:block_masks}), the actual perturbed region still varies because of rounding errors and overlapping. We present the distribution of the occupancy ratio of the generated blocks and the background $N^3$ grid when $t_b\sim\mathcal{U}[0,1]$:
\begin{figure}[htbp]
    \centering
    \includegraphics[width=0.5\linewidth]{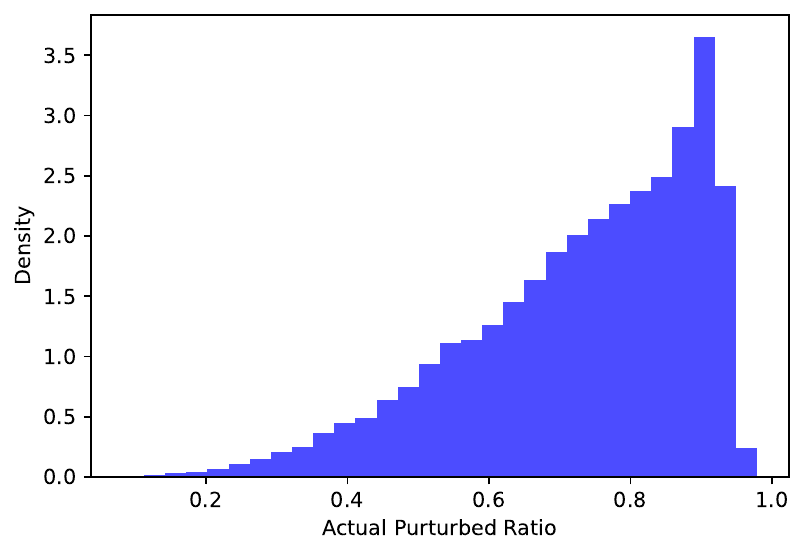}
    \caption{The Ratio of Generated Block Structure and Background Grid. The perturbation fraction $t_b\sim\mathcal{U}[0,1]$}
    \label{fig:blockRatio}
\end{figure}
During fine-tuning, we replace half of the perturbation strategy to BSP with uniformly distributed $t_b$. Within the blocked structures, we apply the forward diffusion kernel with $t\sim \rm{Beta}(3,1)$ to keep the focus on larger perturbation portions.

\subsection{Quantitative Results}\label{appendix:quantativeResults}
 \paragraph{Generation.} Instead of rendering a depth image and unprojecting, we directly normalize and sample 50k points from the surface for the generated mesh and the GT mesh, and compute the averaged Chamfer distance. By this, we evaluate the geometry quality of the whole mesh instead of only the visible part. For image conditioning, we select the image with a pitch angle of 30 degrees, a yaw angle of 20 degrees, and a radius of 2.5. For computing the metric for TRELLIS, we sample voxels with CFG strength as 3 and a sampling step of 50, being consistent with the original paper \cite{xiang2025structured3dlatentsscalable}. For the second stage, we stick with the default setting of TRELLIS's official repo: for the image conditioned task, the CFG strength is set to 5.0 for $t\in[0.5,1.0]$ and for the text conditioned task, the CFG is set to $7.5$ for $t\in[0.5,0.95]$.

Since we have observed that there are misalignment cases in GT meshes and generated meshes, possibly due to the diversity of alignment in the training set and differences in data processing conventions, we pre-align all generated voxels with the GT mesh before we pass the generated sparse voxels to the second stage. For TRELLIS and our method, we compute the chamfer distance between the GT voxels and generated voxels across 24 possible poses (6 oriented axes with 4 different up directions), then we rotate the voxels to the pose with the minimal chamfer distance. We found that this method aligns most of the meshes to the correct direction, which gives us more reliable geometric metrics. To conduct experiments on Direct3D-S2 \cite{wu2025direct3d} and TripoSR \cite{li2025triposghighfidelity3dshape}, we use their official implementations and released checkpoints (\href{https://github.com/DreamTechAI/Direct3D-S2}{Direct3D-S2}
, \href{https://github.com/VAST-AI-Research/TripoSR}{TripoSR}
). For Direct3D-S2, since one of the outputs is a latent index representing voxel coordinates used in the second-stage generation, we store it together with the final mesh. We observe that the generated meshes follow a canonical orientation aligned with one of the six principal axes; therefore, we align each mesh to the ground truth by selecting the axis permutation that minimizes the Chamfer distance. For TripoSR, the generated meshes also exhibit a consistent canonical orientation. We apply a sequence of rotations ($-150^\circ$ about the X-axis, followed by $-60^\circ$ about the Y-axis, and $65^\circ$ about the Z-axis) to align all meshes such that they face the positive z-axis, with the positive y-axis defined as the upward direction. We then apply our voxelization method on the aligned meshes to obtain corresponding voxel representations. All metrics are computed on the aligned meshes and their corresponding aligned voxel representations.

We present the detailed implementation of the evaluation pipeline.
\begin{itemize}
    \item [1.] The point cloud FID following PointNet++ \cite{qi2017pointnetdeephierarchicalfeature}. We sample 4000 points from each mesh and compute the P-FID, denoted as $\text{FID}_{\text{PC}}$. For the FID of Voxels, we sample 4000 points from the cubified mesh of the GT voxel and the generated voxels. For computing the chamfer (CD), we sample 10000 points from each mesh. For computing the chamfer distance of voxels ($\text{CD}_{\text{V}}$), we treat the center of the voxels as the sampled point clouds.
    \item [2.] The FID under DINOv2 \cite{oquab2024dinov2learningrobustvisual} feature spaces, where we render 6 images per asset with yaw angles at every 60 degrees, a pitch angle of 30 degrees, and a radius of 2.5. Totally, there are 15000 images in the training subset and 6000 images in the Toys4k evaluation set.
    \item [3.] We also compute the CLIP score of rendered results and the GT rendering image set. We render 6 images per generated asset with yaw angles at every 60 degrees, a pitch angle of 30 degrees, and a radius of 2.5, and calculate the maximal CLIP score across these 6 images per asset, then report the averaged CLIP score across all assets.
\end{itemize}

\begin{figure}[htbp]
    \centering
    \includegraphics[width=.5\linewidth]{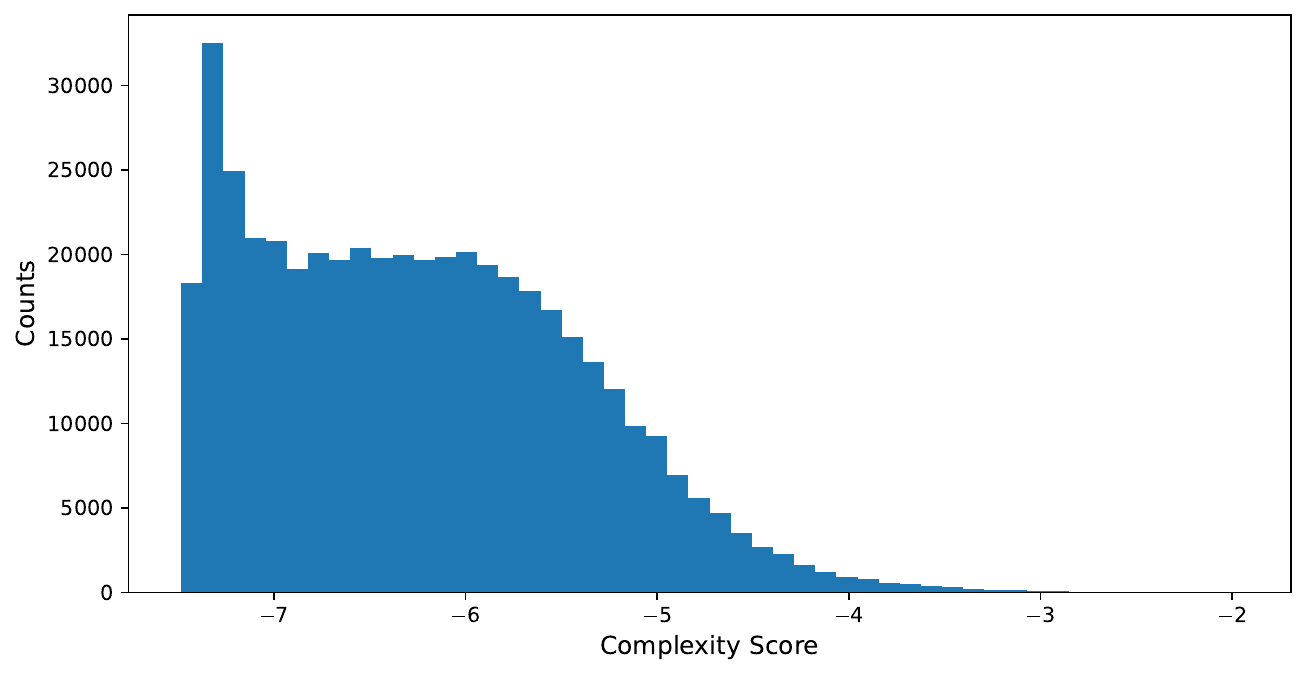}
    \caption{Complexity Scores $\gamma$ of TRELLIS500K Dataset}
    \label{fig:statistics}
\end{figure}

\begin{figure}[htbp]
\centering
\setlength{\tabcolsep}{2pt}      
\renewcommand{\arraystretch}{1.0}
\begin{tabular}{cccc}
\begin{minipage}{0.14\columnwidth}\centering
  \includegraphics[width=\linewidth]{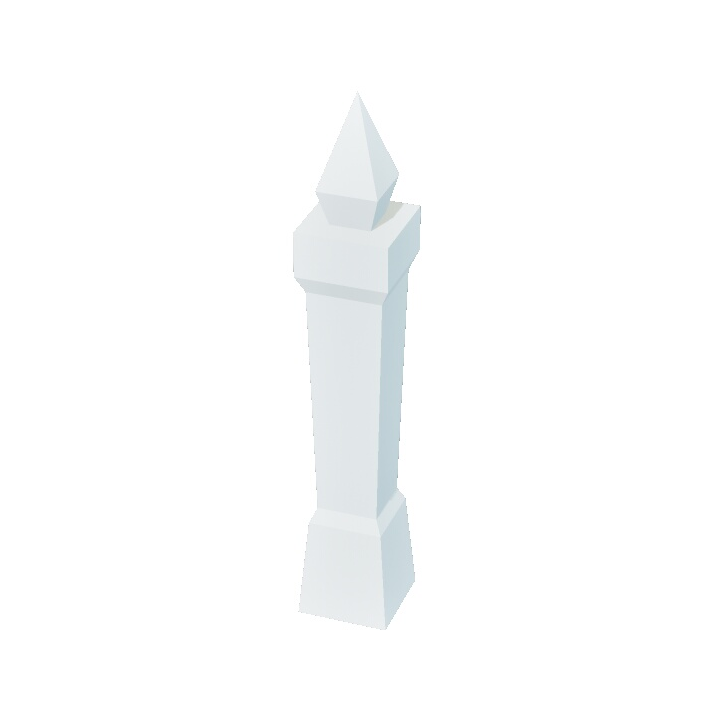}\par
  \vspace{1pt}
  \small\bfseries Score: -7.84
\end{minipage} 
\begin{minipage}{0.14\columnwidth}\centering
  \includegraphics[width=\linewidth]{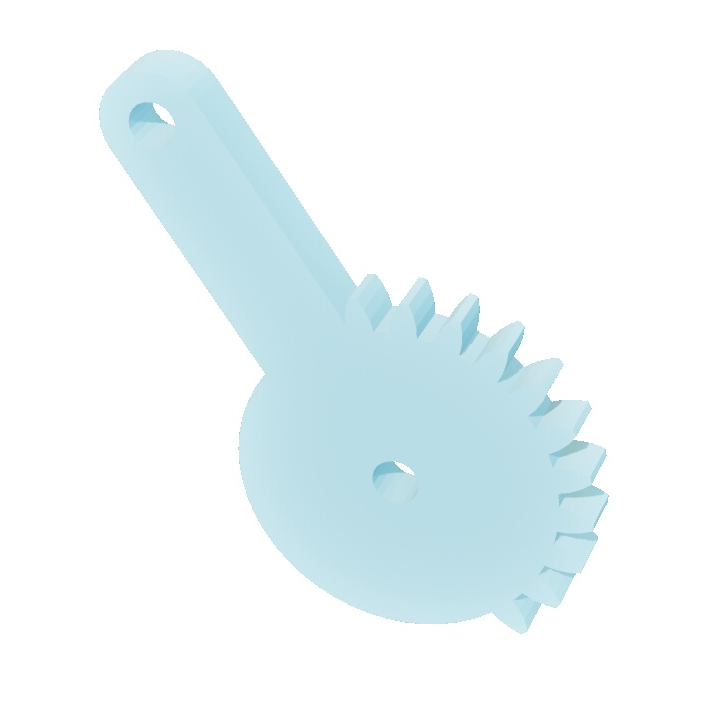}\par
  \vspace{1pt}
  \small\bfseries Score: -7.40
\end{minipage} 
\begin{minipage}{0.14\columnwidth}\centering
  \includegraphics[width=\linewidth]{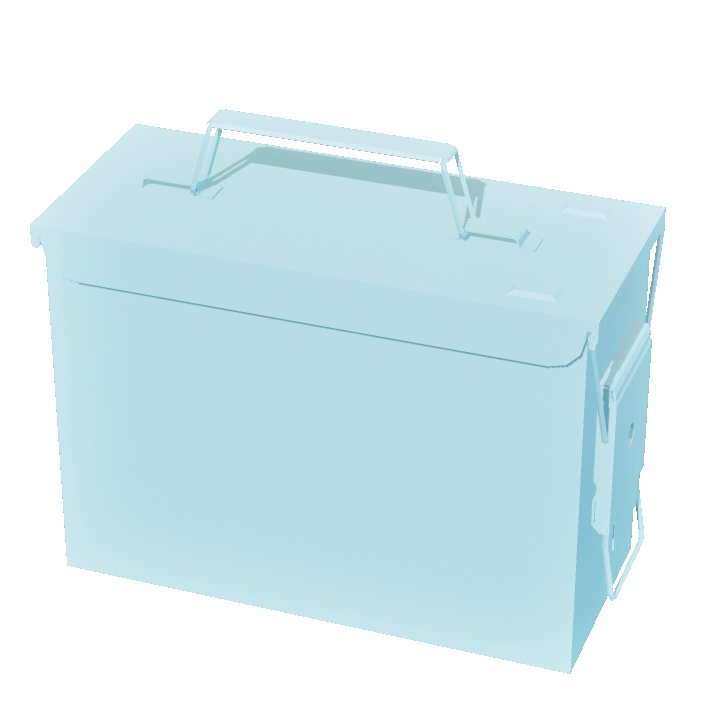}\par
  \vspace{1pt}
  \small\bfseries Score: -7.02
\end{minipage} 
\begin{minipage}{0.14\columnwidth}\centering
  \includegraphics[width=\linewidth]{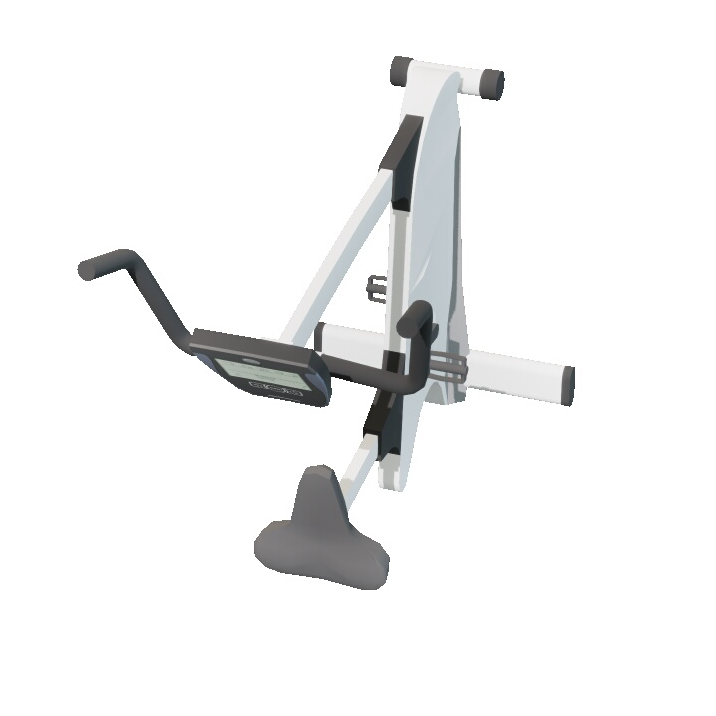}\par
  \vspace{1pt}
  \small\bfseries Score: -6.75
\end{minipage}

\begin{minipage}{0.14\columnwidth}\centering
  \includegraphics[width=\linewidth]{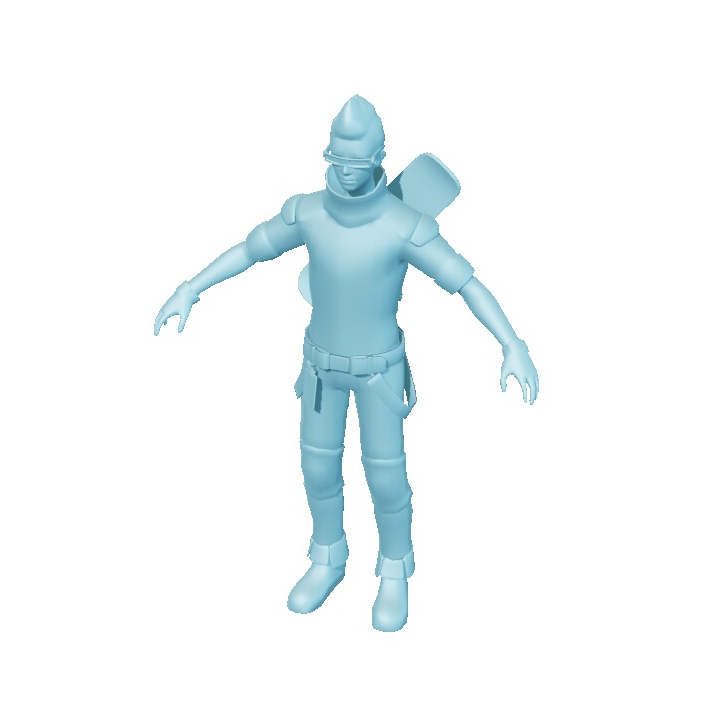}\par
  \vspace{1pt}
  \small\bfseries Score: -6.20
\end{minipage} 
\begin{minipage}{0.14\columnwidth}\centering
  \includegraphics[width=\linewidth]{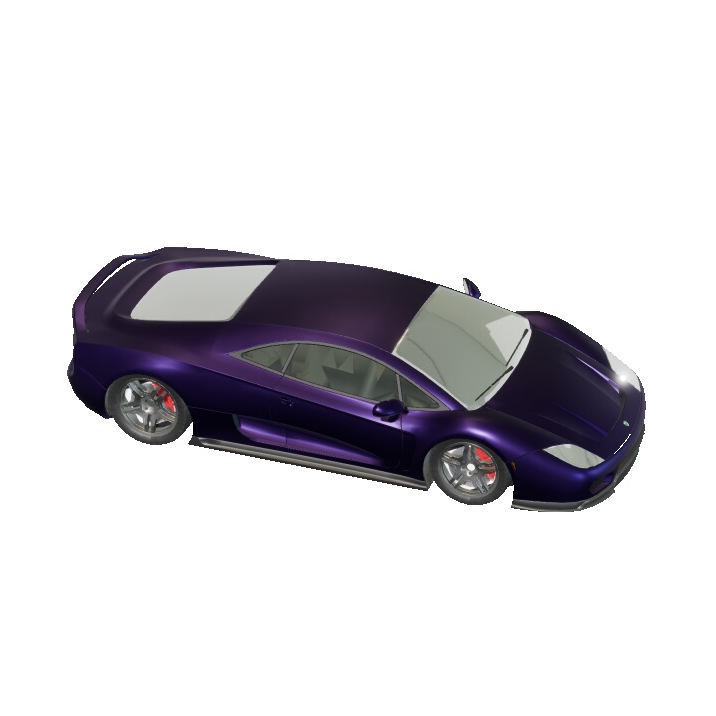}\par
  \vspace{1pt}
  \small\bfseries Score: -5.92
\end{minipage} 
\begin{minipage}{0.14\columnwidth}\centering
  \includegraphics[width=\linewidth]{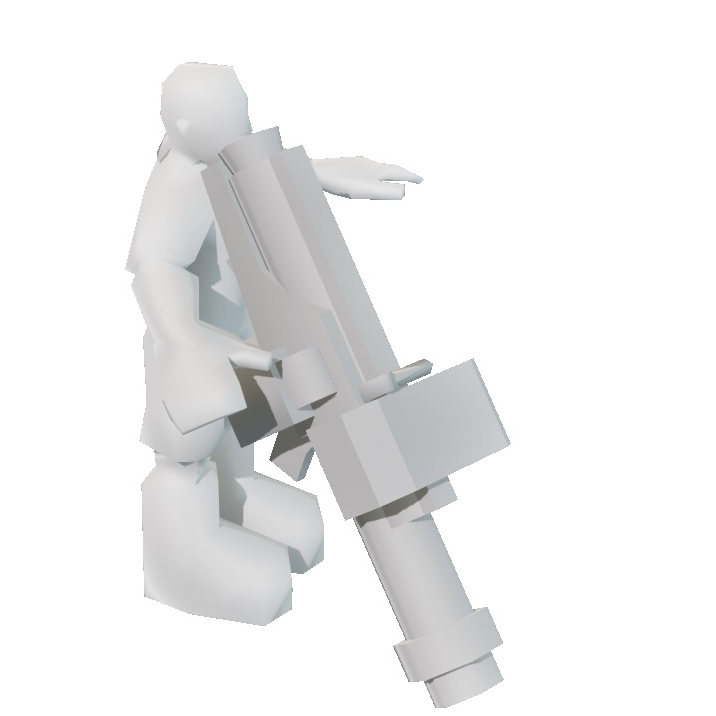}\par
  \vspace{1pt}
  \small\bfseries Score: -5.24
\end{minipage}
\end{tabular}
\vspace{-2pt}
\caption{Uncertainty scores $\gamma$ for selected meshes.}
\label{fig:complexity_scores}

\end{figure}

\paragraph{Uncertainty scores}
The distribution of uncertainty scores of the training set is given in Figure \ref{fig:statistics}, several example meshes with their $\gamma$ scores are shown in Figure \ref{fig:complexity_scores}.

\begin{figure}[htbp]
    \centering
    \includegraphics[width=0.45\linewidth]{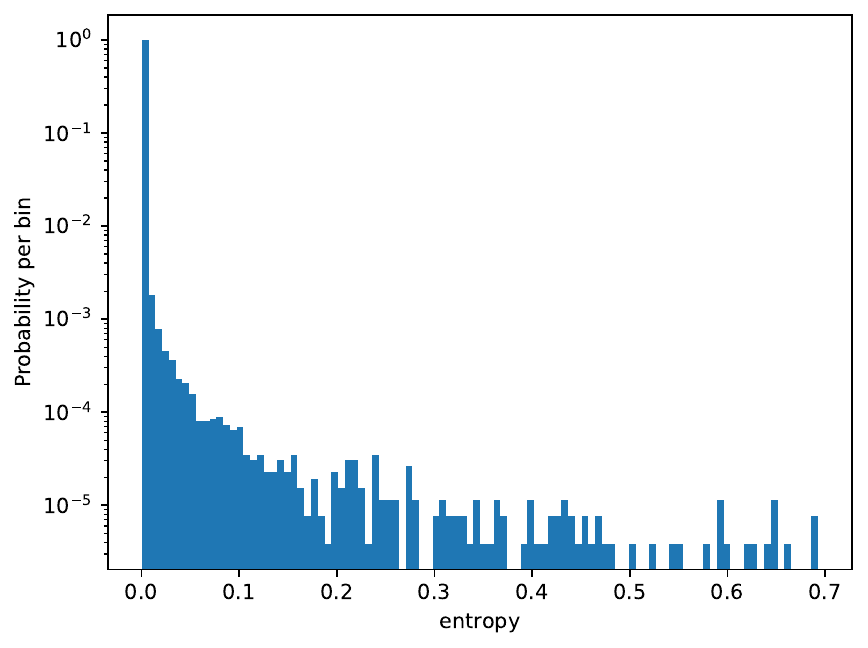}
    \includegraphics[width=0.45\linewidth]{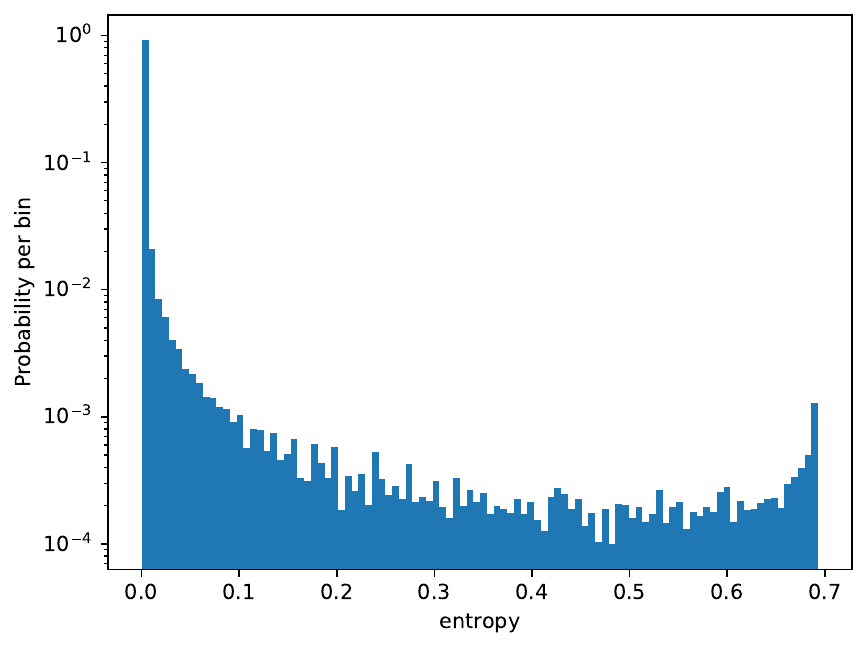}
    \caption{Histogram of entropy of the fire extinguisher (Left) and mushroom house (Right) presented in Figure \ref{fig:voxeluncertainty}}
    \label{fig:uncertaintyHist}
\end{figure}

We also present the histogram of the entropies of the generated voxels shown in Figure \ref{fig:uncertaintyHist}.

\paragraph{Inpainting.} For the inpainting experiment, we apply an axis-aligned half-space mask, setting all voxels to unknown on one side of the grid: for a chosen axis $a \in \{x,y,z\}$, we mask either indices $a \in [32,63]$ or $a \in [0,31]$ (six choices in total).
In algorithm \ref{alg:udlm_ancestral_inpainting}, we present the sampling method we used for sampling and inpainting. The additional steps needed for inpainting are colored blue. While calculating the results of Table \ref{tab:metric_withwoft} for the fine-tuned model, all additional steps, i.e. \cnum{1}\cnum{2} and \cnum{3}, are enabled. For the models that are not fine-tuned, we only enabled \cnum{2}; otherwise, the model generates nothing or a fully occupied voxel in the perturbed part, which makes it pointless to calculate the metric on such obvious failure cases. For the continuous method with Inpaint, we set the sampling steps to 64 and 1 additional re-sampling step within each sampling step, resulting in a total NFE of 128. We also discussed the reason for conducting the inpainting experiment with generated voxels (instead of GT) in Appendix~\ref{appendix:ablationAlignments}.

\section{Additional Ablation Study}\label{appendix:ablation}
\subsection{Ablation on Sampling Steps} 
We studied the influence of the sampling step for both continuous and discrete stage 1 on the Toys4k dataset.

\begin{table}[htbp]
\centering
\caption{Generation result on Toys4K with different NFEs.}
\label{tab:metric_ablation_image}
\begin{tabular}{lccccccc}
\toprule
\multirow{2}{*}{\raisebox{-0.5ex}{Method}}
& \multicolumn{4}{c}{Image-to-3D}
& \multicolumn{3}{c}{Text-to-3D} \\
\cmidrule(lr){2-5}
\cmidrule(lr){6-8}
& CLIP $\uparrow$
& FID$_{\text{DINOv2}}$ $\downarrow$
& FID$_{\text{PC}}$ $\downarrow$
& CD $\downarrow$
& CLIP $\uparrow$
& FID$_{\text{DINOv2}}$ $\downarrow$
& FID$_{\text{PC}}$ $\downarrow$ \\
\midrule
Ours (64)
& {88.78} & \underline{105.77} & 4.78 & \underline{0.0026}& \underline{26.10} & \underline{328.88} & \textbf{7.92}\\
Ours (256)
& \textbf{88.85} & \textbf{104.64} & 4.64 & \textbf{0.0024}& 26.05 & \textbf{312.21} & \underline{8.74}\\
Cont. (50)
& 87.13 & 106.71 & \textbf{4.51} & 0.0039& \underline{26.10} & 328.18 & 8.97\\
Cont. (256)
& 88.81 & 106.14 & \underline{4.56} & 0.0039& \textbf{26.13} & 325.33 & 9.16\\

\bottomrule
\end{tabular}
\end{table}

From Table \ref{tab:metric_ablation_image}, we can see that for the image conditioned generation task, both methods benefit from more sampling steps. Particularly, DVD suffers more from fewer NFEs. While for text-conditioned generation tasks, no clear gain is observed.

\subsection{Ablation on Alignments}\label{appendix:ablationAlignments}
As mentioned in Sec. \ref{appendix:quantativeResults}, we pre-aligned the generated voxel with the GT meshes before we pass the voxels into stage 2. One reason for this is due to differences in voxel orientation conventions and data preprocessing; the generated voxel of TRELLIS and our methods possess different canonical orientations, which leads to significantly worse results in TRELLIS's geometric metrics. We present the result in Table \ref{tab:metric_trainingset_noalign} and Table \ref{tab:metric_toys4k_noalign} as ablations of the alignment processing. Compared with Table \ref{tab:metric_all}, the alignment operation benefits geometric metrics of TRELLIS to a more convincing range instead of being counterintuitively worse.

This also stands for the reason of making use of the generated voxels for inpainting instead of using GT in experiment \ref{sec:inpaintAfterFinetune}, as they keep the internal canonical orientation that the model admits. Otherwise, the continuous method would fail in many cases, and the metric calculated is less convincing.

\begin{table*}[ht]
\centering
\caption{Quantitative comparison on training subset w/o pre-alignment.}
\label{tab:metric_trainingset_noalign}
\begin{tabular}{lccccccc}
\toprule
\multirow{2}{*}{\raisebox{-0.5ex}{Method}}
& \multicolumn{4}{c}{Image-to-3D}
& \multicolumn{3}{c}{Text-to-3D} \\
\cmidrule(lr){2-5}
\cmidrule(lr){6-8}
& CLIP $\uparrow$
& FID$_{\text{D}}$ $\downarrow$
& FID$_{\text{PC}}$ $\downarrow$
& CD $\downarrow$
& CLIP $\uparrow$
& FID$_{\text{D}}$ $\downarrow$
& FID$_{\text{PC}}$ $\downarrow$ \\
\midrule
Cont.(TRELLIS)$^\dagger$
& 87.65& 89.61 & 4.49 & 0.0135
& 26.10 & 238.13 & 6.56 \\

Ours
& \textbf{87.84} & \textbf{84.79} & \textbf{2.39} & \textbf{0.0040}
& \textbf{26.18} & \textbf{208.86} & \textbf{3.58} \\
\bottomrule
\end{tabular}
\end{table*}

\begin{table*}[ht]
\centering
\caption{Quantitative comparison on Toys4K w/o pre-alignment}
\label{tab:metric_toys4k_noalign}
\begin{tabular}{lccccccc}
\toprule
\multirow{2}{*}{\raisebox{-0.5ex}{Method}}
& \multicolumn{4}{c}{Image-to-3D}
& \multicolumn{3}{c}{Text-to-3D} \\
\cmidrule(lr){2-5}
\cmidrule(lr){6-8}
& CLIP $\uparrow$
& FID$_{\text{D}}$ $\downarrow$
& FID$_{\text{PC}}$ $\downarrow$
& CD $\downarrow$
& CLIP $\uparrow$
& FID$_{\text{D}}$ $\downarrow$
& FID$_{\text{PC}}$ $\downarrow$ \\
\midrule
Cont.(TRELLIS)$^\dagger$
& 88.81 & 108.91 & 6.11 & 0.0147
& \textbf{26.41} & 338.40 & 11.56\\
Ours
& \textbf{88.95} & \textbf{102.65} & \textbf{4.63} & \textbf{0.0025}
& \textbf{26.41} & \textbf{312.13} & \textbf{8.64} \\
\bottomrule
\end{tabular}

\vspace{2pt}
{\footnotesize
$^{\dagger}$ Metric reimplemented by us.
}
\end{table*}

\section{User Study}\label{sec:user_study}
\begin{figure}[h]
    \centering
    \includegraphics[width=0.7\linewidth]{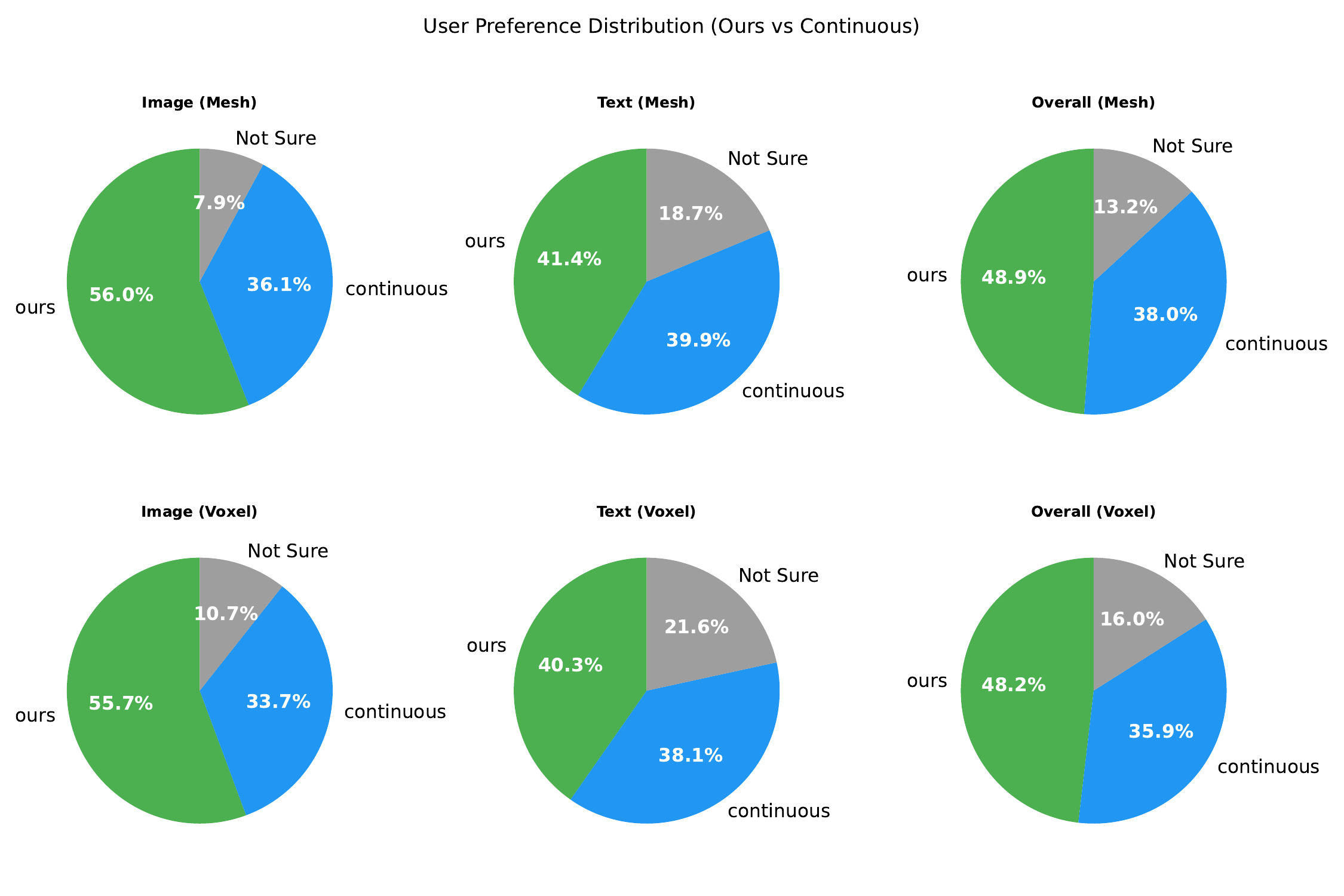}
    \caption{Pie graph of user preference.}
    \label{fig:user_study}
\end{figure}

\begin{figure}
    \centering
    \includegraphics[width=\linewidth]{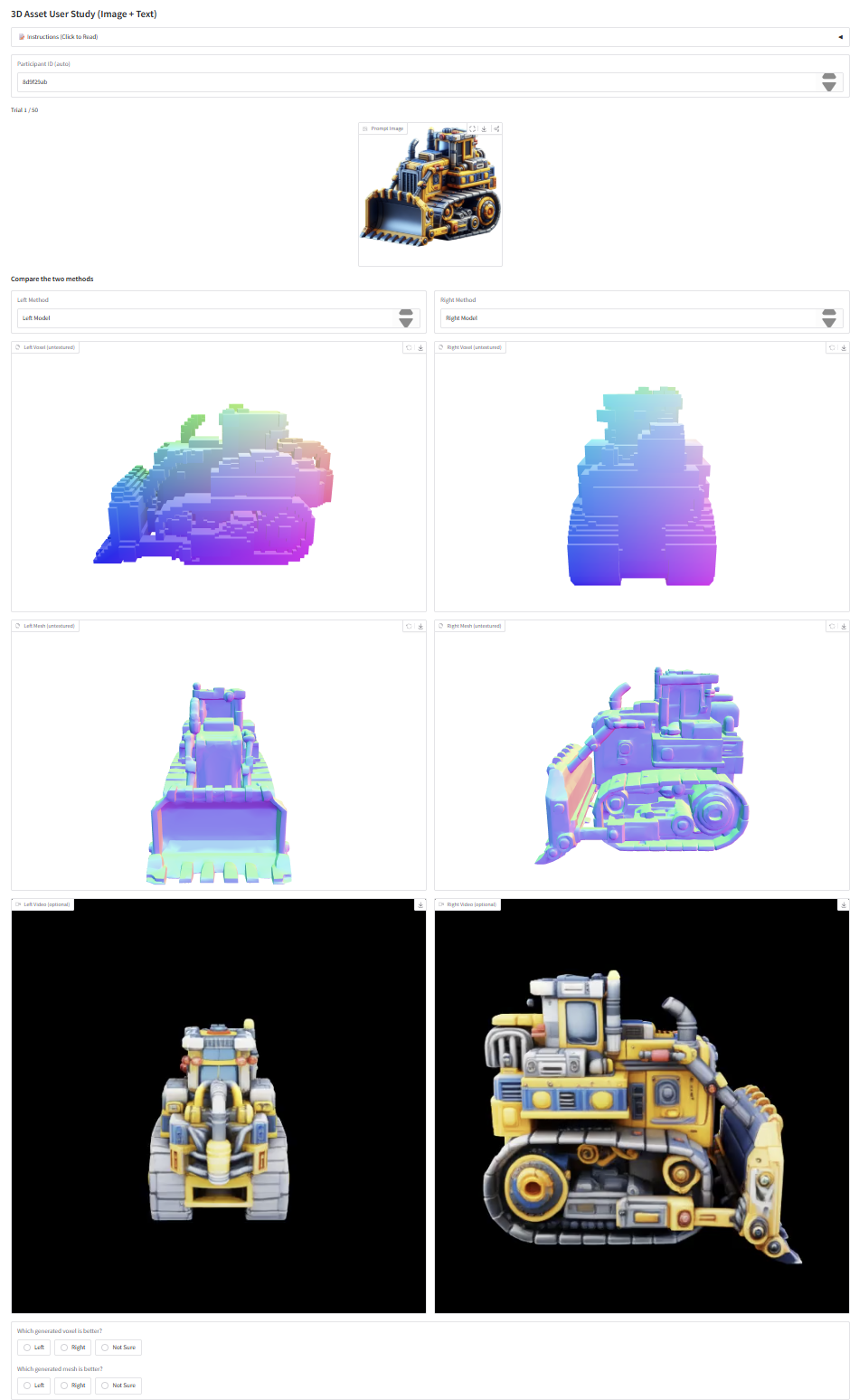}
    \caption{An example screenshot of the user study. The question at the bottom asks about the preferred mesh and voxel.}
    \label{fig:user_study_example}
\end{figure}

We also conducted a user study where we prepared the generation results of DVD and TRELLIS with different image and text prompts. In total, there are 25 image prompts and 25 text prompts. We presented interactive visualizations of the generated voxels, meshes, and rendered videos, and asked the users to choose their preferred voxels and meshes (see an example screenshot in Figure~\ref{fig:user_study_example}). We collect data with around 600 trials. The results are shown in Figure \ref{fig:user_study}. We can see that for image-based tasks, DVD is preferred among users, and for text-based tasks, both methods are on the same level.

\section{Perturbing and Inpainting Algorithm}\label{appendix:PurturbInpaint}

In Algorithm \ref{alg:block_masks}, we introduced the algorithm for generating a block pattern for augmented training. In algorithm \ref{alg:udlm_ancestral_inpainting}, we present the sampling and inpainting algorithm. The additional lines needed for inpainting are colored in blue; the sampling algorithm is modified from \cite{schiff2025simpleguidancemechanismsdiscrete}.

\begin{algorithm}[H]
\caption{Generation of Block-Structured Masks}
\label{alg:block_masks}
\begin{algorithmic}[1]
\REQUIRE Batch size $B$, spatial size $N$, dimension $\mathrm{dim} \in \{1,2,3\}$
, set of block side lengths $\mathcal{L} = \{\ell_1,\dots,\ell_{L_m}\}$,
 target masked fraction $t_b \in [0,1]$ for each sample $b \in \{1,\dots,B\}$
\vspace{0.25em}
\STATE Initialize binary masks $\mathbf{M}_b \gets \mathbf{0} \in \{0,1\}^{N^\mathrm{dim}}$ for all $b$.
\STATE Compute total volume $V \gets N^\mathrm{dim}$.
\FOR{each sample $b = 1,\dots,B$}
    \STATE \textbf{(Per-scale maximum counts)} For each length $\ell_j \in \mathcal{L}$, set
    \[
        A_j \gets \ell_j^{\mathrm{dim}}, \qquad
        m_{b,j}^{\max} \gets \left\lceil \frac{t_b \, V}{A_j} \right\rceil.
    \]
    \STATE \textbf{(Random allocation across scales)} Sample i.i.d.\ weights $u_{b,j} \sim \mathrm{Uniform}(0,1)$ and normalize
    \[
        w_{b,j} \gets \frac{u_{b,j}}{\sum_{k=1}^{L_m} u_{b,k}}.
    \]
    \STATE Compute the number of blocks at each scale
    \[
        m_{b,j} \gets \left\lceil w_{b,j} \, m_{b,j}^{\max} \right\rceil.
    \]
    \FOR{each length $\ell_j \in \mathcal{L}$}
        \FOR{$k = 1,\dots,m_{b,j}$}
            \STATE Sample a start position $\mathbf{s} \in \{0,\dots,N-1\}^\mathrm{dim}$ uniformly (with truncation at borders).
            \STATE Define the block
            \[
                \mathcal{B}(\mathbf{s}, \ell_j)
                = \big\{\mathbf{v} \in \{0,\dots,N-1\}^\mathrm{dim}
                   : s_d \le v_d < s_d + \ell_j \;\; \forall d \in \{1,\dots,\mathrm{dim}\}
                  \big\},
            \]
            where indices exceeding $N-1$ are clipped.
            \STATE Set $\mathbf{M}_b[\mathbf{v}] \gets 1$ for all $\mathbf{v} \in \mathcal{B}(\mathbf{s}, \ell_j)$ (union of blocks).
        \ENDFOR
    \ENDFOR
\ENDFOR
\STATE Optionally, compute the realized masked fraction for each sample
\[
    t_b \gets \frac{1}{V} \sum_{\mathbf{v}} \mathbf{M}_b[\mathbf{v}].
\]
\OUTPUT Block masks $\mathbf{M}_1,\dots,\mathbf{M}_B$ (and $t_1,\dots,t_B$ if desired).
\end{algorithmic}
\end{algorithm}

\begin{algorithm}[H]
\caption{USDM Ancestral Sampling with Optional Inpainting}
\label{alg:udlm_ancestral_inpainting}
\begin{algorithmic}[1]
\REQUIRE Initial input $\x^{1:L}$ (noise or corrupted input) at $t = 1$ (e.g., pure noise), number of steps $T$, optional inpainting parameters $\{\mathbf{M}^{1:L},\mathbf{C}^{1:L}\}$, where mask $\mathbf{M}^{1:L} \in \{0,1\}^{L}$ and target classes $\mathbf{C}^{1:L} \in \{0,\dots,K-1\}^{L}$
\vspace{0.25em}

\STATE Choose a decreasing time grid $1 = t_0 > t_1 > \dots > t_T \approx 0$, e.g.\ $t_k$ from a rescaled linear or cosine schedule.
\STATE Set $\x_{t_0}^{1:L} \gets \x^{1:L}$ initial noise (or corrupted input).

\FOR{$k = 0$ to $T-1$}
    \STATE $t \gets t_k,\quad s \gets t_{k+1}$.
    {\color{blue}
    \STATE \cnum{1}\textbf{(Calculate Actual Timestep)}
    \[
        \tilde{t}\gets  
        \begin{cases}
            t \cdot\frac{L-\text{sum}(\mathbf{M}^{1:L})}{L} & \text{if } \{\mathbf{M}^{1:L},\mathbf{C}^{1:L}\} \text{ is provided}\\
            t,& \text{otherwise}
        \end{cases}
    \]}
    \STATE \textbf{(Model prediction)} 
    \[
        \p^{1:L} = \x_\theta^{1:L}(\x_t^{1:L},\tilde t)
    \]
    {\color{blue}
    \STATE \cnum{2}\textbf{(Substituting Probabilities)}
    \IF{$\{\mathbf{M}^{1:L},\mathbf{C}^{1:L}\}$ is provided} 

    \STATE    \[
            \mathbf{p}^{i}\gets
            \begin{cases}
                \text{Onehot}(\mathbf{C}^i), & \text{if } \mathbf{M}^i = 1, \\
                \p^i, & \text{otherwise}.
            \end{cases} \quad\text{for } i\in\{1,...,L\}
        \]
    \ENDIF}
    \STATE \textbf{(Computing Posterior)} Compute $q_{s|t}^{1:L}(\cdot|\x_t^{1:L},\x^{1:L})$ according to equation \ref{equ:posterior}

    \IF{$k < T-1$}
        \STATE \textbf{(Ancestral step)} Sample
        \[
            \x_{s}^{1:L} \sim q_{s|t}^{1:L}
        \]
    \ELSE
        \STATE \textbf{(Final noise removal)} Set
        \[
            \x_{s}^{1:L} = \text{Onehot}(\arg\max q_{s|t}^{1:L})
        \]
        
    \ENDIF
    {\color{blue}
    \STATE \cnum{3}\textbf{(Replace with Known Categories)}
    \IF{$\{\mathbf{M}^{1:L},\mathbf{C}^{1:L}\}$ is provided}
        
    \STATE\[
            \x_s^{i} \gets
            \begin{cases}
                \text{Onehot}(\mathbf{C}), & \text{if } \mathbf{M}^i = 1, \\
                x_s^{i}, & \text{otherwise}.
            \end{cases} \quad\text{for } i\in\{1,...,L\}
        \]
    \ENDIF}
    \STATE $\x_t^{1:L} \gets \x_{s}^{1:L}$.
\ENDFOR

\STATE \textbf{Output:} final sample $\x_{t_T}^{1:L}$ 
\end{algorithmic}
\end{algorithm}

\section{Limitations and Future Works}
Although our discrete stage-1 model improves sparse voxel generation over continuous counterparts, it still requires hundreds of neural function evaluations (NFEs) to obtain strong results, which hinders real-time interactive applications. Future work includes accelerating sampling via fast samplers or distillation, and extending the representation and uncertainty modeling to richer or hierarchical voxel scaffolds. Another potential direction could be modeling both latent features and voxels via a joint discrete-continuous diffusion process.

\newpage
\section{More Generation Results}\label{appendix:moreGenResults}

\begin{figure}[h]
\centering
\reflectbox{
\includegraphics[width=0.13\linewidth]{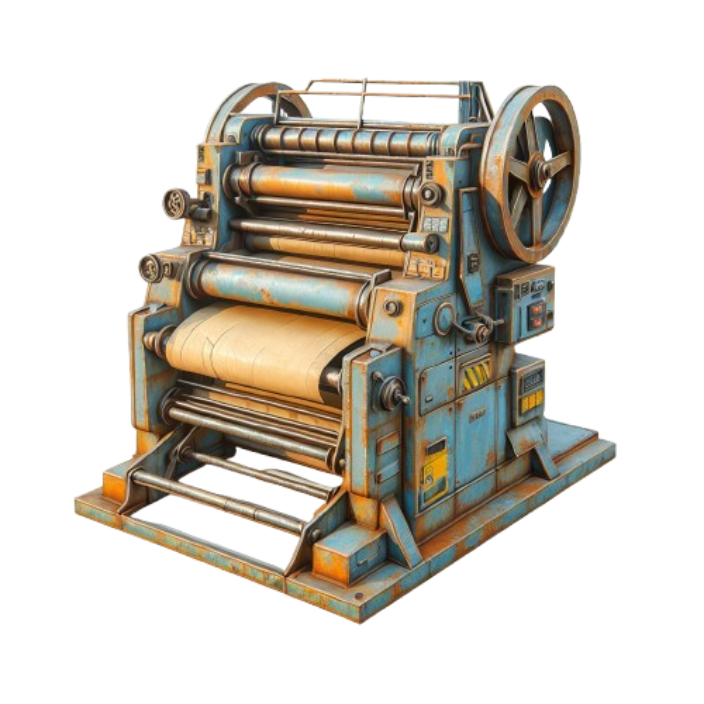}
}
\includegraphics[width=0.13\linewidth]{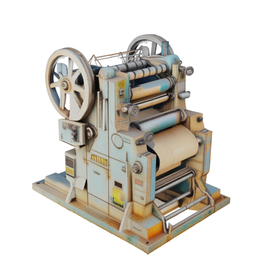}
\includegraphics[width=0.13\linewidth]{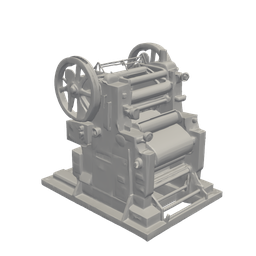}
\includegraphics[width=0.13\linewidth]{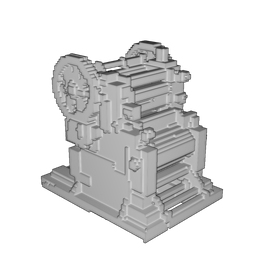}
\includegraphics[width=0.13\linewidth]{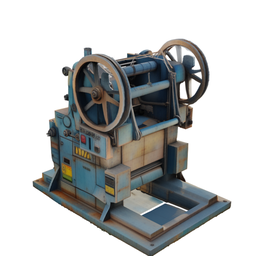}
\includegraphics[width=0.13\linewidth]{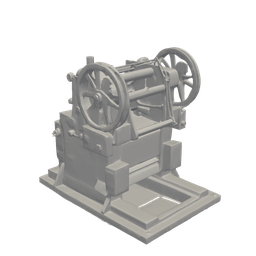}
\includegraphics[width=0.13\linewidth]{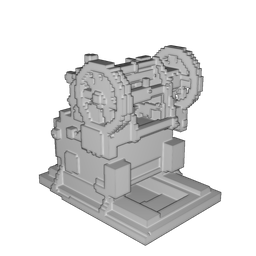}

\includegraphics[width=0.13\linewidth]{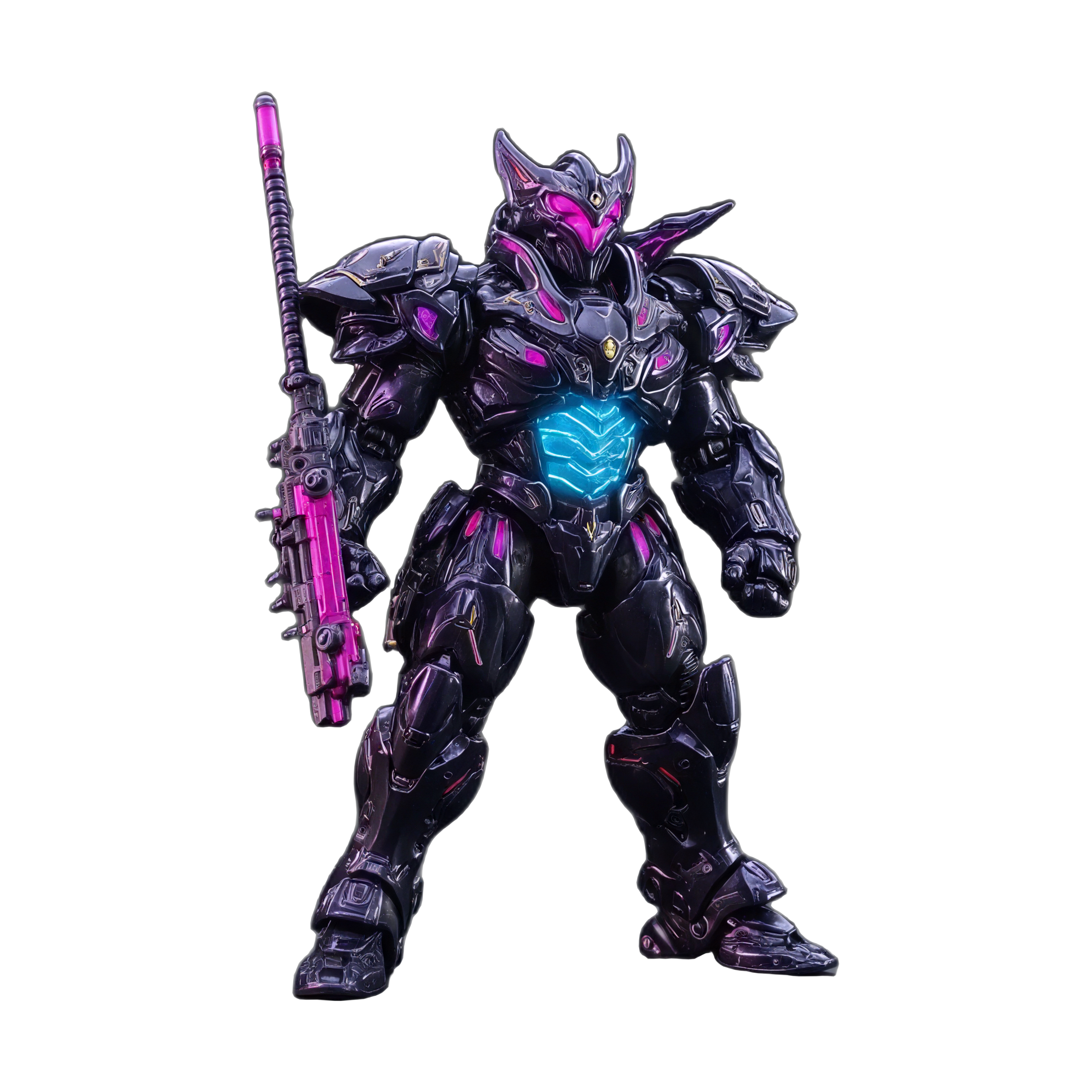}
\includegraphics[width=0.13\linewidth]{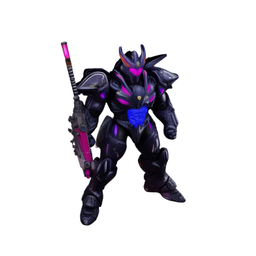}
\includegraphics[width=0.13\linewidth]{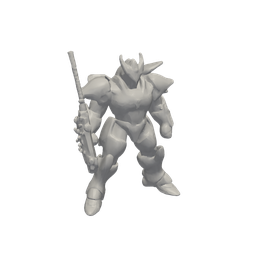}
\reflectbox{
\includegraphics[width=0.13\linewidth]{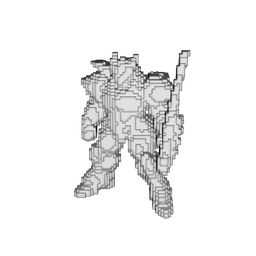}}
\includegraphics[width=0.13\linewidth]{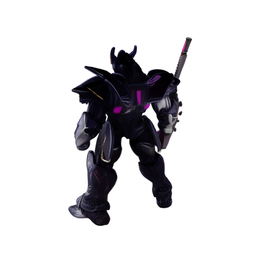}
\includegraphics[width=0.13\linewidth]{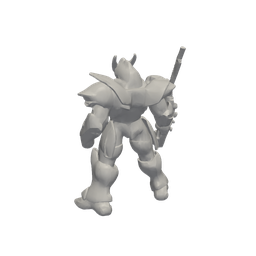}
\reflectbox{
\includegraphics[width=0.13\linewidth]{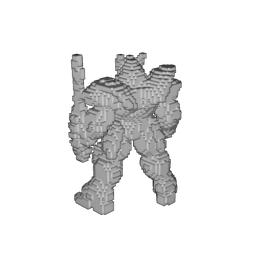}}

\includegraphics[width=0.13\linewidth]{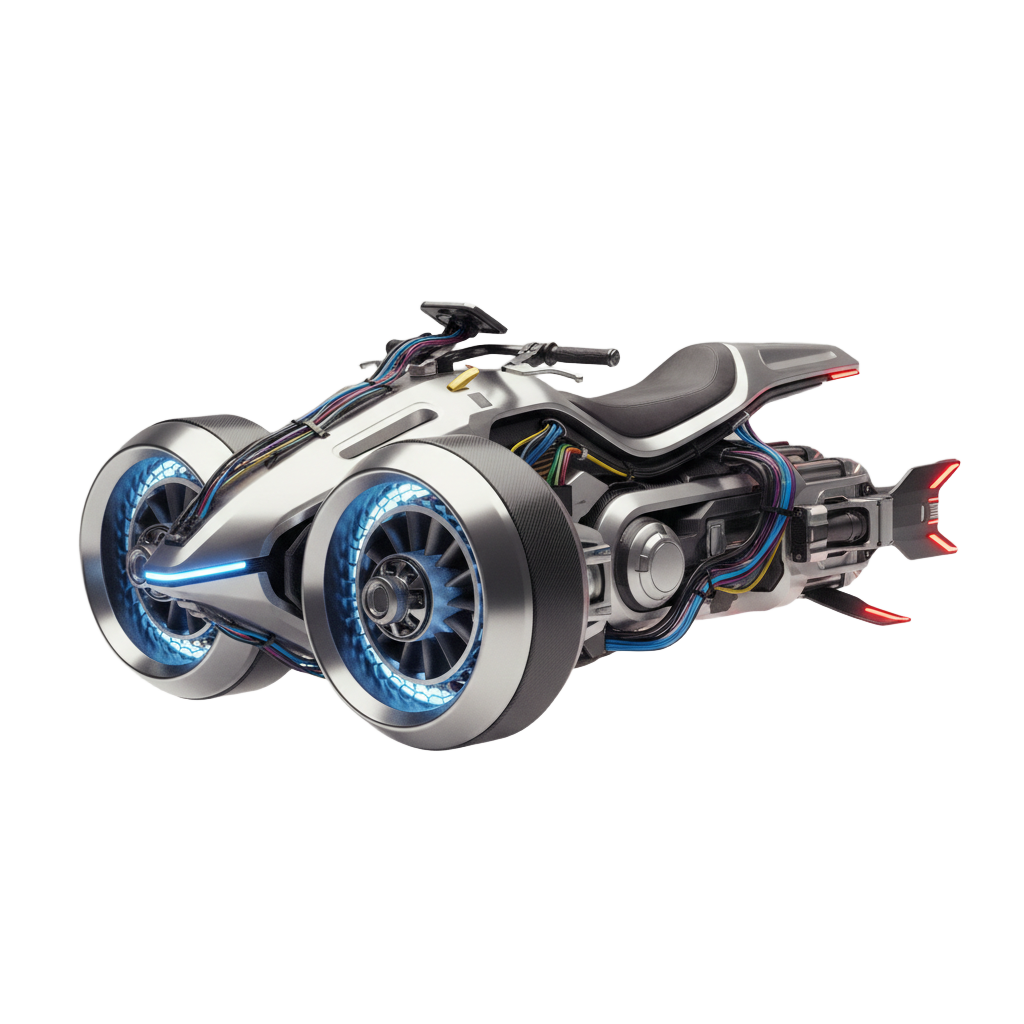}
\includegraphics[width=0.13\linewidth]{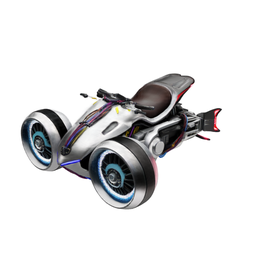}
\includegraphics[width=0.13\linewidth]{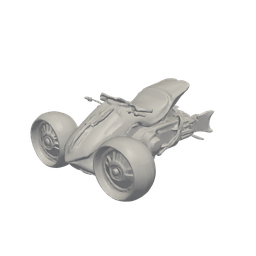}
\includegraphics[width=0.13\linewidth]{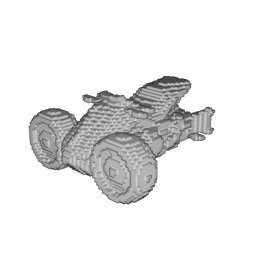}
\includegraphics[width=0.13\linewidth]{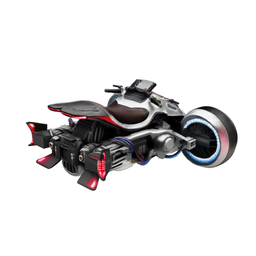}
\includegraphics[width=0.13\linewidth]{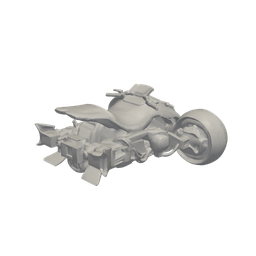}
\includegraphics[width=0.13\linewidth]{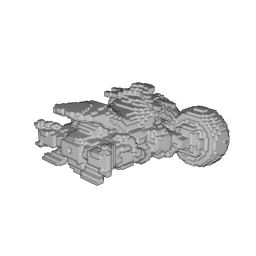}

\includegraphics[width=0.13\linewidth]{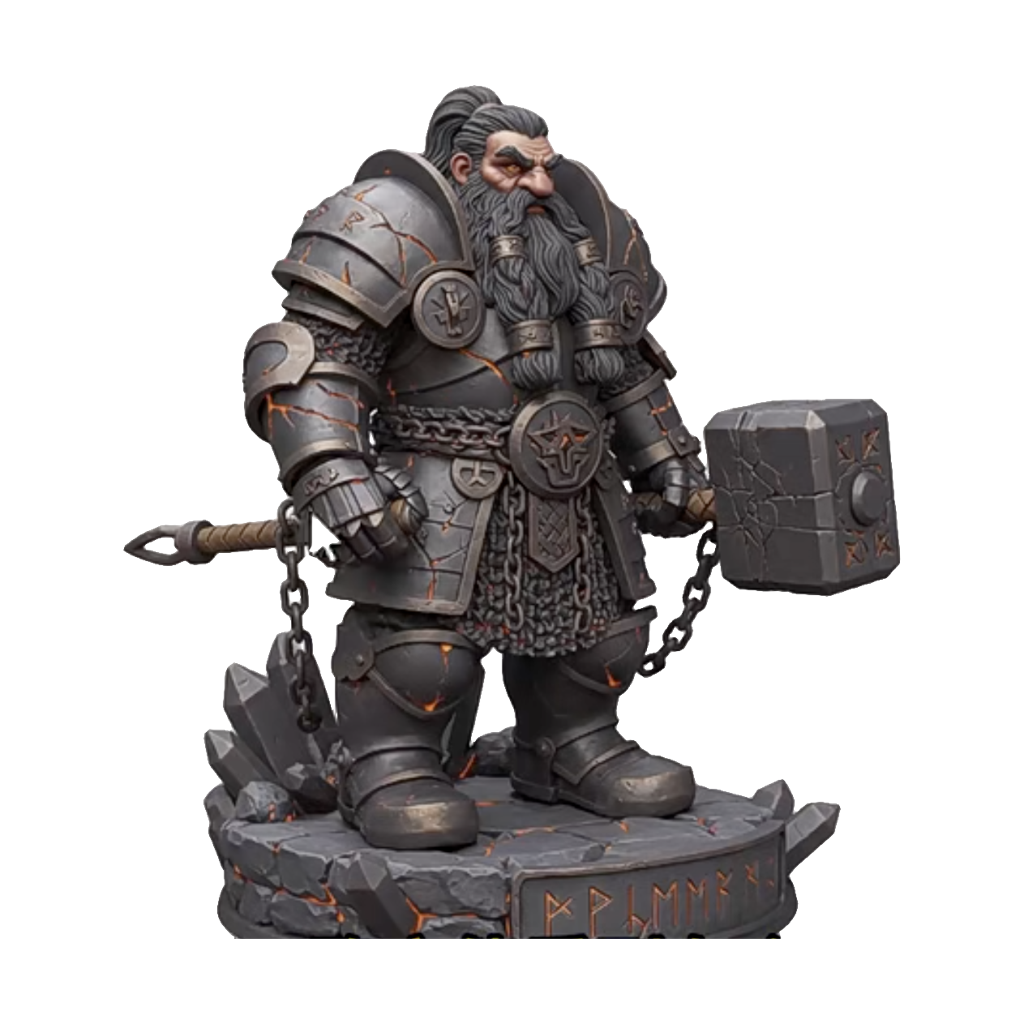}
\includegraphics[width=0.13\linewidth]{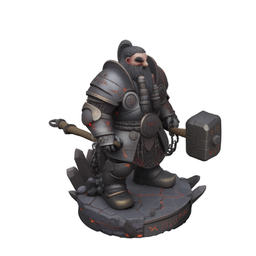}
\includegraphics[width=0.13\linewidth]{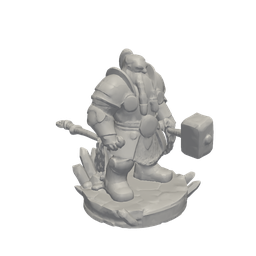}
\includegraphics[width=0.13\linewidth]{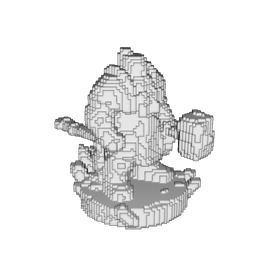}
\includegraphics[width=0.13\linewidth]{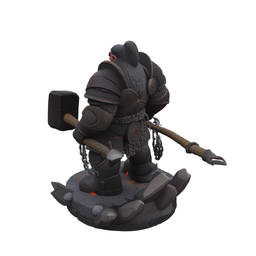}
\includegraphics[width=0.13\linewidth]{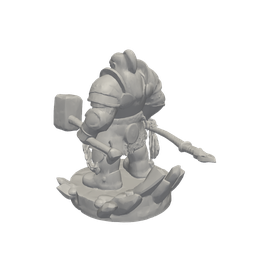}
\includegraphics[width=0.13\linewidth]{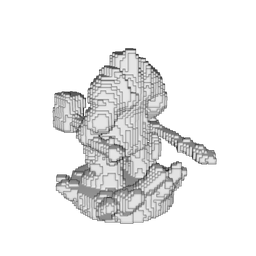}

\includegraphics[width=0.13\linewidth]{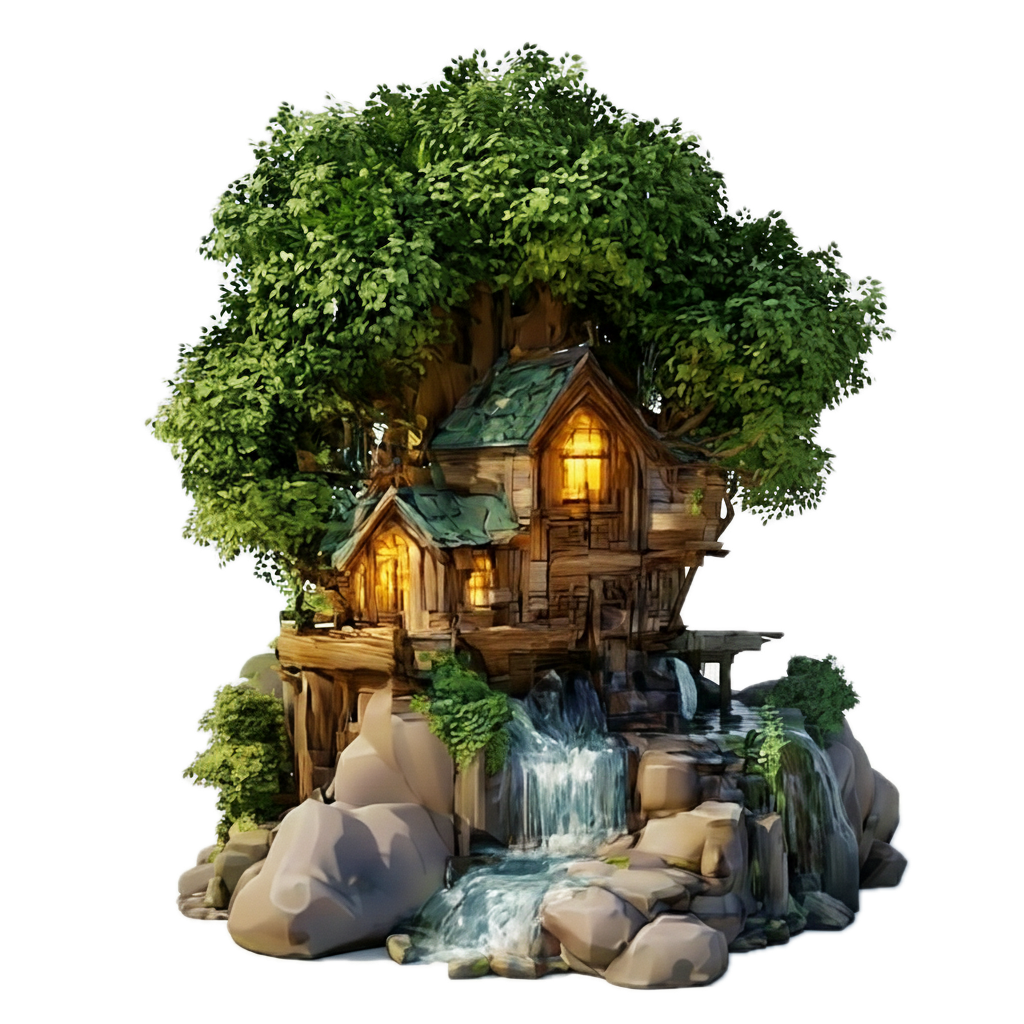}
\includegraphics[width=0.13\linewidth]{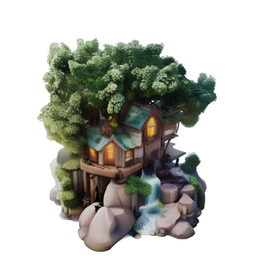}
\includegraphics[width=0.13\linewidth]{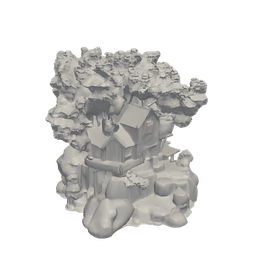}
\reflectbox{
\includegraphics[width=0.13\linewidth]{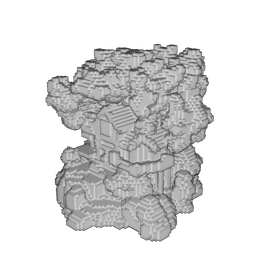}}
\includegraphics[width=0.13\linewidth]{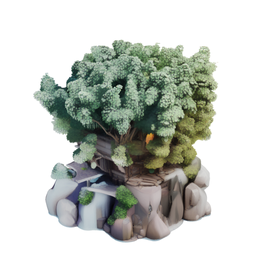}
\includegraphics[width=0.13\linewidth]{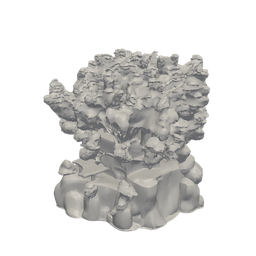}
\reflectbox{
\includegraphics[width=0.13\linewidth]{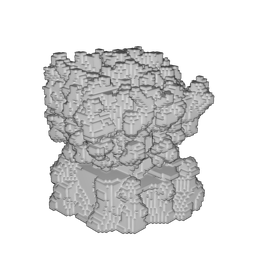}}

\includegraphics[width=0.13\linewidth]{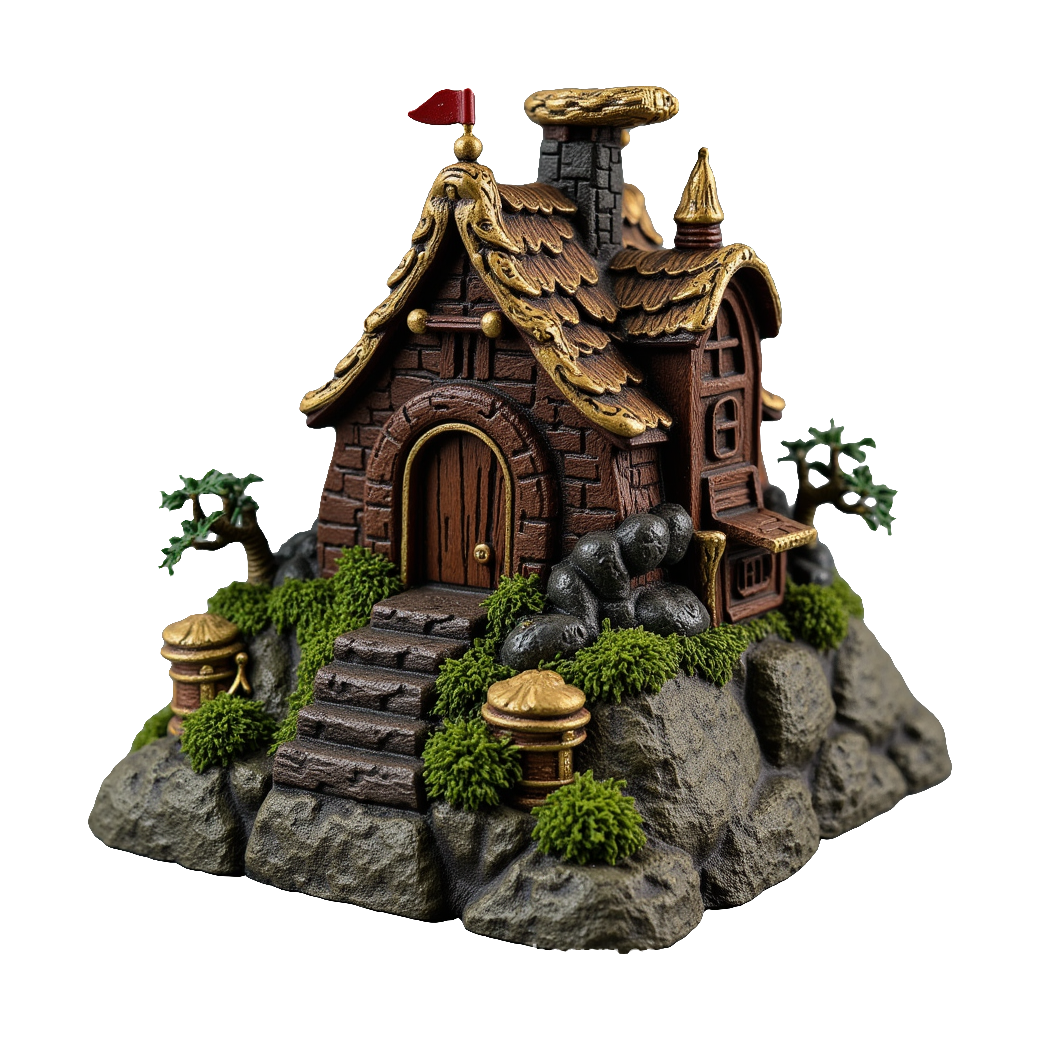}
\includegraphics[width=0.13\linewidth]{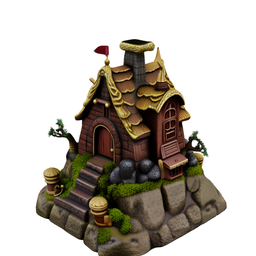}
\includegraphics[width=0.13\linewidth]{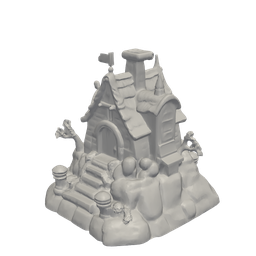}
\includegraphics[width=0.13\linewidth]{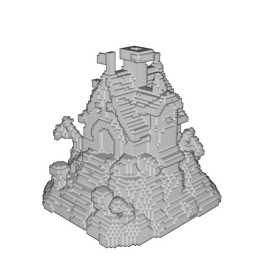}
\includegraphics[width=0.13\linewidth]{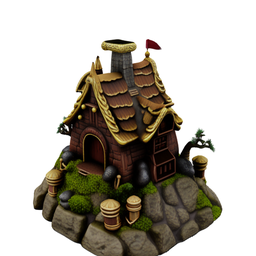}
\includegraphics[width=0.13\linewidth]{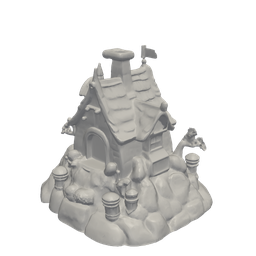}
\includegraphics[width=0.13\linewidth]{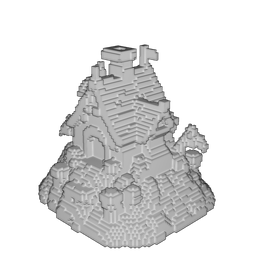}

\includegraphics[width=0.13\linewidth]{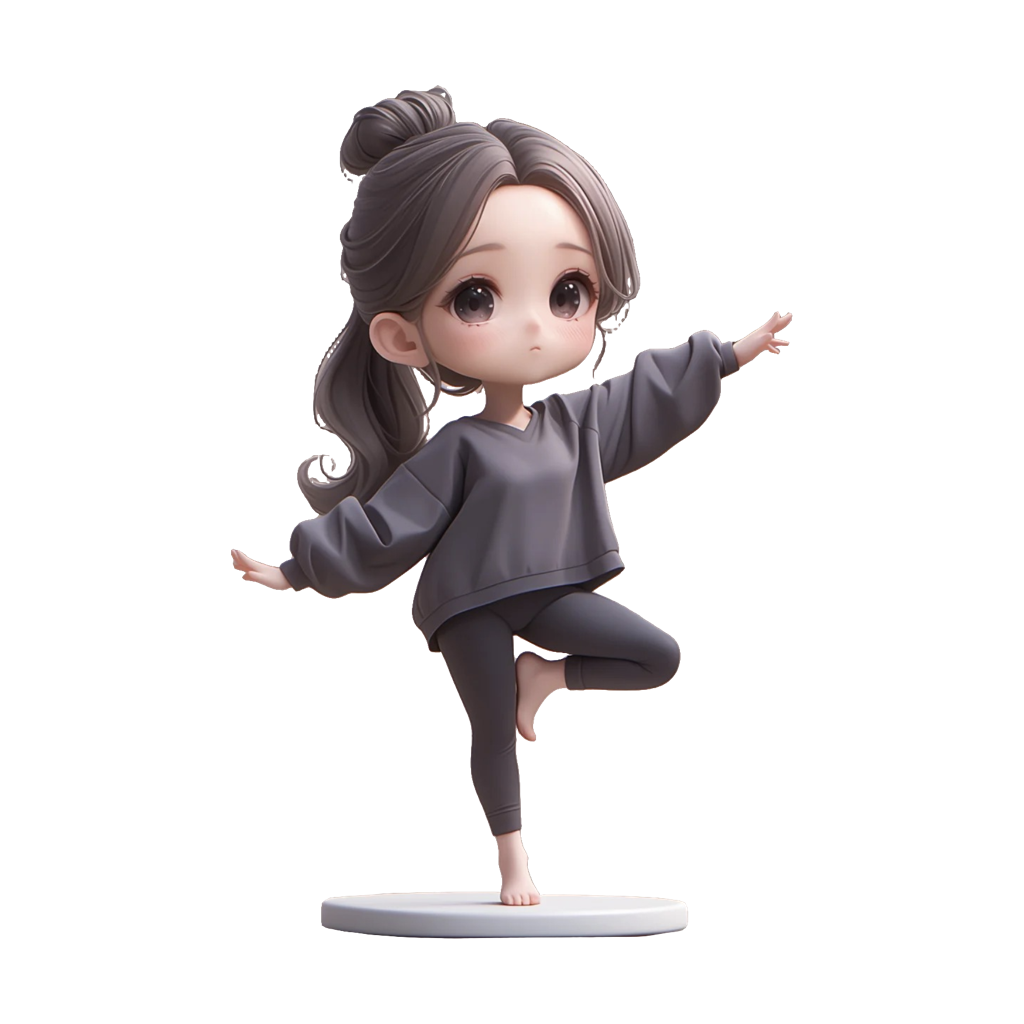}
\includegraphics[width=0.13\linewidth]{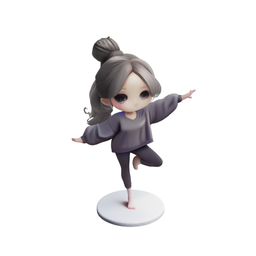}
\includegraphics[width=0.13\linewidth]{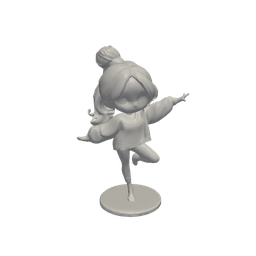}
\includegraphics[width=0.13\linewidth]{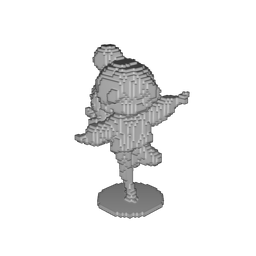}
\includegraphics[width=0.13\linewidth]{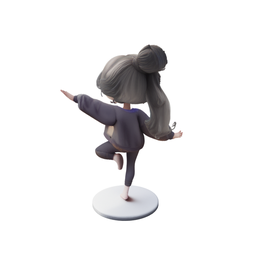}
\includegraphics[width=0.13\linewidth]{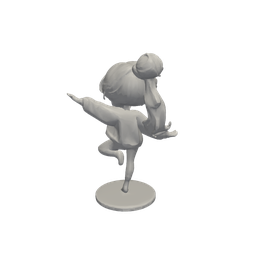}
\includegraphics[width=0.13\linewidth]{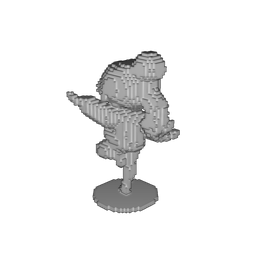}
\includegraphics[width=0.13\linewidth]{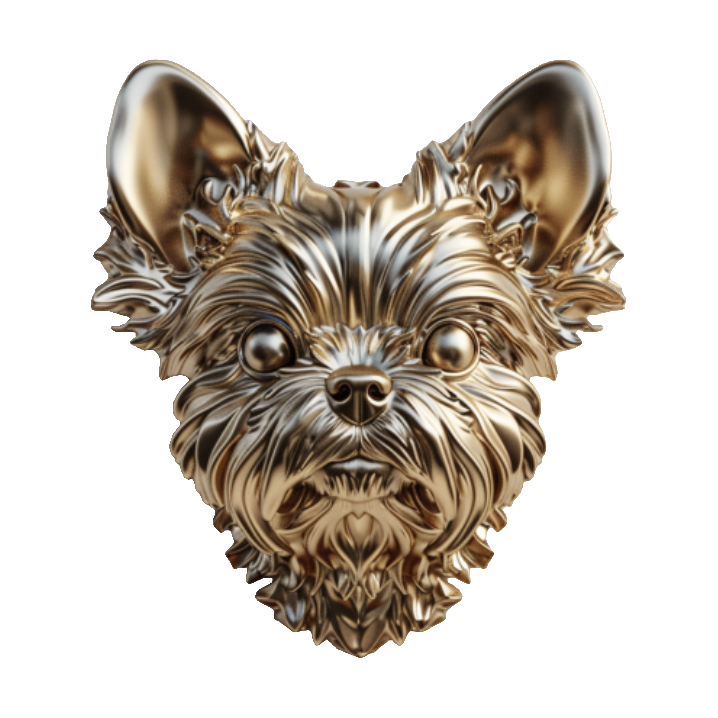}
\includegraphics[width=0.13\linewidth]{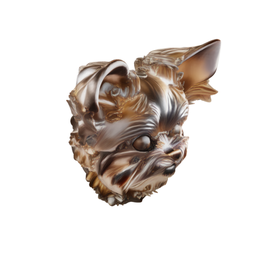}
\includegraphics[width=0.13\linewidth]{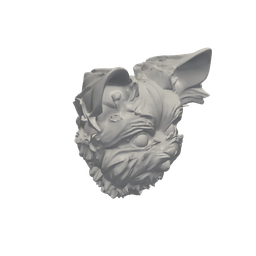}
\includegraphics[width=0.13\linewidth]{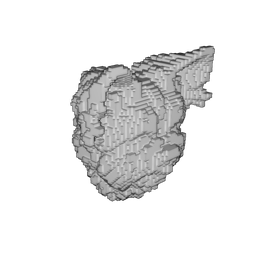}
\includegraphics[width=0.13\linewidth]{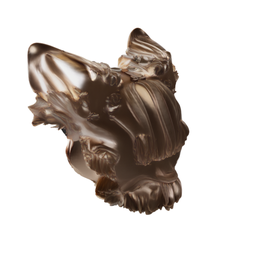}
\includegraphics[width=0.13\linewidth]{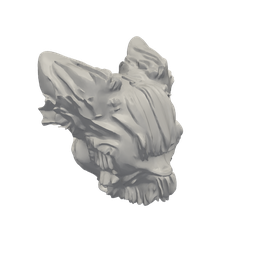}
\includegraphics[width=0.13\linewidth]{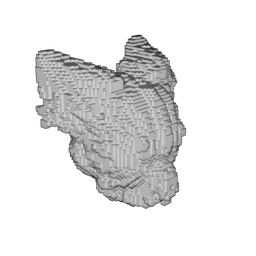}

\includegraphics[width=0.13\linewidth]{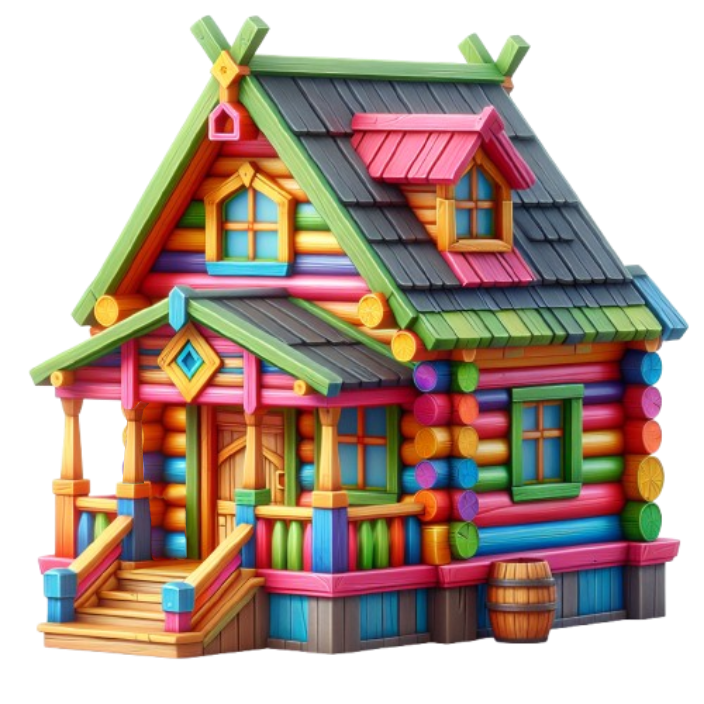}
\includegraphics[width=0.13\linewidth]{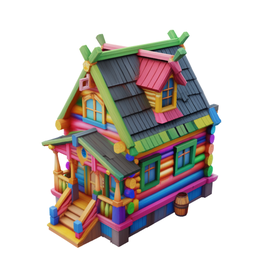}
\includegraphics[width=0.13\linewidth]{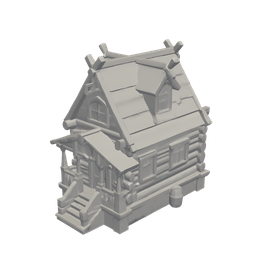}
\includegraphics[width=0.13\linewidth]{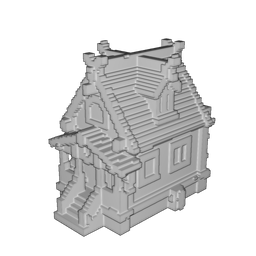}
\includegraphics[width=0.13\linewidth]{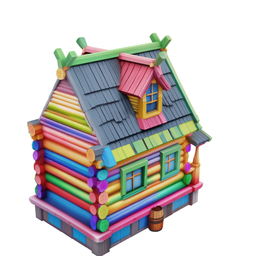}
\includegraphics[width=0.13\linewidth]{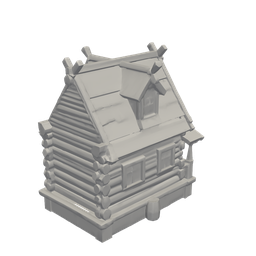}
\includegraphics[width=0.13\linewidth]{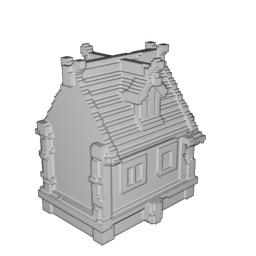}

\includegraphics[width=0.13\linewidth]{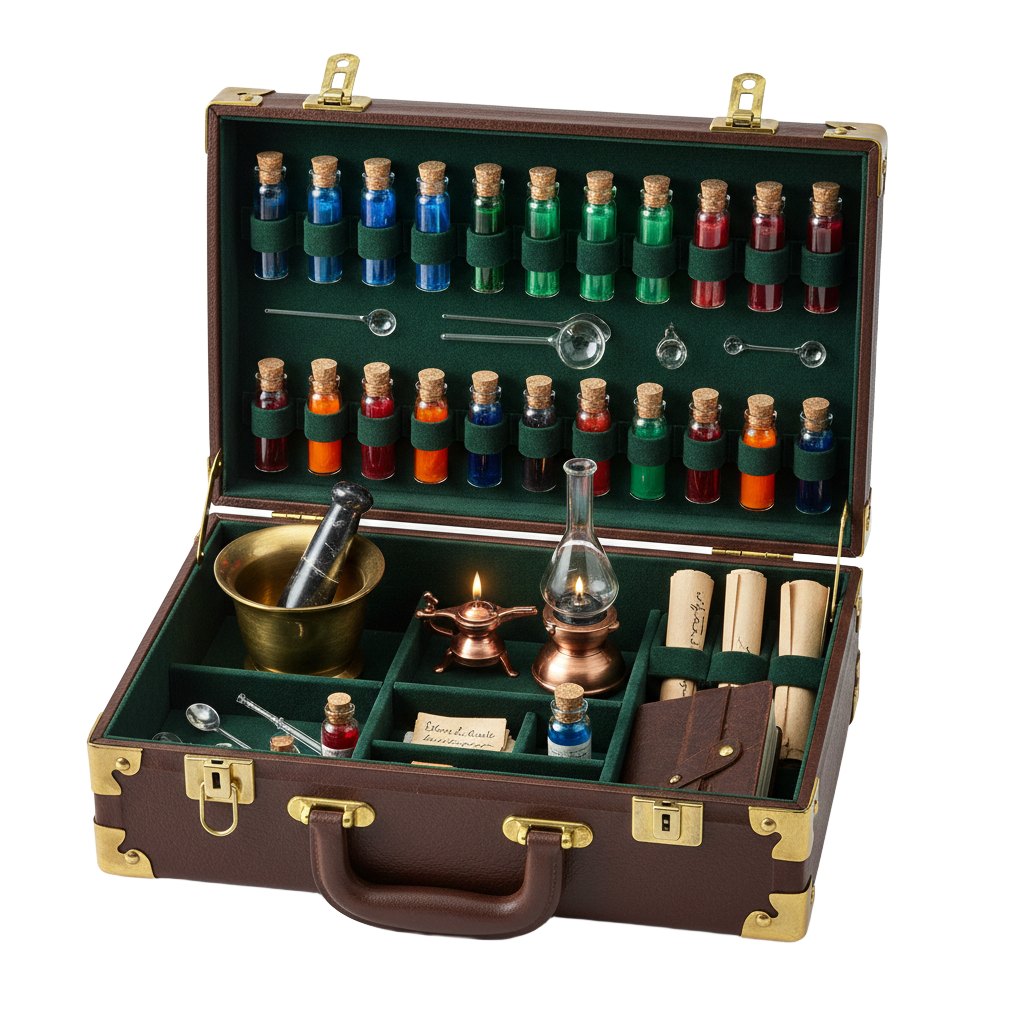}
\includegraphics[width=0.13\linewidth]{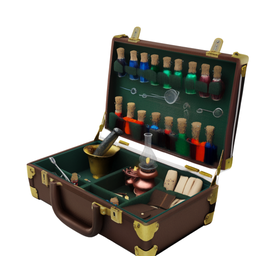}
\includegraphics[width=0.13\linewidth]{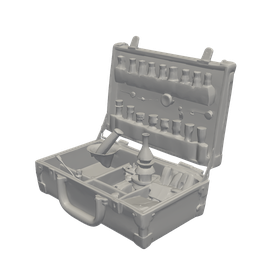}
\reflectbox{
\includegraphics[width=0.13\linewidth]{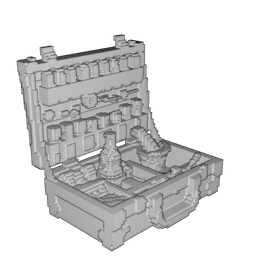}}
\includegraphics[width=0.13\linewidth]{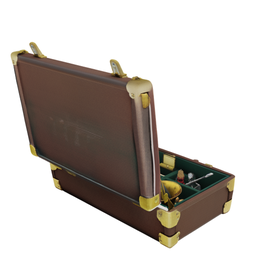}
\includegraphics[width=0.13\linewidth]{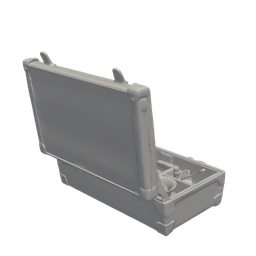}
\reflectbox{
\includegraphics[width=0.13\linewidth]{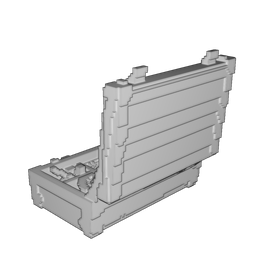}}

\caption{More image-conditioned generation result. Image prompt in the first column.}
\label{fig:imgresult_1}
\end{figure}

\begin{figure}
    \centering

\includegraphics[width=0.13\linewidth]{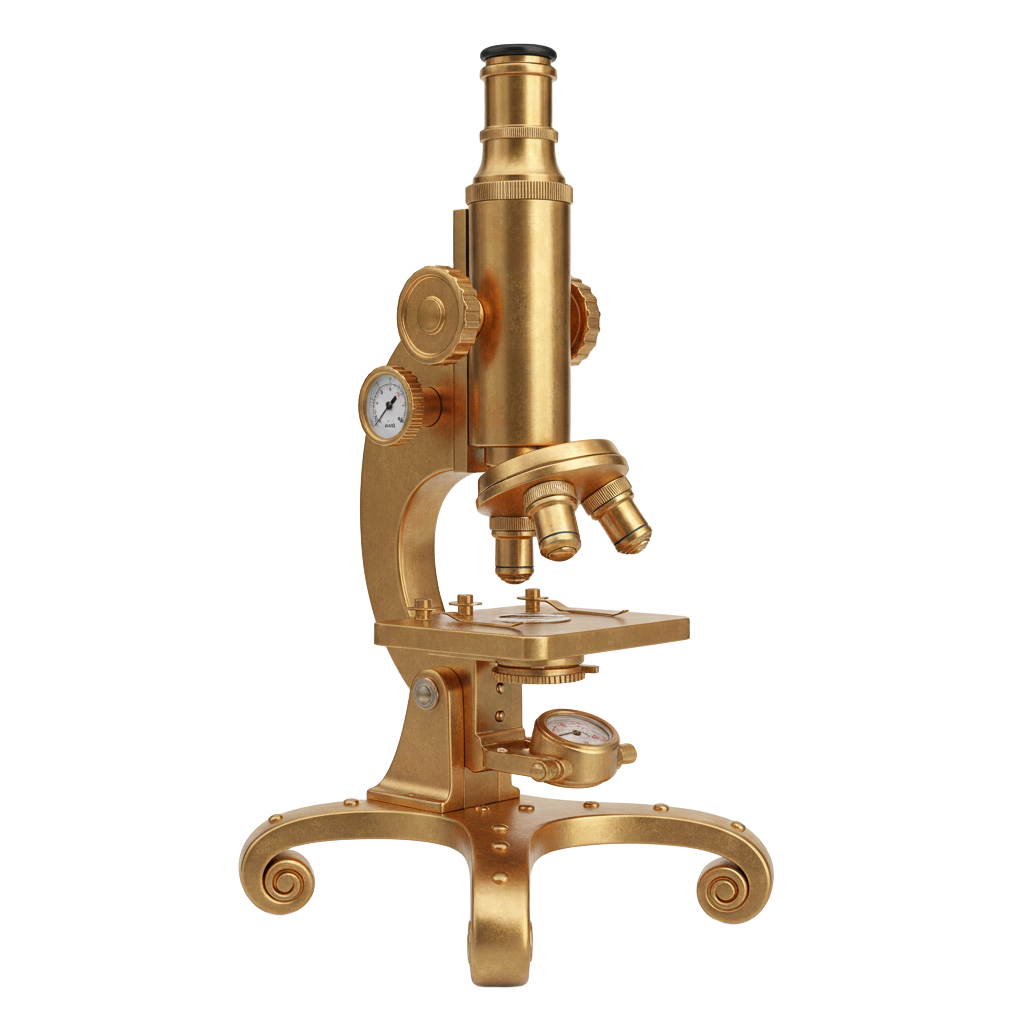}
\includegraphics[width=0.13\linewidth]{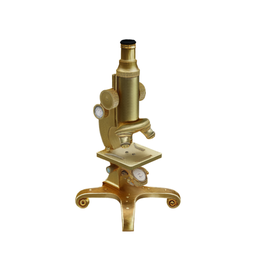}
\includegraphics[width=0.13\linewidth]{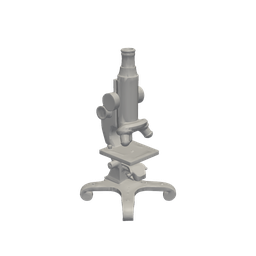}
\reflectbox{
\includegraphics[width=0.13\linewidth]{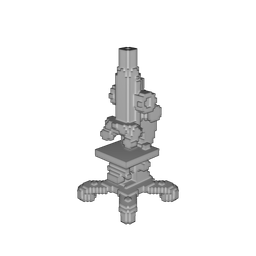}}
\includegraphics[width=0.13\linewidth]
{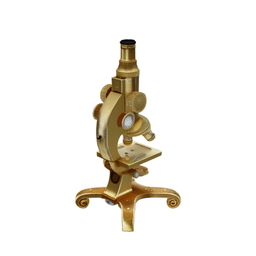}
\includegraphics[width=0.13\linewidth]{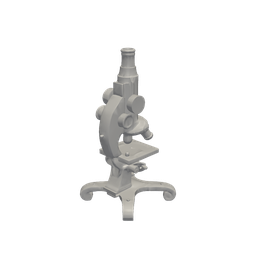}
\reflectbox{
\includegraphics[width=0.13\linewidth]{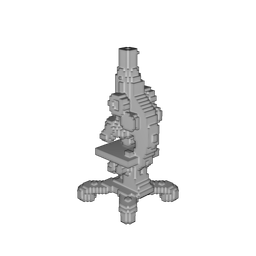}}

\includegraphics[width=0.13\linewidth]{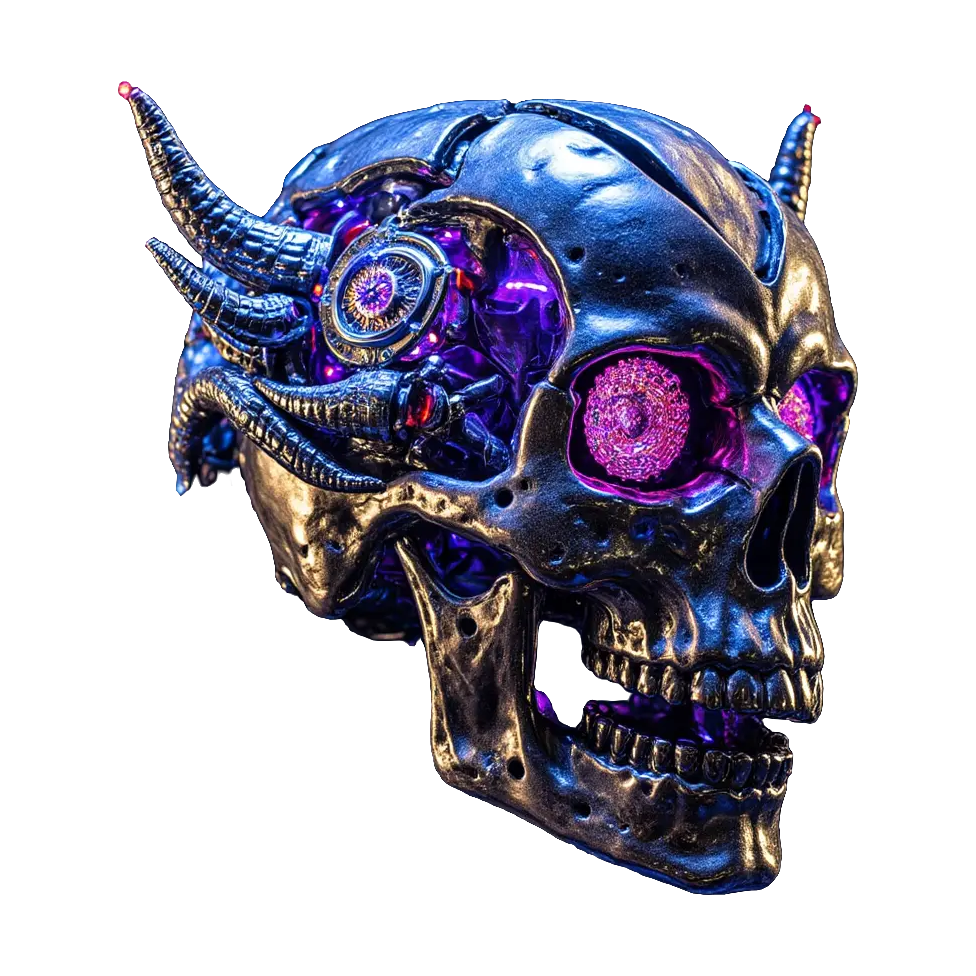}
\includegraphics[width=0.13\linewidth]{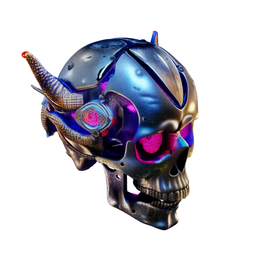}
\includegraphics[width=0.13\linewidth]{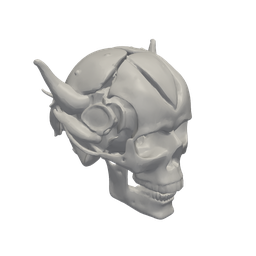}
\includegraphics[width=0.13\linewidth]{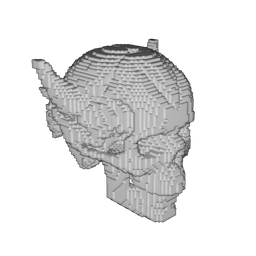}
\includegraphics[width=0.13\linewidth]{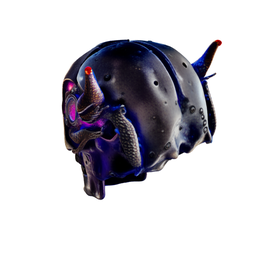}
\includegraphics[width=0.13\linewidth]{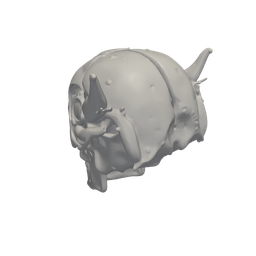}
\includegraphics[width=0.13\linewidth]{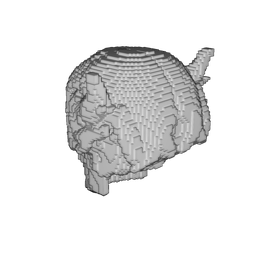}

\includegraphics[width=0.13\linewidth]{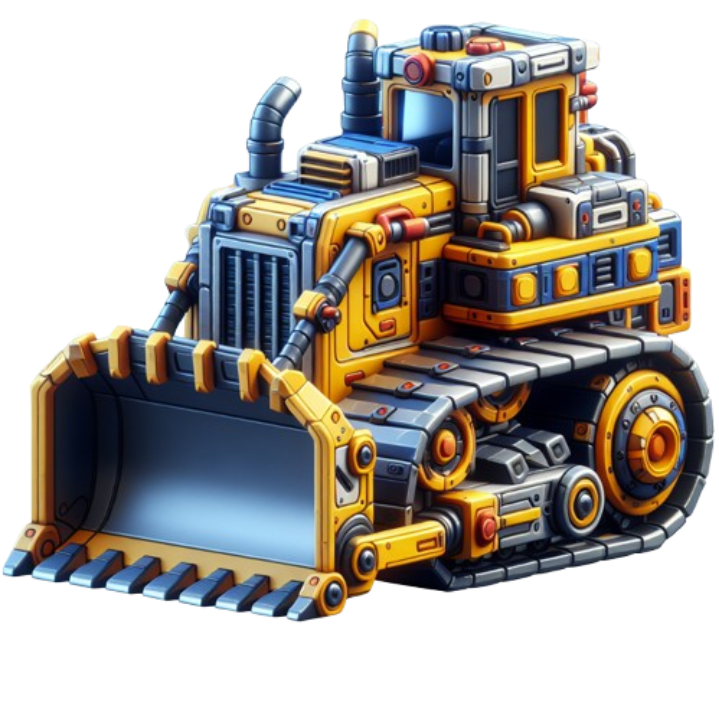}
\includegraphics[width=0.13\linewidth]{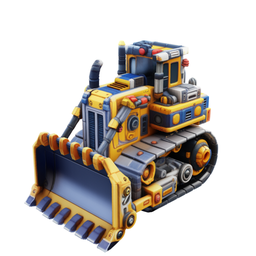}
\includegraphics[width=0.13\linewidth]{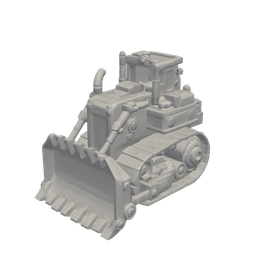}
\includegraphics[width=0.13\linewidth]{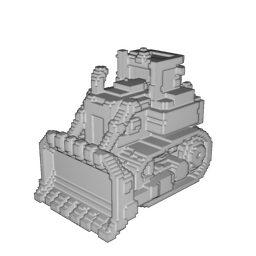}
\includegraphics[width=0.13\linewidth]{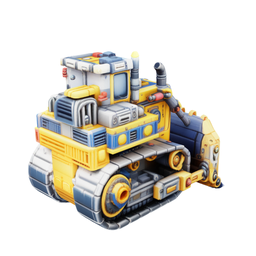}
\includegraphics[width=0.13\linewidth]{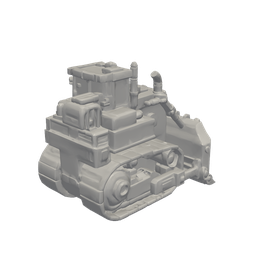}
\includegraphics[width=0.13\linewidth]{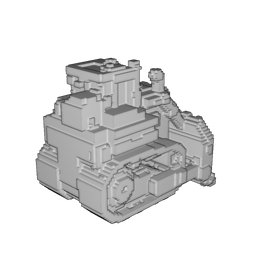}

\caption{More image-conditioned generation result. Image prompt in the first column.}
\label{fig:imgresult_2}
\end{figure}

\begin{figure}
    \centering
\includegraphics[width=0.16\linewidth]{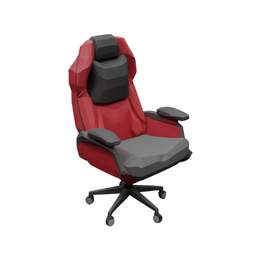}
\includegraphics[width=0.16\linewidth]{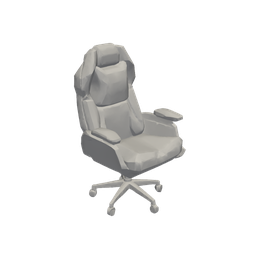}
\includegraphics[width=0.16\linewidth]{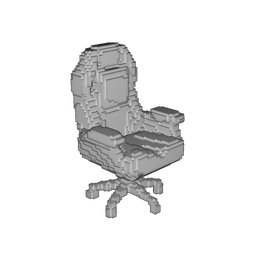}
\includegraphics[width=0.16\linewidth]{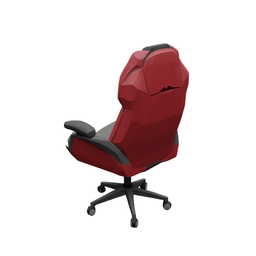}
\includegraphics[width=0.16\linewidth]{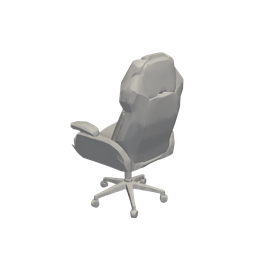}
\includegraphics[width=0.16\linewidth]{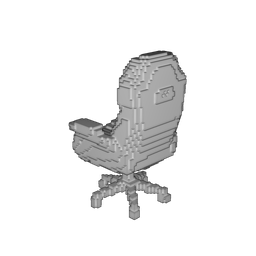}
\\A low-poly gaming chair in red and gray with stitched cushions, wheels, and metal frame.\\
\includegraphics[width=0.16\linewidth]{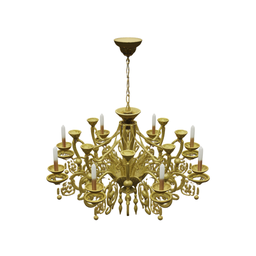}
\includegraphics[width=0.16\linewidth]{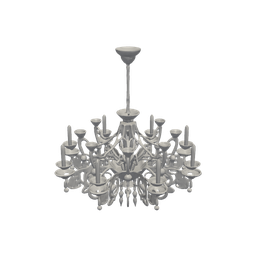}
\includegraphics[width=0.16\linewidth]{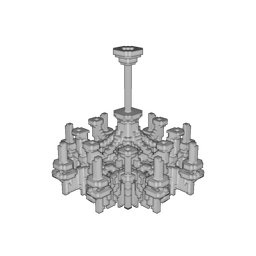}
\includegraphics[width=0.16\linewidth]{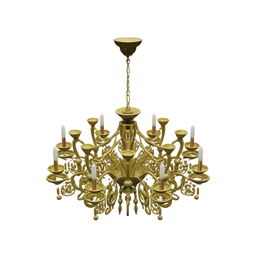}
\includegraphics[width=0.16\linewidth]{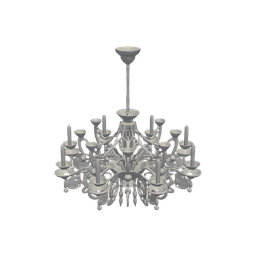}
\includegraphics[width=0.16\linewidth]{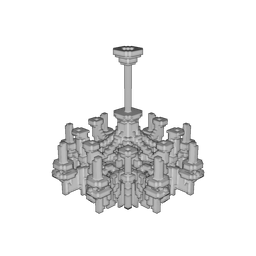}
\\An ornate chandelier in gold metal with multiple thin arms, candle-shaped bulbs, and dangling crystal beads.\\
\includegraphics[width=0.16\linewidth]{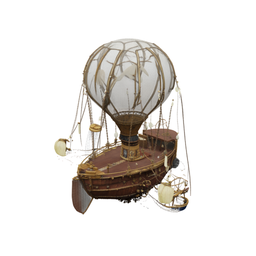}
\includegraphics[width=0.16\linewidth]{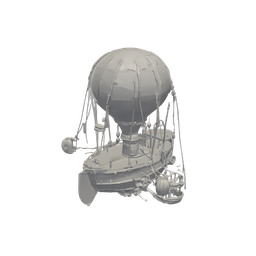}
\reflectbox{
\includegraphics[width=0.16\linewidth]
{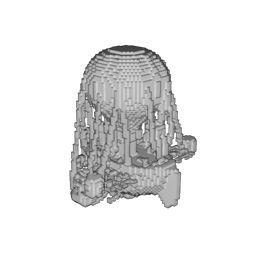}}
\includegraphics[width=0.16\linewidth]{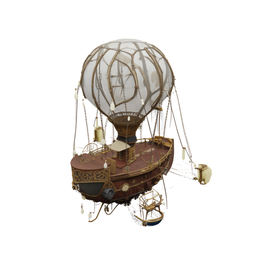}
\includegraphics[width=0.16\linewidth]{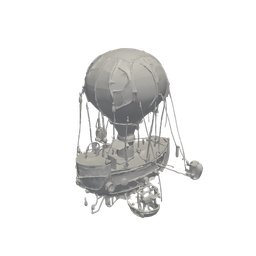}
\reflectbox{
\includegraphics[width=0.155\linewidth]{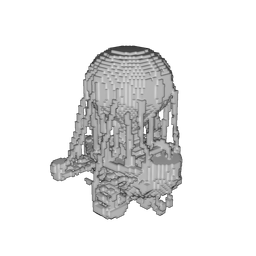}} 
\\A steampunk airship with a brass gondola, cream balloon envelope, side propellers, hanging ropes, and small lanterns.\\

    \caption{More text-conditioned generation result. Prompts Generated by GPT5.2.}
\end{figure}

\begin{figure}
    \centering
\includegraphics[width=0.16\linewidth]{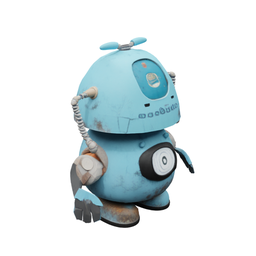}
\includegraphics[width=0.16\linewidth]{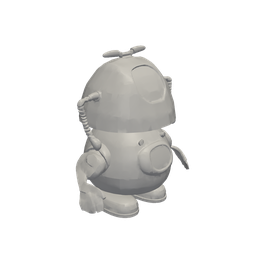}
\includegraphics[width=0.16\linewidth]{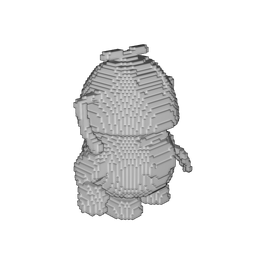}
\includegraphics[width=0.16\linewidth]{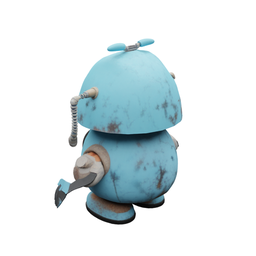}
\includegraphics[width=0.16\linewidth]{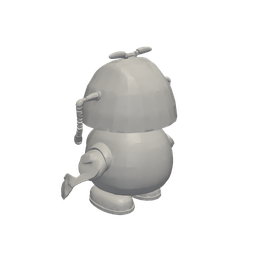}
\includegraphics[width=0.16\linewidth]{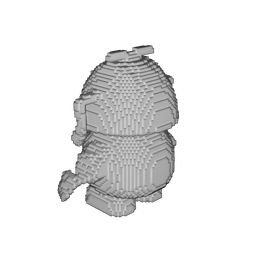}
\\A small helper robot with a round body, sky-blue paint, screen face, and tiny tool arms.\\
\includegraphics[width=0.16\linewidth]{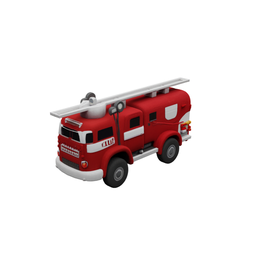}
\includegraphics[width=0.16\linewidth]{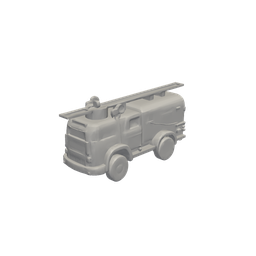}
\reflectbox{
\includegraphics[width=0.16\linewidth]{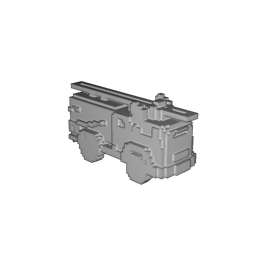}}
\includegraphics[width=0.16\linewidth]{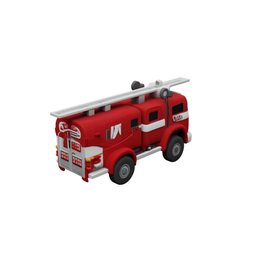}
\includegraphics[width=0.16\linewidth]{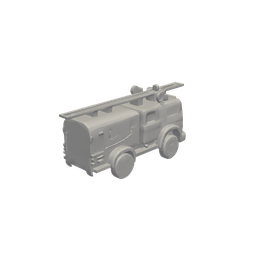}
\reflectbox{
\includegraphics[width=0.155\linewidth]{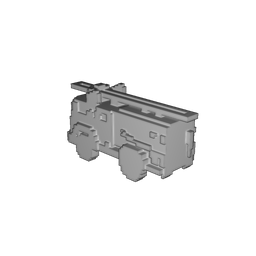}}
\\A cartoon fire truck in deep red with white accents, silver ladder on top, black tires.\\
\includegraphics[width=0.16\linewidth]{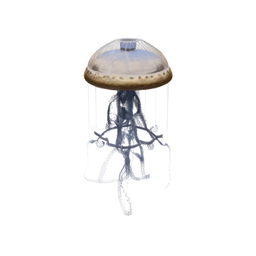}
\includegraphics[width=0.16\linewidth]{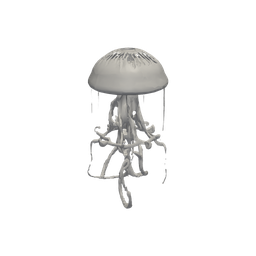}
\reflectbox{
\includegraphics[width=0.16\linewidth]{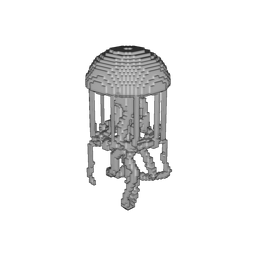}}
\includegraphics[width=0.16\linewidth]{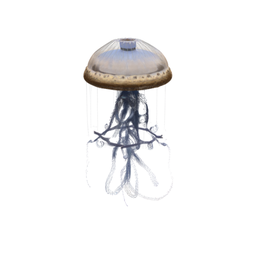}
\includegraphics[width=0.16\linewidth]{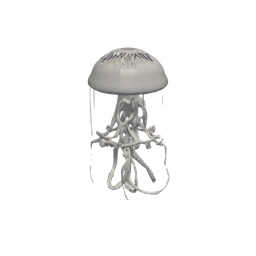}
\reflectbox{
\includegraphics[width=0.155\linewidth]{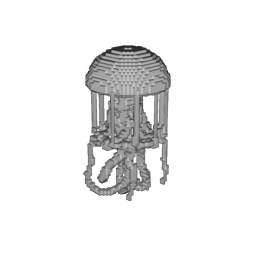}}
\\A 3D model of a mechanical jellyfish with a round bell and separated tentacles, centered, plain background, sharp silhouette.\\
\includegraphics[width=0.16\linewidth]{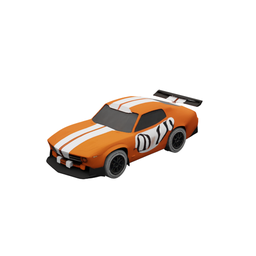}
\includegraphics[width=0.16\linewidth]{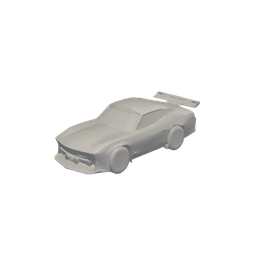}
\reflectbox{
\includegraphics[width=0.16\linewidth]{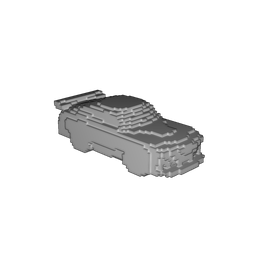}}
\includegraphics[width=0.16\linewidth]{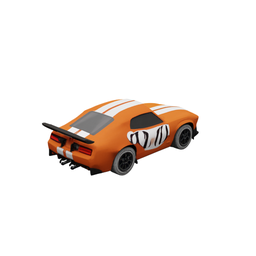}
\includegraphics[width=0.16\linewidth]{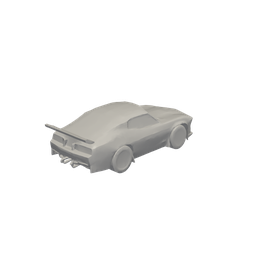}
\reflectbox{
\includegraphics[width=0.155\linewidth]{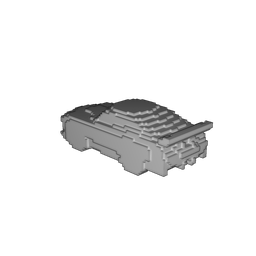}}
\\A die-cast race car in bright orange with white stripes, spoiler, and rubbery tires.\\
\includegraphics[width=0.16\linewidth]{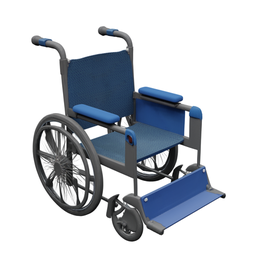}
\includegraphics[width=0.16\linewidth]{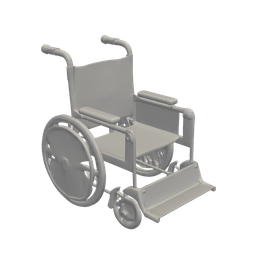}
\includegraphics[width=0.16\linewidth]{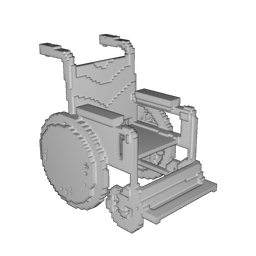}
\includegraphics[width=0.16\linewidth]{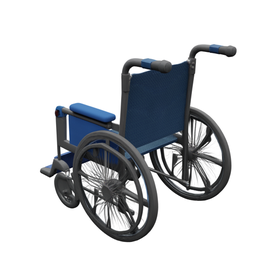}
\includegraphics[width=0.16\linewidth]{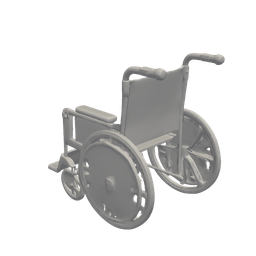}
\includegraphics[width=0.16\linewidth]{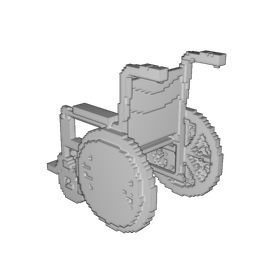}
\\A wheelchair with large rear wheels, small front casters, dark gray frame, blue seat cushion, and visible push handles.\\
\includegraphics[width=0.16\linewidth]{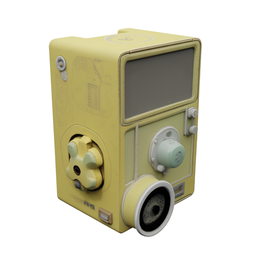}
\includegraphics[width=0.16\linewidth]{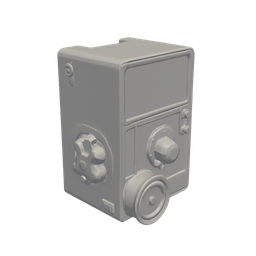}
\includegraphics[width=0.16\linewidth]{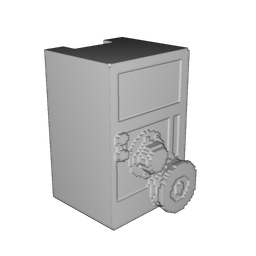}
\includegraphics[width=0.16\linewidth]{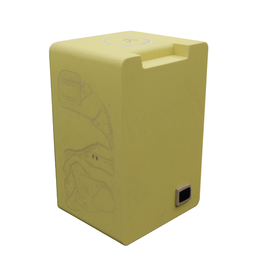}
\includegraphics[width=0.16\linewidth]{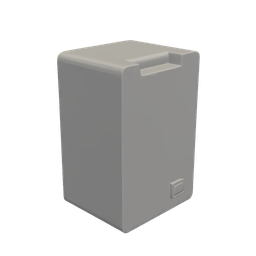}
\includegraphics[width=0.16\linewidth]{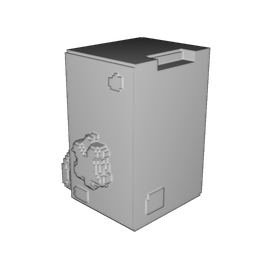}
\\A retro instant camera in pastel yellow with a big lens, flash cube, and worn corners.\\
\includegraphics[width=0.16\linewidth]{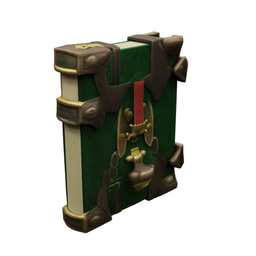}
\includegraphics[width=0.16\linewidth]{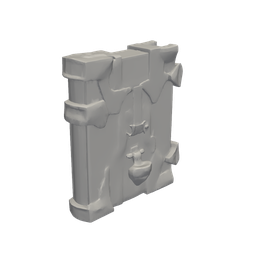}
\includegraphics[width=0.16\linewidth]{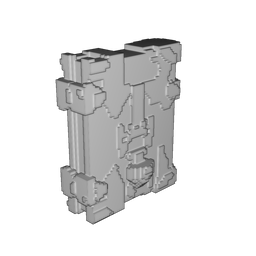}
\includegraphics[width=0.16\linewidth]{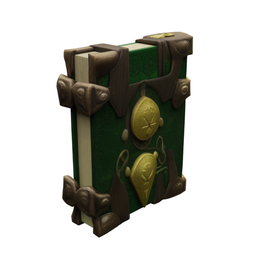}
\includegraphics[width=0.16\linewidth]{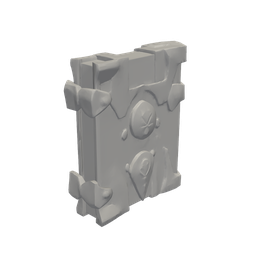}
\includegraphics[width=0.16\linewidth]{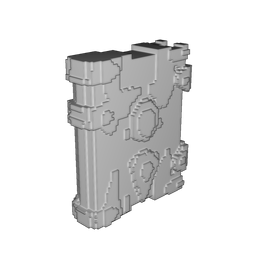}
\\A spellbook with deep green cover, gold embossing, corner metal guards, and a ribbon bookmark.\\

    \caption{More text-conditioned generation result. Prompts Generated by GPT5.2.}
\end{figure}

\begin{figure}[t]
    \centering

\includegraphics[width=0.16\linewidth]{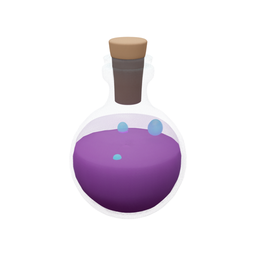}
\includegraphics[width=0.16\linewidth]{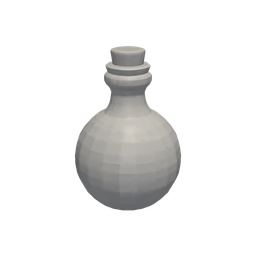}
\includegraphics[width=0.16\linewidth]{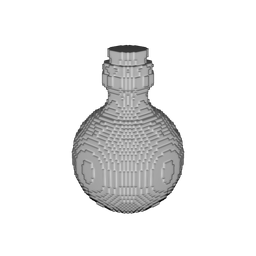}
\includegraphics[width=0.16\linewidth]{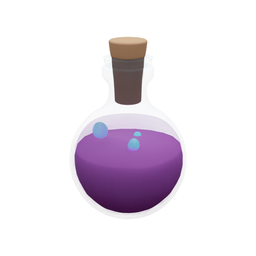}
\includegraphics[width=0.16\linewidth]{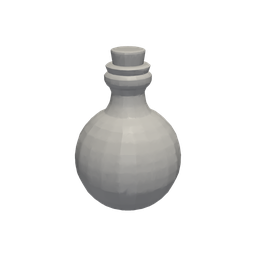}
\includegraphics[width=0.16\linewidth]{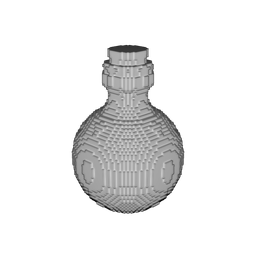}
\\A round glass potion bottle with purple liquid, cork stopper, label tag, and tiny bubbles.\\
\includegraphics[width=0.16\linewidth]{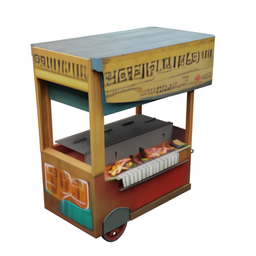}
\includegraphics[width=0.16\linewidth]{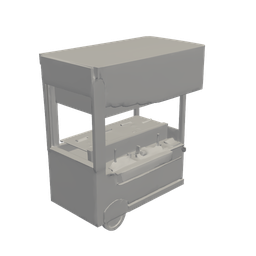}
\reflectbox{
\includegraphics[width=0.16\linewidth]{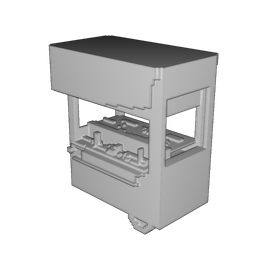}}
\includegraphics[width=0.16\linewidth]{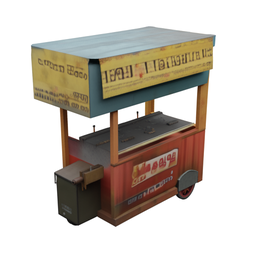}
\includegraphics[width=0.16\linewidth]{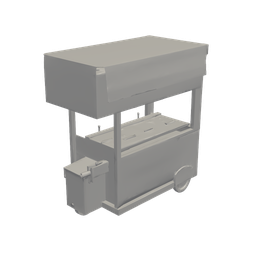}
\reflectbox{
\includegraphics[width=0.155\linewidth]{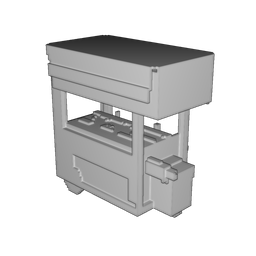}}
\\A street food cart: canopy, menu board, skewers on a grill, and a small cash box; include wooden wheels and handle.\\
\includegraphics[width=0.16\linewidth]{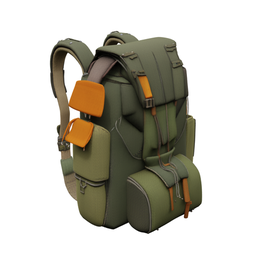}
\includegraphics[width=0.16\linewidth]{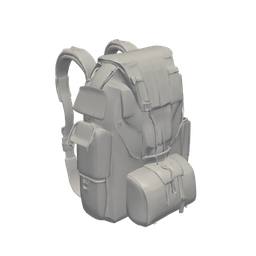}
\includegraphics[width=0.16\linewidth]{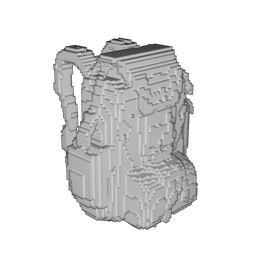}
\includegraphics[width=0.16\linewidth]{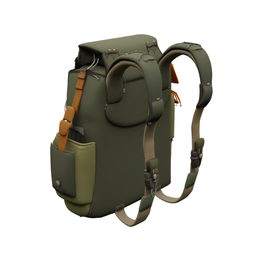}
\includegraphics[width=0.16\linewidth]{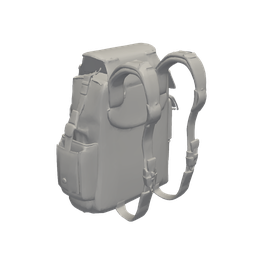}
\includegraphics[width=0.16\linewidth]{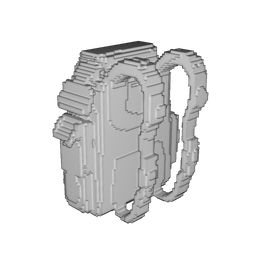}
\\A compact hiking backpack in olive green with orange straps, zipper pulls, and a rolled sleeping mat.\\
\includegraphics[width=0.16\linewidth]{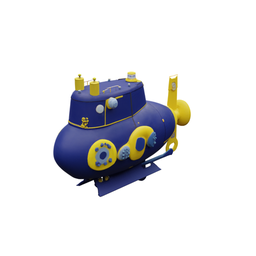}
\includegraphics[width=0.16\linewidth]{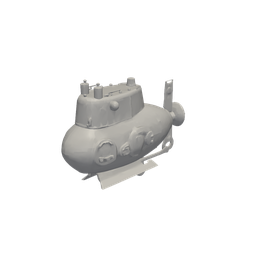}
\reflectbox{
\includegraphics[width=0.16\linewidth]{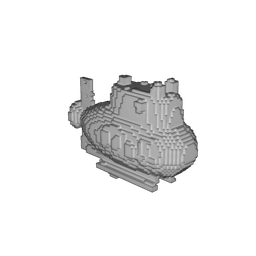}}
\includegraphics[width=0.16\linewidth]{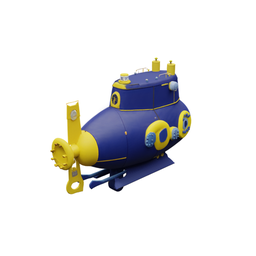}
\includegraphics[width=0.16\linewidth]{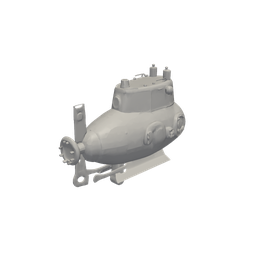}
\reflectbox{
\includegraphics[width=0.155\linewidth]{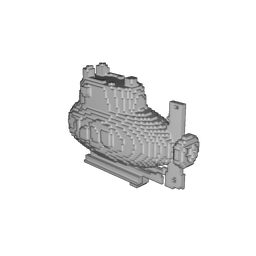}}
\\A cartoon submarine with a rounded navy-blue hull, yellow accents, circular portholes, a small periscope, and a rear propeller.\\
    \caption{More text-conditioned generation result. Prompts Generated by GPT5.2.}
\end{figure}

We also present some editing results giving different image and text prompts.
\begin{figure}
    \centering
    \includegraphics[width=0.1\linewidth]{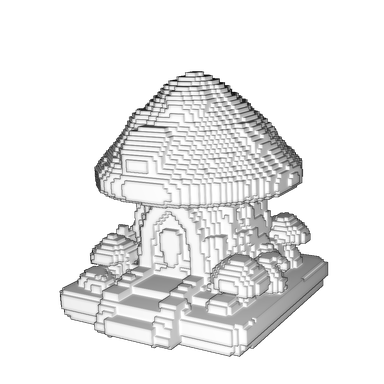}
    \includegraphics[width=0.1\linewidth]{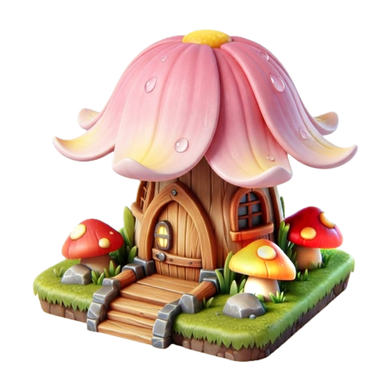}
    \includegraphics[width=0.1\linewidth]{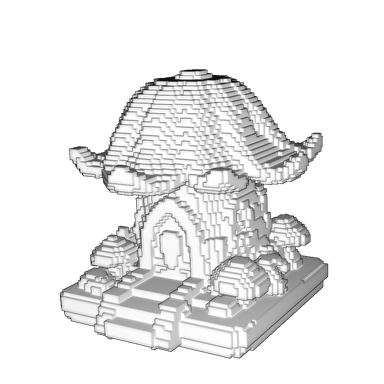}
    \includegraphics[width=0.1\linewidth]{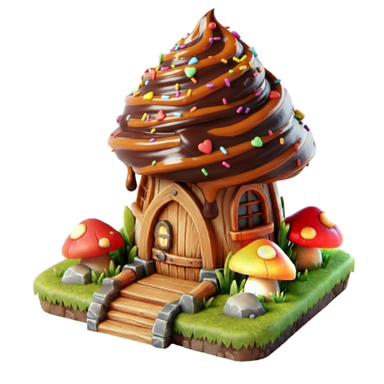}
    \includegraphics[width=0.1\linewidth]{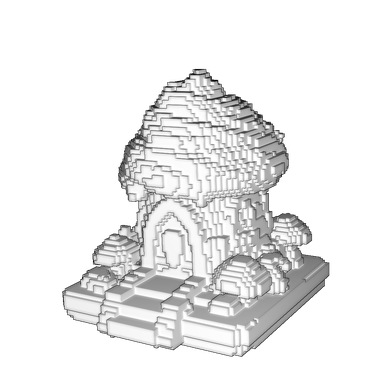}
    \includegraphics[width=0.1\linewidth]{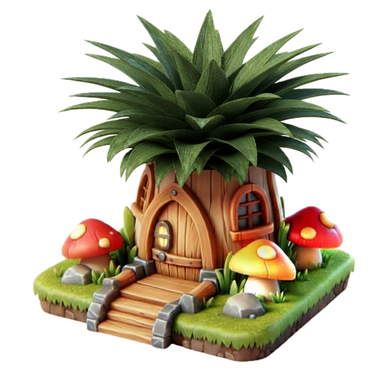}
    \includegraphics[width=0.1\linewidth]{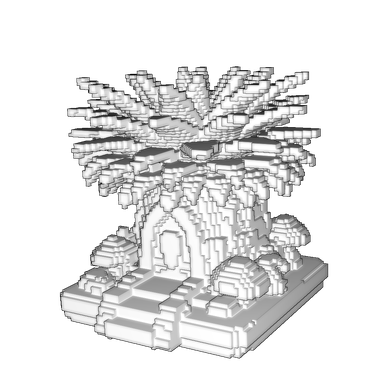}
    \includegraphics[width=0.1\linewidth]{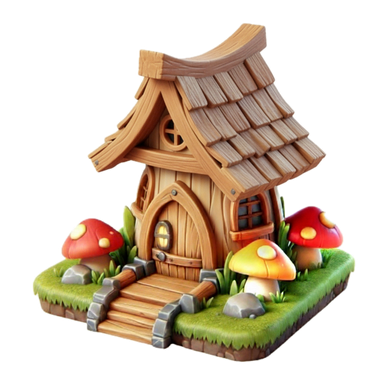}
    \includegraphics[width=0.1\linewidth]{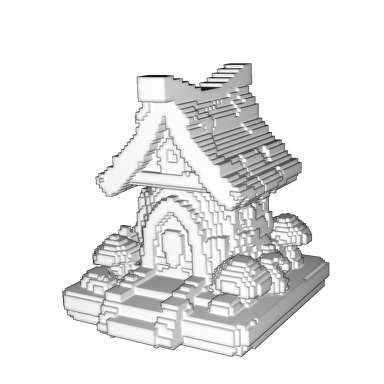}
    \caption{Examples of editing samples with different image prompts. The most left is treated as the reference.}
    \label{fig:variant}
\end{figure}

\begin{figure}
    \centering
    \includegraphics[width=\linewidth]{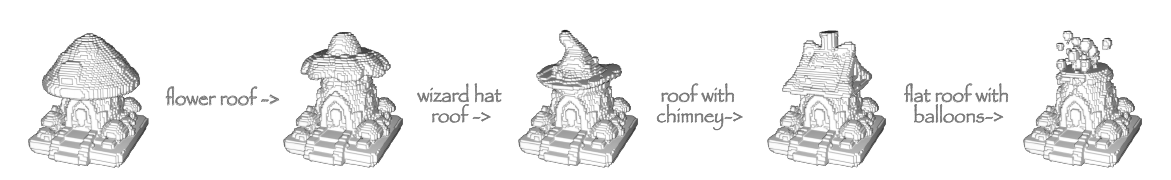}
    \caption{Examples of editing samples with different text prompts. The most left is treated as the reference.}
    \label{fig:variant_txt}
\end{figure}

\clearpage
\newpage
\section{Generation Comparisons with Different Second Stages}\label{appendix:diffpipeline}
\begin{figure}
    \centering
    \includegraphics[width=\linewidth]{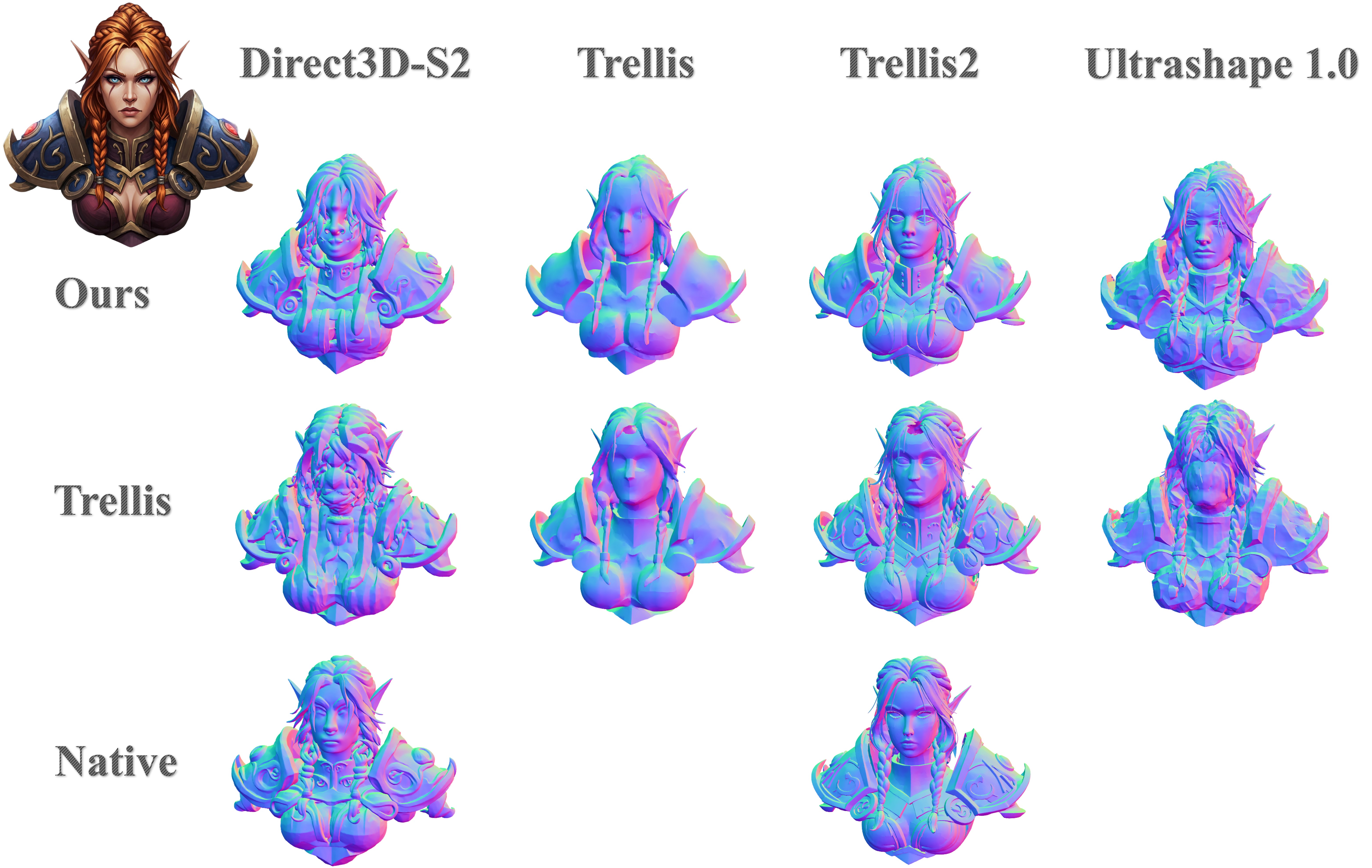}
    \caption{Qualitative comparison using DVD, TRELLIS, and pipeline-native voxels}
    \label{fig:diffpipeline_3}
\end{figure}

\begin{figure}
    \centering
    \includegraphics[width=\linewidth]{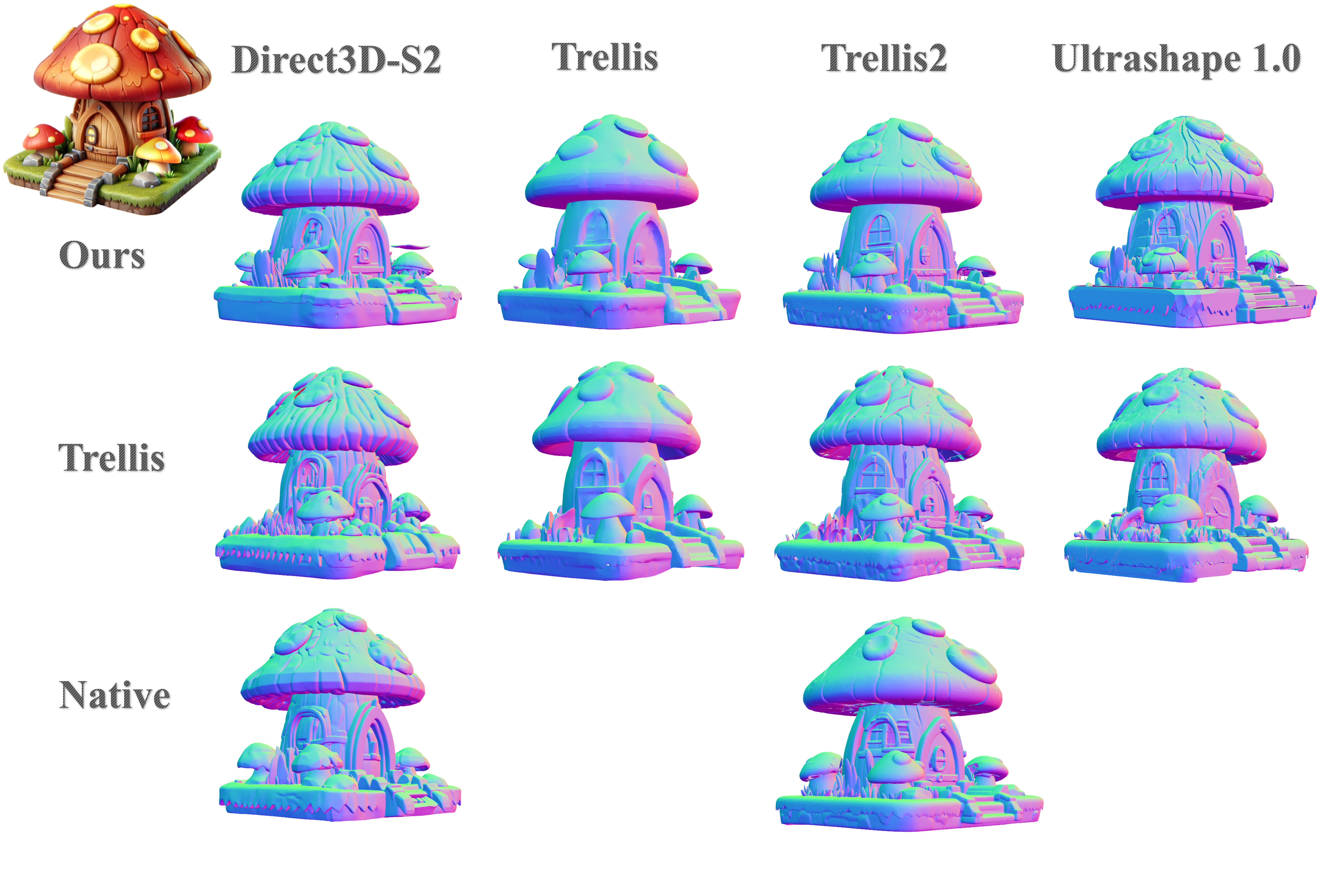}
    \includegraphics[width=\linewidth]{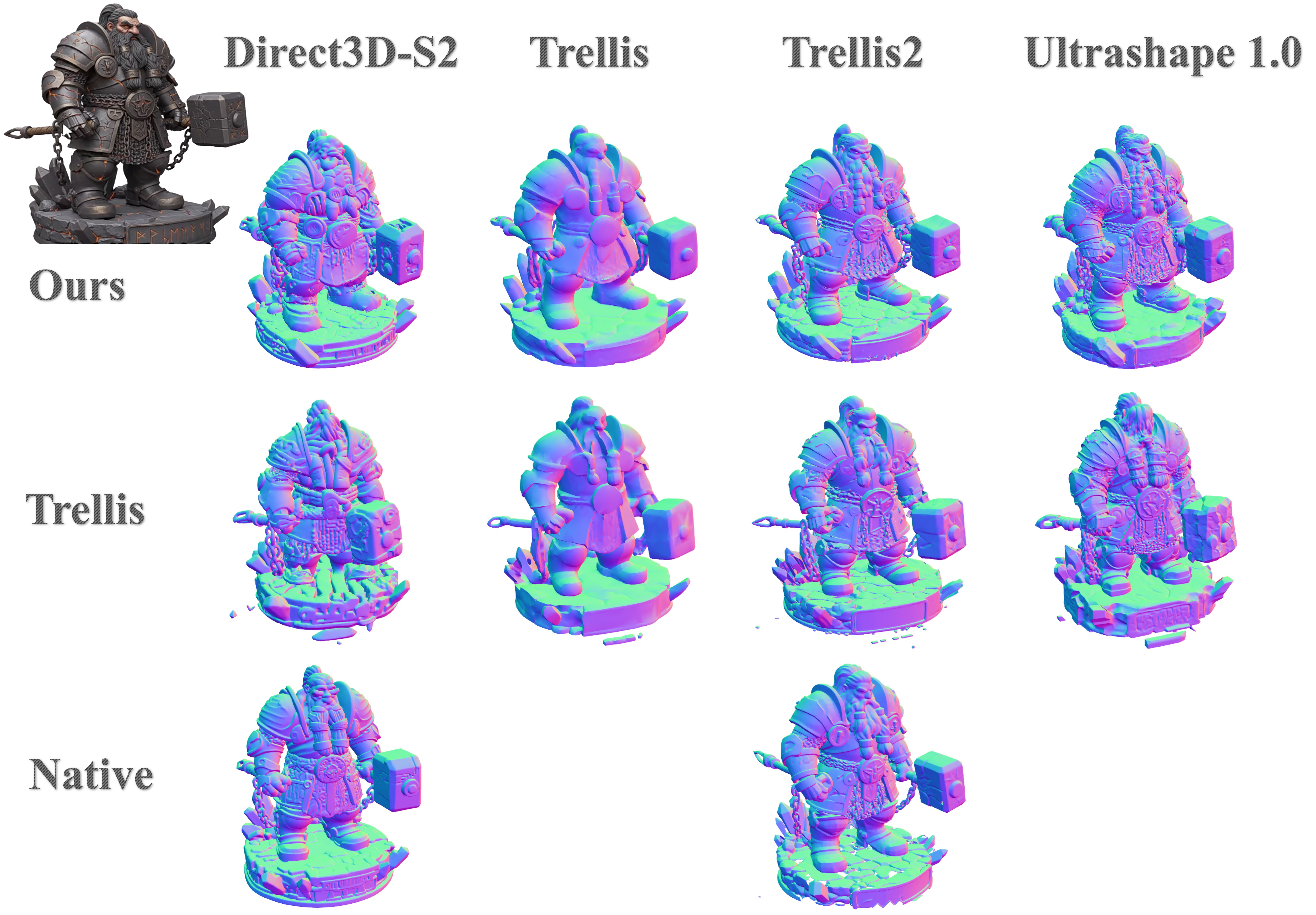}
    \caption{Qualitative comparison using DVD, TRELLIS, and pipeline-native voxels}
    \label{fig:diffpipeline_1}
\end{figure}

\begin{figure}
    \centering
    \includegraphics[width=\linewidth]{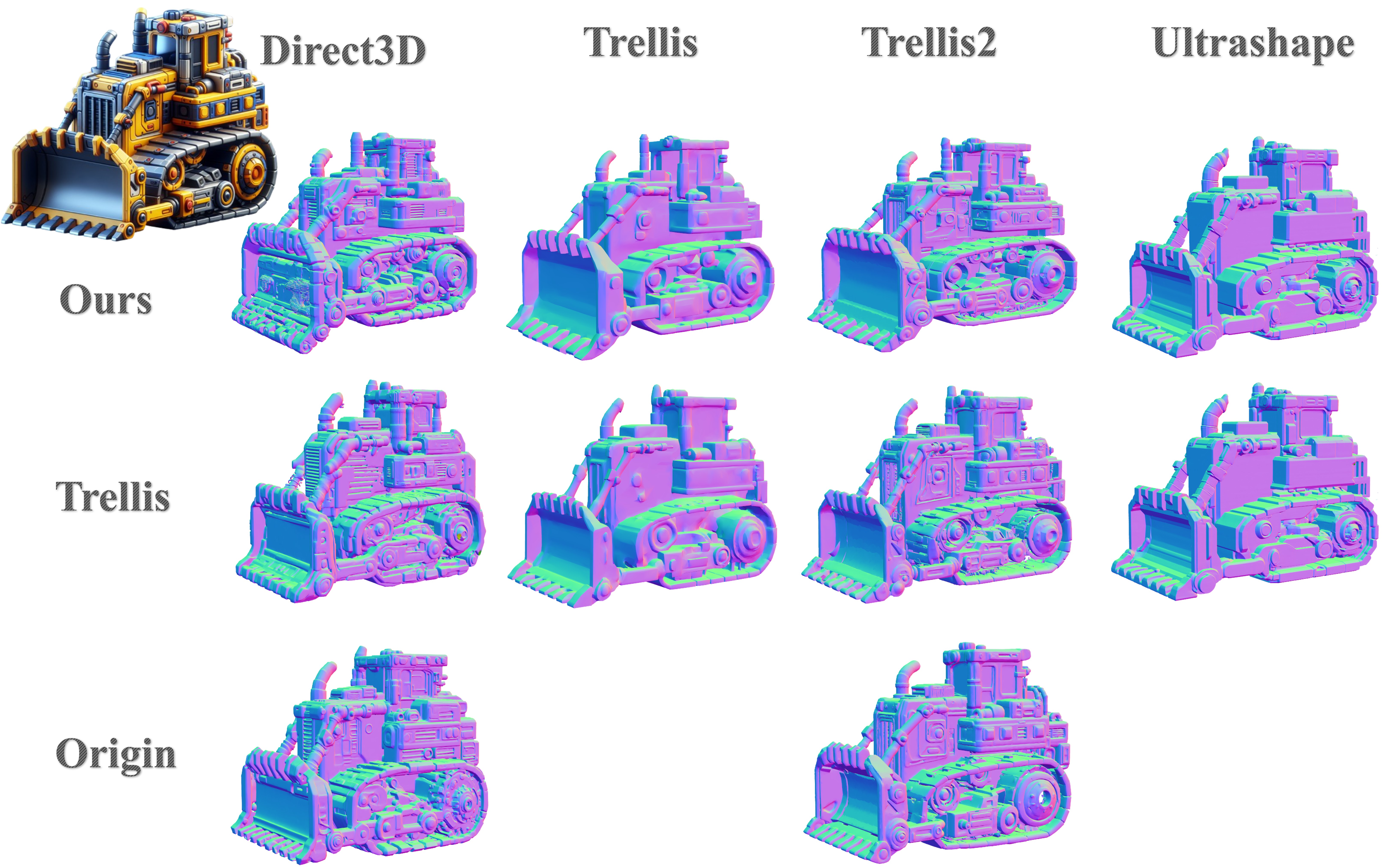}
    \includegraphics[width=\linewidth]{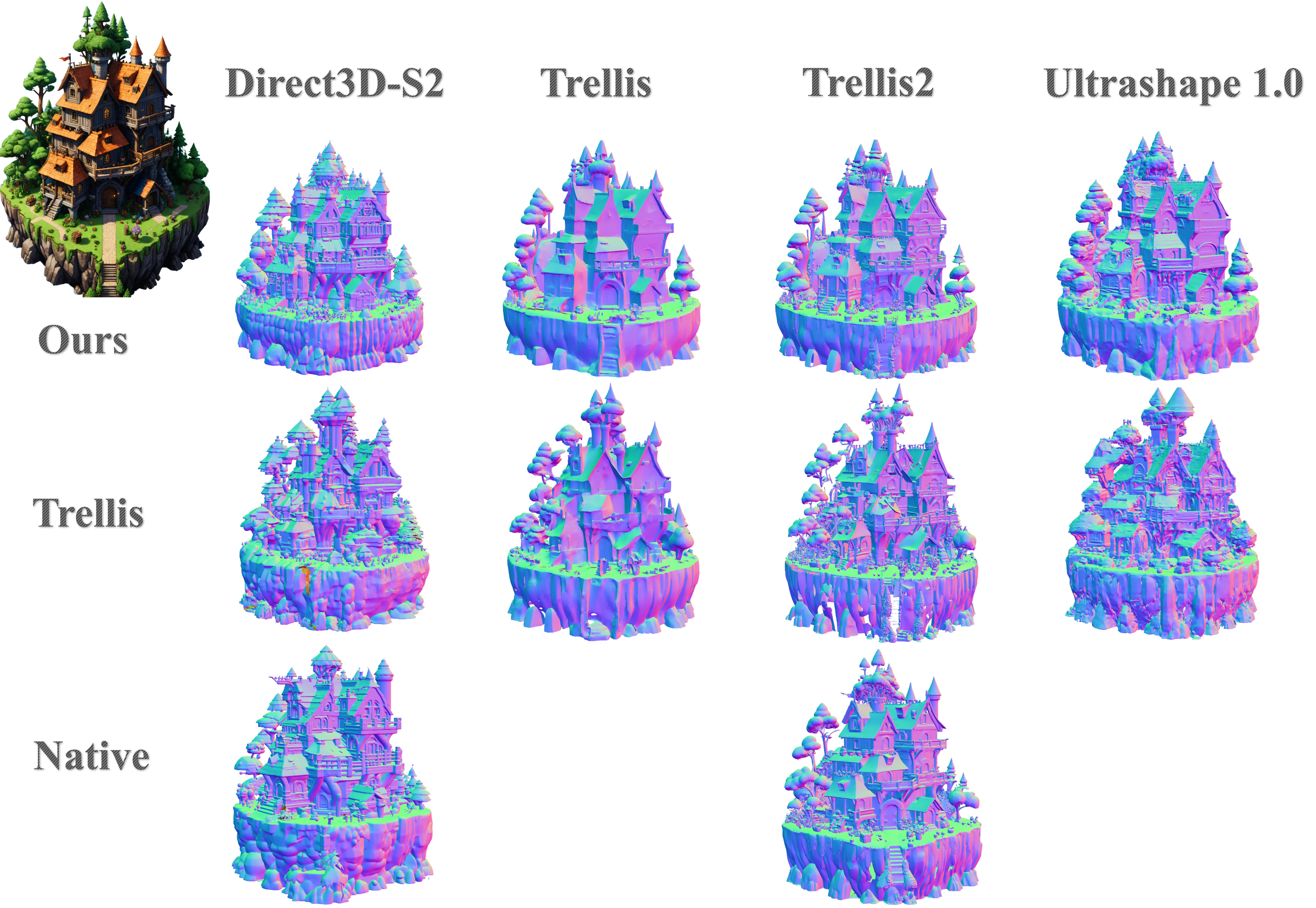}
    \caption{Qualitative comparison using DVD, TRELLIS, and pipeline-native voxels}
    \label{fig:diffpipeline_2}
\end{figure}

Figure~\ref{fig:diffpipeline_3}, \ref{fig:diffpipeline_1}, and \ref{fig:diffpipeline_2} show selected results with generated voxel with different second stages, where each row indicates the source of voxel, image prompts from TRELLIS, Direct3d-S2, and Google Nano Banana.

\paragraph{Implementation details.}
For cross-pipeline evaluation, we replace the first-stage structure of each pipeline with voxels generated by DVD or TRELLIS. Since the downstream pipelines adopt different voxel conventions and input resolutions, we apply pipeline-specific adapter preprocessing before Stage-2 decoding.

For TRELLIS.2, the first-stage structure is a sparse voxel grid at resolution $64^3$, matching the output resolution of DVD and TRELLIS. We therefore directly substitute the original first-stage voxel grid with the voxel grid generated by DVD or TRELLIS, and run the TRELLIS.2 second-stage pipeline at its output resolution of $1024$.

For Direct3D-S2, we found that the second-stage model is sensitive to voxel thickness and empirically expects denser occupancy patterns, likely due to differences in its voxelization convention. Directly using the sparse voxels generated by DVD or TRELLIS often leads to degenerate outputs. To match this convention, we apply one step of 6-neighborhood dilation before feeding the voxels into Direct3D-S2. Let $\mathcal{V} \subset \{0,\ldots,N-1\}^3$ denote the occupied voxel set. We define the dilated set as
\[
\mathcal{V}' =
\mathcal{V}
\cup
\{p + e_x, p - e_x, p + e_y, p - e_y, p + e_z, p - e_z : p \in \mathcal{V}\},
\]
where out-of-bound indices are discarded. Equivalently, a voxel is occupied after dilation if it was originally occupied or if one of its six axis-aligned neighbors was occupied.

UltraShape 1.0 follows a mesh-based first-stage convention. Its official pipeline suggests obtaining an initial mesh from an external pretrained model, such as Hunyuan3D 2.1~\cite{hunyuan3d2025hunyuan3d21imageshighfidelity}, and then voxelizing this mesh internally. In addition, UltraShape 1.0 expects voxel inputs at resolution $128^3$, whereas DVD and TRELLIS generate voxels at $64^3$. Therefore, to adapt DVD and TRELLIS voxels to UltraShape 1.0, we first densely upsample each $64^3$ voxel grid to $128^3$. Since UltraShape's voxelization is surface-oriented, we then remove interior voxels and keep only the surface shell before passing the voxel grid to the UltraShape second-stage pipeline. 

\end{document}